\documentclass{article}

\usepackage[final, nonatbib]{neurips_2021}

\usepackage[utf8]{inputenc} 
\usepackage[T1]{fontenc}    
\usepackage[colorlinks,linkcolor=black,citecolor=blue,urlcolor=blue,bookmarks=true]{hyperref}
\usepackage{url}            
\usepackage{booktabs}       
\usepackage{amsfonts}       
\usepackage{nicefrac}       
\usepackage{microtype}      
\usepackage{xcolor}         
\usepackage{amsmath}
\usepackage{amssymb}
\usepackage{amsthm}
\usepackage{commath}        
\usepackage{wrapfig}
\usepackage{graphicx}
\usepackage{cancel}
\usepackage{tablefootnote}
\usepackage{multirow}
\usepackage{float}
\usepackage[skip=1ex]{caption}
\usepackage{subcaption}\usepackage{algorithm}
\usepackage{algorithmicx}
\usepackage[algo2e,linesnumbered,ruled,vlined]{algorithm2e}

\newcommand{\KL}[2]{\text{KL}\!\left(#1 || #2\right)}
\renewcommand{\H}[1]{\text{H}\!\left(#1\right)}
\newcommand{\CE}[2]{\text{CE}(#1 || #2)}
\newcommand{\bx}{{\mathbf{x}}}
\newcommand{\bz}{{\mathbf{z}}}
\newcommand{\bh}{{\mathbf{h}}}
\newcommand{\bff}{{\mathbf{f}}}
\newcommand{\bw}{{\mathbf{w}}}
\newcommand{\bH}{{\text{\bf H}}}

\newcommand{\beps}{\boldsymbol{\epsilon}}

\newcommand{\bmu}{\boldsymbol{\mu}}
\newcommand{\btheta}{{\boldsymbol{\theta}}}
\newcommand{\bphi}{{\boldsymbol{\phi}}}
\newcommand{\bpsi}{{\boldsymbol{\psi}}}
\newcommand{\balpha}{{\boldsymbol{\alpha}}}
\newcommand{\bzero}{\mathbf{0}}
\newcommand{\beye}{\mathbf{I}}

\newcommand{\E}{{\mathbb{E}}}
\newcommand{\N}{\mathcal{N}}
\newcommand{\U}{\mathcal{U}}
\renewcommand{\L}{\mathcal{L}}

\newcommand{\vmax}{\sigma^2_{\text{max}}}
\newcommand{\vmin}{\sigma^2_{\text{min}}}
\newcommand{\vn}{\mathring{\sigma}^2}

\newcommand{\vo}{\sigma^2_0}
\newcommand{\vt}{\sigma^2_t}
\newcommand{\vtil}{\tilde{\sigma}^2}

\newcommand{\R}{\mathbb{R}}
\newcommand{\D}[1]{\dif #1}

\newtheorem{theorem}{Theorem}

\newcommand{\re}{\text{re}}
\newcommand{\un}{\text{un}}
\newcommand{\we}{\text{ll}}

\title{Score-based Generative Modeling in Latent Space}

\author{%
  Arash Vahdat\thanks{Equal contribution.} \\
  NVIDIA \\
  \texttt{avahdat@nvidia.com} \\
   \And
  Karsten Kreis\footnotemark[1] \\
  NVIDIA \\
  \texttt{kkreis@nvidia.com} \\
   \And
  Jan Kautz \\
  NVIDIA \\
  \texttt{jkautz@nvidia.com} \\
}

\begin{document}

\maketitle

\begin{abstract}
Score-based generative models (SGMs) have recently demonstrated impressive results in terms of both sample quality and distribution coverage. However, they are usually applied directly in data space and often require thousands of network evaluations for sampling. Here, we propose the \textit{Latent Score-based Generative Model} (LSGM), a novel approach that trains SGMs in a latent space, relying on the variational autoencoder framework. Moving from data to latent space allows us to train more expressive generative models, apply SGMs to non-continuous data, and learn smoother SGMs in a smaller space, resulting in fewer network evaluations and faster sampling. To enable training LSGMs end-to-end in a scalable and stable manner, we (i) introduce a new score-matching objective suitable to the LSGM setting, (ii) propose a novel parameterization of the score function that allows SGM to focus on the mismatch of the target distribution with respect to a simple Normal one, and (iii) analytically derive multiple techniques for variance reduction of the training objective. LSGM obtains a state-of-the-art FID score of 2.10 on CIFAR-10, outperforming all existing generative results on this dataset. On CelebA-HQ-256, LSGM is on a par with previous SGMs in sample quality while outperforming them in sampling time by two orders of magnitude. In modeling binary images, LSGM achieves state-of-the-art likelihood on the binarized OMNIGLOT dataset. Our project page and code can be found at \url{https://nvlabs.github.io/LSGM}. 
\end{abstract}
\vspace{-0.3cm}
\section{Introduction}

The long-standing goal of likelihood-based generative learning is to faithfully learn a data distribution, while also generating high-quality samples. 
Achieving these two goals simultaneously is a tremendous challenge, which has led to the development of a plethora of different generative models.
Recently, score-based generative models (SGMs) demonstrated astonishing results in terms of both high sample quality and 
likelihood~\cite{ho2020denoising,song2021scoreSDE}. 
These models define a forward diffusion process that maps data to noise by gradually 
perturbing the input data. 
Generation corresponds to a reverse process that synthesizes novel data via iterative denoising, starting from random noise.
The problem then reduces to learning \textit{the score function}---the gradient of the log-density---of the perturbed data~\cite{song2019scorematching}. 
In a seminal work, Song et al.~\cite{song2021scoreSDE} show how this modeling approach is described with a stochastic differential equation (SDE) framework which can be converted to maximum likelihood training~\cite{song2021MLSGM}.
Variants of SGMs have been applied to images~\cite{ho2020denoising,song2021scoreSDE,song2021denoising,dhariwal2021diffusion}, audio~\cite{chen2020wavegrad,kong2020diffwave,jeong2021difftts,mittal2021symbolic}, graphs~\cite{niu2020scorebasedgraphs} and point clouds~\cite{cai2020gradientfield,luo2021diffusion}. 

Albeit high quality, sampling from SGMs is computationally expensive. This is because generation amounts to solving a complex SDE, or equivalently ordinary differential equation (ODE) (denoted as the \textit{probability flow ODE} in~\cite{song2021scoreSDE}), that maps a simple base distribution to the complex data distribution. 
The resulting differential equations are typically complex and solving them accurately requires numerical integration with very small step sizes, which results in thousands of neural network evaluations~\cite{ho2020denoising,song2021scoreSDE,dhariwal2021diffusion}.
Furthermore, generation complexity is uniquely defined by the underlying data distribution and the forward SDE for data perturbation, implying that synthesis speed cannot be increased easily without sacrifices. Moreover, SDE-based generative models are currently defined for continuous data and cannot be applied effortlessly to binary, categorical, or graph-structured data.

Here, we propose the \textit{Latent Score-based Generative Model} (LSGM), a new approach for learning SGMs in latent space, leveraging a variational autoencoder (VAE) framework~\cite{kingma2014vae,rezende2014stochastic}. We map the input data to latent space and apply the score-based generative model there. The score-based model is then tasked with modeling the distribution over the embeddings of the data set. Novel data synthesis is achieved by first generating embeddings via drawing from a simple base distribution followed by iterative denoising, and then transforming this embedding via a decoder to data space (see Fig.~\ref{fig:pipeline}). We can consider this model a VAE with an SGM prior.
Our approach has several key advantages:
\begin{figure}[t!]
    \centering
    \vspace{-0.3cm}
    \includegraphics[width=0.96\textwidth]{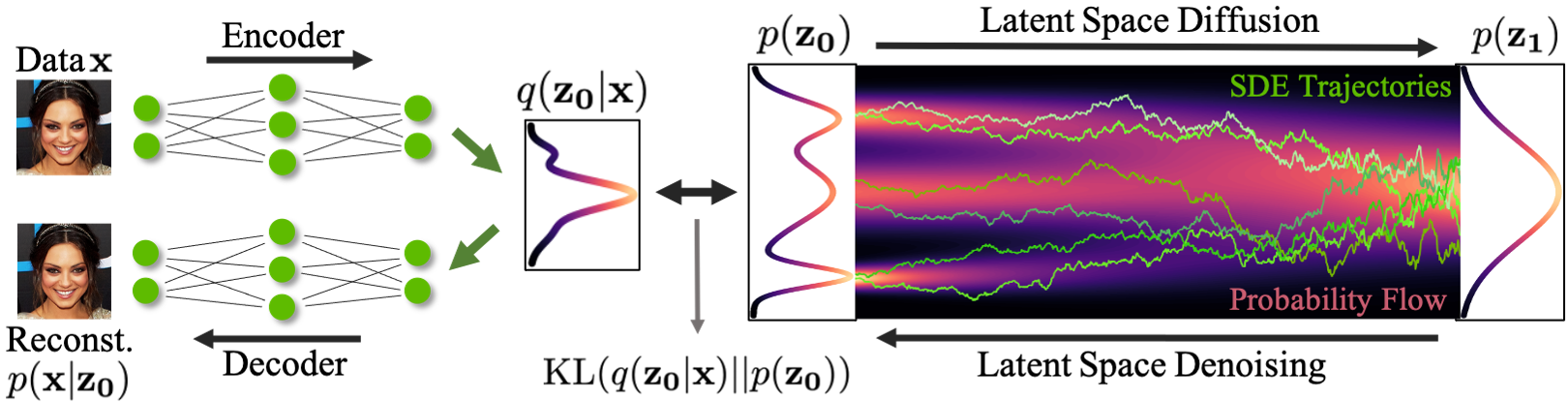}
    \caption{\small In our latent score-based generative model (LSGM), data is mapped to latent space via an encoder $q(\bz_0|\bx)$ and a diffusion process is applied in the latent space ($\bz_0 \rightarrow \bz_1$). Synthesis starts from the base distribution $p(\bz_1)$ and generates samples in latent space via denoising ($\bz_0 \leftarrow \bz_1$). Then, the samples are mapped from latent to data space using a decoder $p(\bx|\bz_0)$. The model is trained end-to-end.}
    \vspace{-5mm}
    \label{fig:pipeline}
\end{figure}

\vspace{-0.1cm}
\textbf{Synthesis Speed:} By pretraining the VAE with a Normal prior first, we can bring the marginal distribution over encodings (the \textit{aggregate posterior}) close to the Normal prior, which is also the SGM's base distribution. Consequently, the SGM only needs to model the remaining mismatch, resulting in a less complex model from which sampling becomes easier. Furthermore, we can tailor the latent space according to our needs. For example, we can use hierarchical latent variables and apply the diffusion model only over a subset of them, further improving synthesis speed.

\vspace{-0.1cm}
\textbf{Expressivity:} Training a regular SGM can be considered as training a neural ODE directly on the data~\cite{song2021scoreSDE}. However, previous works found that augmenting neural ODEs~\cite{dupont2019augmented,kong2020expressive} and more generally generative models~\cite{huang2020anf,chen2020vflow,vahdat2020nvae,child2021VDVAE} with latent variables improves their expressivity. Consequently, we expect similar performance gains from combining SGMs with a latent variable framework.

\vspace{-0.1cm}
\textbf{Tailored Encoders and Decoders:} Since we use the SGM in latent space, we can utilize carefully designed encoders and decoders mapping between latent and data space, further improving expressivity. Additionally, the LSGM method can therefore be naturally applied to non-continuous data.

LSGMs can be trained end-to-end by maximizing the variational lower bound on the data likelihood. Compared to regular score matching, our approach comes with additional challenges, since both the score-based denoising model and its target distribution, formed by the latent space encodings, are learnt simultaneously. To this end, we make the following technical contributions: \textbf{(i)} We derive a new denoising score matching objective that allows us to efficiently learn the VAE model and the latent SGM prior at the same time. \textbf{(ii)} We introduce a new parameterization of the latent space score function, which mixes a Normal distribution with a learnable SGM, allowing the SGM to model only the mismatch between the distribution of latent variables and the Normal prior. \textbf{(iii)} We propose techniques for variance reduction of the training objective by designing a new SDE and by analytically deriving importance sampling schemes, allowing us to stably train deep LSGMs. Experimentally, we achieve state-of-the-art 2.10 FID on CIFAR-10 and 7.22 FID on CelebA-HQ-256, and significantly improve upon likelihoods of previous SGMs. On CelebA-HQ-256, we outperform previous SGMs in synthesis speed by two orders of magnitude.
We also model binarized images, MNIST and OMNIGLOT, achieving state-of-the-art likelihood on the latter.

\vspace{-0.2cm}
\section{Background} \label{sec:background}
\vspace{-0.2cm}
Here, we review continuous-time score-based generative models (see \cite{song2021scoreSDE} for an in-depth discussion). Consider a forward diffusion process $\{\bz_t\}^{t=1}_{t=0}$ for continuous time variable $t \in [0, 1]$, where $\bz_0$ is the starting variable and $\bz_t$ its perturbation at time $t$. The diffusion process is defined by an It\^{o} SDE: 
\begin{equation} \label{eq:forward_ito}
    \D{\bz} = f(t)\bz\D{t} + g(t)\D{\bw}
\end{equation}
where $f: \R \rightarrow \R$ and $g: \R \rightarrow \R$ are scalar drift and diffusion coefficients, respectively, and $\bw$ is the standard Wiener process. $f(t)$ and $g(t)$ can be designed such that $\bz_1 \sim \N(\bz_1; \bzero, \beye)$ follows a Normal distribution at the end of the diffusion process.\footnote{Other distributions at $t=1$ are possible; for instance, see the ``variance-exploding'' SDE in \cite{song2021scoreSDE}. In this paper, however, we use only SDEs converging towards $\N(\bz_1; \bzero, \beye)$ at $t=1$.}
Song et al.~\cite{song2021scoreSDE} show that the SDE in Eq.~\ref{eq:forward_ito} can be converted to a generative model by first sampling from 
$\bz_1 \sim \N(\bz_1; \bzero, \beye)$ and then running the reverse-time SDE
$\D{\bz} = [f(t)\bz - g(t)^2 \nabla_\bz \log q_t(\bz)] \D{t} + g(t)\D{\bar{\bw}}$, 
where $\bar{\bw}$ is a reverse-time standard Wiener process and $\D{t}$ is an infinitesimal negative time step. The reverse SDE requires knowledge of $\nabla_{\bz_t} \log q_t(\bz_t)$, the score function of the marginal 
distribution under the forward diffusion at time $t$. One approach for estimating it is via the score matching objective\footnote{We omit the $t$-subscript of the diffused distributions $q_t$ in all score functions of the form $\nabla_{\bz_t}\log q_t(\bz_t)$.}:
\begin{equation} \label{eq:score_matching}
    \min_{\btheta} \E_{t \sim \U[0, 1]} \left[\lambda(t) \E_{q(\bz_0)} \E_{q(\bz_t|\bz_0)} [||\nabla_{\bz_t} \log q(\bz_t) - \nabla_{\bz_t} \log p_{\btheta}(\bz_t) ||_2^2 ]\right]
\end{equation}
that trains the parameteric score function $\nabla_{\bz_t} \log p_{\btheta}(\bz_t)$ at time $t \sim \U[0, 1]$ for a given weighting coefficient $\lambda(t)$. 
$q(\bz_0)$ is the $\bz_0$-generating distribution and $q(\bz_t |\bz_0)$ is the diffusion kernel, which is available in closed form for certain $f(t)$ and $g(t)$. Since $\nabla_{\bz_t} \log q(\bz_t)$ is not analytically available, Song et al.~\cite{song2021scoreSDE} rely on denoising score matching~\cite{vincent2011connection} that converts the objective in Eq.~\ref{eq:score_matching} to:
\begin{equation} \label{eq:training_song}
    \min_{\btheta} \E_{t \sim \U[0, 1]} \left[\lambda(t) \E_{q(\bz_0)} \E_{q(\bz_t|\bz_0)} [||\nabla_{\bz_t} \log q(\bz_t |\bz_0) - \nabla_{\bz_t} \log p_{\btheta}(\bz_t) ||_2^2 ]\right] + C
\end{equation}
Vincent~\cite{vincent2011connection} shows $C =  \E_{t \sim \U[0, 1]} [\lambda(t) \E_{q(\bz_0)} \E_{q(\bz_t|\bz_0)} [||\nabla_{\bz_t} \log q(\bz_t)||_2^2 - ||\nabla_{\bz_t} \log q(\bz_t|\bz_0) ||_2^2 ]]$
is independent of $\btheta$, making the minimizations in Eq.~\ref{eq:training_song} and Eq.~\ref{eq:score_matching} equivalent. 
Song et al.~\cite{song2021MLSGM} show that for $\lambda(t) = g(t)^2/2$, the minimizations correspond to approximate maximum likelihood training based on an upper on the Kullback-Leibler (KL) divergence between the target distribution and the distribution defined by the reverse-time generative SDE with the learnt score function. In particular, the objective of Eq.~\ref{eq:score_matching} can then be written:
{\small
\begin{align} \label{eq:durkan_kl}
     \KL{q(\bz_0)}{ p_{\btheta}(\bz_0)} \leq \E_{t \sim \U[0, 1]}\left[ \frac{g(t)^2}{2} \E_{q(\bz_0)} \E_{q(\bz_t|\bz_0)} \left[||\nabla_{\bz_t} \log q(\bz_t) - \nabla_{\bz_t} \log p_{\btheta}(\bz_t) ||_2^2 \right]\right]
\end{align}}
which can again be transformed into denoising score matching (Eq.~\ref{eq:training_song}) following Vincent~\cite{vincent2011connection}.

\vspace{-0.2cm}
\section{Score-based Generative Modeling in Latent Space}
\vspace{-0.1cm}

The LSGM framework in Fig.~\ref{fig:pipeline} consists of the encoder $q_\bphi(\bz_0|\bx)$, SGM prior $p_\btheta(\bz_0)$, 
and decoder $p_\bpsi(\bx | \bz_0)$. The SGM prior leverages a diffusion process as defined in Eq.~\ref{eq:forward_ito} and diffuses $\bz_0 \sim q_\bphi(\bz_0|\bx)$ samples in latent space to the standard Normal distribution $p(\bz_1) = \N(\bz_1; \bzero, \beye)$. Generation uses the reverse SDE to sample from $p_\btheta(\bz_0)$ with time-dependent score function $\nabla_{\bz_t} \log p_\btheta(\bz_t)$, and the decoder $p_\bpsi(\bx | \bz_0)$ to map the synthesized encodings $\bz_0$ to data space. 
Formally, the generative process is written as $p(\bz_0, \bx)=p_\btheta(\bz_0)p_\bpsi(\bx | \bz_0 )$. The goal of training is to learn $\{\bphi, \btheta, \bpsi\}$, the parameters of the encoder $q_\bphi(\bz_0|\bx)$, score function $\nabla_{\bz_t} \log p_\btheta(\bz_t)$, and decoder $p_\bpsi(\bx | \bz_0 )$, respectively.

We train LSGM by minimizing the variational upper bound on negative data log-likelihood $\log p(\bx)$:
{\small
\begin{align} \label{eq:elbo_kl} \hspace{-0.7cm}
    \L(\bx, \bphi, \btheta, \bpsi) &= \E_{q_\bphi(\bz_0|\bx)}\left[-\log p_\bpsi(\bx|\bz_0) \right]\! +\!\KL{q_\bphi(\bz_0|\bx)}{p_\btheta(\bz_0)} \\
    &= \underbrace{\E_{q_\bphi(\bz_0|\bx)}\left[-\log p_\bpsi(\bx|\bz_0) \right]}_{\text{reconstruction term}}\!+\!\underbrace{\E_{q_\phi(\bz_0|\bx)}\left[\log q_\bphi(\bz_0 | \bx) \right]}_{\text{negative encoder entropy}}\!+\! \underbrace{\E_{q_\bphi(\bz_0|\bx)}\left[-\log p_\btheta(\bz_0) \right]}_{\text{cross entropy}} \label{eq:elbo_ce}
\end{align}}%
following a VAE approach~\cite{kingma2014vae,rezende2014stochastic}, where $q_\bphi(\bz_0|\bx)$ approximates the true posterior $p(\bz_0|\bx)$.

In this paper, we use Eq.~\ref{eq:elbo_ce} with decomposed KL
divergence 
into its entropy and cross entropy terms. The reconstruction and entropy terms
are estimated easily for any explicit encoder as long as the reparameterization trick is available~\cite{kingma2014vae}. The challenging part in training LSGM is to train the cross entropy term that involves the SGM prior. We motivate and present our expression for the cross-entropy term in Sec.~\ref{sec:ce}, the parameterization of the SGM prior in Sec.~\ref{sec:param}, different weighting mechanisms for the training objective in Sec.~\ref{sec:weighting}, and variance reduction techniques in Sec.~\ref{sec:variance}.

\vspace{-0.2cm}
\subsection{The Cross Entropy Term} \label{sec:ce}
One may ask, why not train LSGM with Eq.~\ref{eq:elbo_kl} and rely on the KL in Eq.~\ref{eq:durkan_kl}. Directly using the KL expression in Eq.~\ref{eq:durkan_kl} is not possible, as it involves the marginal score $\nabla_{\bz_t} \log q(\bz_t)$, which is unavailable analytically for common non-Normal distributions $q(\bz_0)$ such as Normalizing flows. Transforming into denoising score matching does not help either, since in that case the problematic $\nabla_{\bz_t} \log q(\bz_t)$ term appears in the $C$ term (see Eq.~\ref{eq:training_song}). In contrast to previous works~\cite{song2021scoreSDE, vincent2011connection}, we cannot simply drop $C$, since it is, in fact, not constant but depends on $q(\bz_t)$, which is trainable in our setup.

To circumvent this problem, we instead decompose the KL in Eq.~\ref{eq:elbo_kl} and rather work directly with the cross entropy between the encoder distribution $q(\bz_0|\bx)$ and the SGM prior $p(\bz_0)$. We show:

\begin{theorem} \label{th:simple_ce}
Given two distributions $q(\bz_0|\bx)$ and $p(\bz_0)$, defined in the continuous space $\R^D$, denote the marginal distributions of diffused samples under the SDE in Eq.~\ref{eq:forward_ito} at time $t$ with $q(\bz_t|\bx)$ and $p(\bz_t)$. Assuming mild smoothness conditions on $\log q(\bz_t|\bx)$ and $\log p(\bz_t)$, the cross entropy is:
{\small
\begin{align*}
 \CE{q(\bz_0|\bx)}{p(\bz_0)} = \E_{t \sim \U[0, 1]}\left[\frac{g(t)^2}{2} \E_{q(\bz_t, \bz_0|\bx)} \left[||\nabla_{\bz_t} \log q(\bz_t|\bz_0)\!-\!\nabla_{\bz_t} \log p(\bz_t)||_2^2 \right]\right]\!+\!\frac{D}{2} \log\left(2\pi e\vo\right),
\end{align*}}%
with $q(\bz_t, \bz_0|\bx) = q(\bz_t|\bz_0)q(\bz_0|\bx)$ and a Normal transition kernel $q(\bz_t | \bz_0) = \N(\bz_t; \bmu_t(\bz_0), \vt \beye)$, where $\bmu_t$ and $\vt$ are obtained from $f(t)$ and $g(t)$ for a fixed initial variance $\vo$ at $t=0$. 
\end{theorem}
A proof with generic expressions for $\bmu_t$ and $\vt$ as well as an intuitive interpretation are in App.~\ref{app:proof}.

Importantly, unlike for the KL objective of Eq.~\ref{eq:durkan_kl}, no problematic terms depending on the marginal score $\nabla_{\bz_t} \log q(\bz_t|\bx)$ arise. This allows us to use this denoising score matching objective for the cross entropy term in Theorem~\ref{th:simple_ce} not only for optimizing $p(\bz_0)$ (which is commonly done in the score matching literature), but also for the $q(\bz_0|\bx)$ encoding distribution. It can be used even with complex $q(\bz_0|\bx)$ distributions, defined, for example, in a hierarchical fashion~\cite{vahdat2020nvae,child2021VDVAE} or via Normalizing flows~\cite{rezendeICML15Normalizing,kingma2016improved}. Our novel analysis shows that, for diffusion SDEs following Eq.~\ref{eq:forward_ito}, only the cross entropy can be expressed purely with $\nabla_{\bz_t} \log q(\bz_t|\bz_0)$. Neither KL nor entropy in \cite{song2021MLSGM} can be expressed without the problematic term $\nabla_{\bz_t} \log q(\bz_t|\bx)$ (details in the Appendix).

Note that in Theorem~\ref{th:simple_ce}, 
the term $\nabla_{\bz_t} \log p(\bz_t)$ in the score matching expression corresponds to the score that originates from diffusing an initial $p(\bz_0)$ distribution. In practice, we use the expression to learn an SGM prior $p_{\boldsymbol{\theta}}(\bz_0)$, which models $\nabla_{\bz_t} \log p(\bz_t)$ by a neural network. With the learnt score $\nabla_{\bz_t} \log p_{\boldsymbol{\theta}}(\bz_t)$ (here we explicitly indicate the parameters $\boldsymbol{\theta}$ to clarify that this is the learnt model), the actual SGM prior is defined via the generative reverse-time SDE (or, alternatively, a closely-connected ODE, see Sec.~\ref{sec:background} and App.~\ref{app:prob_flow_ode}), which generally defines its own, separate marginal distribution $p_{\boldsymbol{\theta}}(\bz_0)$ at $t=0$. Importantly, the learnt, approximate score $\nabla_{\bz_t} \log p_{\boldsymbol{\theta}}(\bz_t)$ is not necessarily the same as one would obtain when diffusing $p_{\boldsymbol{\theta}}(\bz_0)$.
Hence, when considering the learnt score $\nabla_{\bz_t} \log p_{\boldsymbol{\theta}}(\bz_t)$, the score matching expression in our Theorem only corresponds to an upper bound on the cross entropy between $q(\bz_0|\bx)$ and $p_{\boldsymbol{\theta}}(\bz_0)$ defined by the generative reverse-time SDE.
This is discussed in detail in concurrent works~\cite{song2021MLSGM, huang2021variational}. Hence, from the perspective of the learnt SGM prior, we are training with an upper bound on the cross entropy (similar to the bound on the KL in Eq.~\ref{eq:durkan_kl}), which can also be considered as the continuous version of the discretized variational objective derived by Ho et al.~\cite{ho2020denoising}.

\vspace{-0.2cm}
\subsection{Mixing Normal and Neural Score Functions} \label{sec:param}

In VAEs~\cite{kingma2014vae}, $p(\bz_0)$ is often chosen as a standard Normal $\N(\bz_0; \bzero, \beye)$. For recent hierarchical VAEs~\cite{vahdat2020nvae,child2021VDVAE}, using the reparameterization trick, the prior can be converted to $\N(\bz_0; \bzero, \beye)$ (App.~\ref{app:hvae}).

Considering a single dimensional latent space, we can assume that the prior at time $t$ is in the form of a geometric mixture $p(z_t) \propto \N(z_t; 0, 1)^{1 - \alpha} p'_\btheta(z_t)^\alpha$ where $p'_\btheta(z_t)$ is a trainable SGM prior and $\alpha \in [0, 1]$ is a learnable scalar mixing coefficient. Formulating the prior this way has crucial advantages: (i) We can pretrain LSGM's autoencoder networks assuming $\alpha{=}0$, which corresponds to training the VAE with a standard Normal prior. This pretraining step will bring the distribution of latent variable close to $\N(z_0; 0, 1)$, allowing the SGM prior to learn a much simpler distribution in the following end-to-end training stage. (ii) The score function for this mixture is of the form $\nabla_{z_t} \log p(z_t) = -(1 - \alpha)z_t + \alpha \nabla_{z_t} \log p'_{\btheta}(z_t)$.
When the score function is dominated by the linear term, we expect that the reverse SDE can be solved faster, as its drift is dominated by this linear term. 

For our multivariate latent space, we obtain diffused samples at time $t$ by sampling $\bz_t \sim q(\bz_t|\bz_0)$ with $\bz_t = \bmu_t(\bz_0) + \sigma_t \beps$, where $\beps \sim \N(\beps; \bzero, \beye)$. Since we have $\nabla_{\bz_t} \log q(\bz_t|\bz_0) = -{\beps}/{\sigma_t}$, similar to \cite{ho2020denoising}, we parameterize the score function by $\nabla_{\bz_t} \log p(\bz_t) := - {\beps_\theta(\bz_t, t)}/{\sigma_t}$, where $\beps_\theta(\bz_t, t) := \sigma_t (1 - \balpha) \odot \bz_t + \balpha \odot \beps'_\theta(\bz_t, t)$ is defined by our \textit{mixed score parameterization} that is applied elementwise to the components of the score. With this, we simplify the cross entropy expression to:

{\small \vspace{-0.5cm}
\begin{align}\label{eq:simple_ce_eps}
 \CE{q_\bphi(\bz_0|\bx)}{p_\btheta(\bz_0)} = \E_{t \sim \U[0, 1]}\left[ \frac{w(t)}{2} \E_{q_\bphi(\bz_t, \bz_0|\bx), \beps} \left[||\beps\!-\!\beps_\btheta(\bz_t, t) ||_2^2 \right]\right]\!+\!\frac{D}{2} \log\left(2\pi e\sigma_0^2\right),
\end{align}}
where $w(t) = g(t)^2 / \vt$ is a time-dependent weighting scalar.

\vspace{-0.1cm}
\subsection{Training with Different Weighting Mechanisms} \label{sec:weighting}

\begin{wraptable}{r}{4.2cm}
\vspace{-11mm}
\centering
{\footnotesize
\setlength{\tabcolsep}{2pt}
\caption{\small Weighting mechanisms}\label{table:weighting}
    \begin{tabular}{ll}
    \toprule
    Mechanism    & Weights \\
    \midrule
    Weighted     &  $w_{\we}(t) = g(t)^2 / \vt $  \\
    Unweighted   &  $w_{\un}(t) = 1$ \\
    Reweighted   &  $w_{\re}(t) = g(t)^2$ \\ 
    \bottomrule
    \end{tabular}}
\vspace{-4mm}
\end{wraptable}
The weighting term $w(t)$ in Eq.~\ref{eq:simple_ce_eps} trains the prior with maximum likelihood. Similar to \cite{ho2020denoising,song2021scoreSDE}, we observe that when $w(t)$ is dropped while training the SGM prior (i.e., $w(t)=1$), LSGM often yields higher quality samples at a small cost in likelihood. However, in our case, we can only drop the weighting when training the prior. When updating the encoder parameters, we still need to use the maximum likelihood weighting to ensure that the encoder $q(\bz_0|\bx)$ is brought closer to the true posterior $p(\bz_0|\bx)$\footnote{Minimizing $\L(\bx, \bphi, \btheta, \bpsi)$ w.r.t $\bphi$ is equivalent to minimizing $\KL{q(\bz_0|\bx)}{p(\bz_0|\bx)}$ w.r.t $q(\bz_0|\bx)$.}. Tab.~\ref{table:weighting} summarizes three weighting mechanisms we consider in this paper: $w_{\we}(t)$ corresponds to maximum likelihood, $w_{\un}(t)$ is the unweighted objective used by \cite{ho2020denoising, song2021scoreSDE}, and $w_{\re}(t)$ is a variant obtained by dropping only $1/\vt$. 
This weighting mechanism has a similar affect on the sample quality as $w_{\un}(t) = 1$; however, in Sec.~\ref{sec:variance}, we show that it is easier to define a variance reduction scheme for this weighting mechanism.  

The following summarizes our training objectives (with $t \sim \U[0, 1]$ and $\beps\sim \N(\beps; \bzero, \beye)$):

{\small \vspace{-0.5cm}\phantom{.}
\begin{align}
\hspace{-0.6cm} \min_{\bphi, \bpsi} \ & \E_{q_\bphi(\bz_0|\bx)}\left[-\!\log p_\bpsi(\bx|\bz_0) \right]\!+\!\E_{q_\phi(\bz_0|\bx)}\left[\log q_\bphi(\bz_0 | \bx) \right]\!+\!\E_{t,\beps,q(\bz_t|\bz_0),q_\bphi(\bz_0|\bx)} \left[\frac{w_\we(t)}{2}||\beps\!-\!\beps_\btheta(\bz_t, t) ||_2^2 \right]  \label{eq:vae_obj} \\
\hspace{-0.6cm} \min_{\btheta} \ & \E_{t,\beps, q(\bz_t|\bz_0),q_\bphi(\bz_0|\bx)} \left[\frac{ w_{\we/\un/\re}(t)}{2}||\beps\!-\!\beps_\btheta(\bz_t, t) ||_2^2 \right]  \quad \text{with} \quad q(\bz_t | \bz_0) = \N(\bz_t; \bmu_t(\bz_0), \vt \beye), \label{eq:prior_obj}
\end{align}}%
where Eq.~\ref{eq:vae_obj} trains the VAE encoder and decoder parameters $\{\bphi, \bpsi\}$ using the variational bound $\L(\bx, \bphi, \btheta, \bpsi)$ from Eq.~\ref{eq:elbo_ce}. Eq.~\ref{eq:prior_obj} trains the prior with one of the three weighting mechanisms. Since the SGM prior participates in the objective only in the cross entropy term, we only consider this term when training the prior. Efficient algorithms for training with the objectives are presented in App.~\ref{app:impl}.

\vspace{-0.2cm}
\subsection{Variance Reduction} \label{sec:variance}
\vspace{-0.1cm}
The objectives in Eqs.~\ref{eq:vae_obj} and~\ref{eq:prior_obj} involve 
sampling of the time variable $t$,
which has high variance~\cite{nichol2021improved}. 
We introduce several techniques for reducing this variance for all three objective weightings. 
We focus on the ``variance preserving'' SDEs (VPSDEs)~\cite{song2021scoreSDE, ho2020denoising, sohldickstein2015thermodynamics}, defined by $\D{\bz} = - \frac{1}{2} \beta(t) \bz \D{t} + \sqrt{\beta(t)} \D{\bw}$ where $\beta(t)=\beta_0+(\beta_1-\beta_0)t$ linearly interpolates in $[\beta_0, \beta_1]$ (other SDEs discussed in App.~\ref{app:var}).

We denote the marginal distribution of latent variables by $q(\bz_0) := \E_{p_{\text{data}}(\bx)}[q(\bz_0|\bx)]$. Here, we derive variance reduction techniques for $\CE{q(\bz_0)}{p(\bz_0)}$, assuming that both $q(\bz_0) = p(\bz_0)=\N(\bz_0; \bzero, \beye)$. This is a reasonable simplification for our analysis because pretraining our LSGM model with a $\N(\bz_0; \bzero, \beye)$ prior brings $q(\bz_0)$ close to $\N(\bz_0; \bzero, \beye)$ and our SGM prior is often dominated by the fixed Normal mixture component. 
We empirically observe that the variance reduction techniques developed with this assumption still work well when $q(\bz_0)$ and $p(\bz_0)$ are not exactly $\N(\bz_0; \bzero, \beye)$.

\textbf{Variance reduction for likelihood weighting:}  In App.~\ref{app:var}, for $q(\bz_0) = p(\bz_0) = \N(\bz_0; \bzero, \beye)$, we show $\CE{q(\bz_0)}{p(\bz_0)}$ is given by $\frac{D}{2}\E_{t\sim \U[0, 1]}[ {\D\,\log \vt }/{\D{t}}] + \text{const.}$
We consider two approaches:

\textbf{(1)} \textit{Geometric VPSDE}: To reduce the variance sampling uniformly from $t$, we can design the SDE such that ${\D\,\log \vt }/{\D{t}}$ is constant for $t \in [0, 1]$. We show in App.~\ref{app:var} that a $\beta(t) = \log({\sigma^2_\text{max}}/{\sigma^2_\text{min}})\frac{\vt}{(1 - \vt)}$ with geometric variance $\vt = \sigma^2_\text{min}({\sigma^2_\text{max}}/{\sigma^2_\text{min}})^t$ satisfies this condition. We call a VPSDE with this $\beta(t)$ a geometric VPSDE. $\sigma^2_\text{min}$ and $\sigma^2_\text{max}$ are the hyperparameters of the SDE, with $0\!<\!\vmin\!<\!\vmax\!<\!1$. 
Although our geometric VPSDE has a geometric variance progression similar to the ``variance exploding'' SDE (VESDE)~\cite{song2021scoreSDE}, it still enjoys the ``variance preserving'' property of the VPSDE.
In App.~\ref{app:var}, we show that the VESDE does not come with a reduced variance for $t$-sampling by default.

\textbf{(2)} \textit{Importance sampling (IS):}
We can keep $\beta(t)$ and $\vt$ unchanged for the original linear VPSDE, and instead use IS  to minimize variance. The theory of IS shows that the proposal $r(t) \propto {\D\,\log \vt}/{\D{t}}$ has minimum variance~\cite{owenMCbook}. In App.~\ref{app:var}, we show that we can sample from $r(t)$ using inverse transform sampling $t = \text{var}^{-1}((\sigma_1^2)^\rho (\sigma_0^2)^{1 - \rho})$ where $\text{var}^{-1}$ is the inverse of $\vt$ and $\rho \sim \U[0, 1]$. 
This variance reduction technique is available for any VPSDE with arbitrary $\beta(t)$.

\begin{wrapfigure}{r}{0.3\textwidth}
\vspace{-0.65cm}
  \begin{center}
    \includegraphics[scale=0.5, trim={10px, 10px, 10px, 10px}, clip=True]{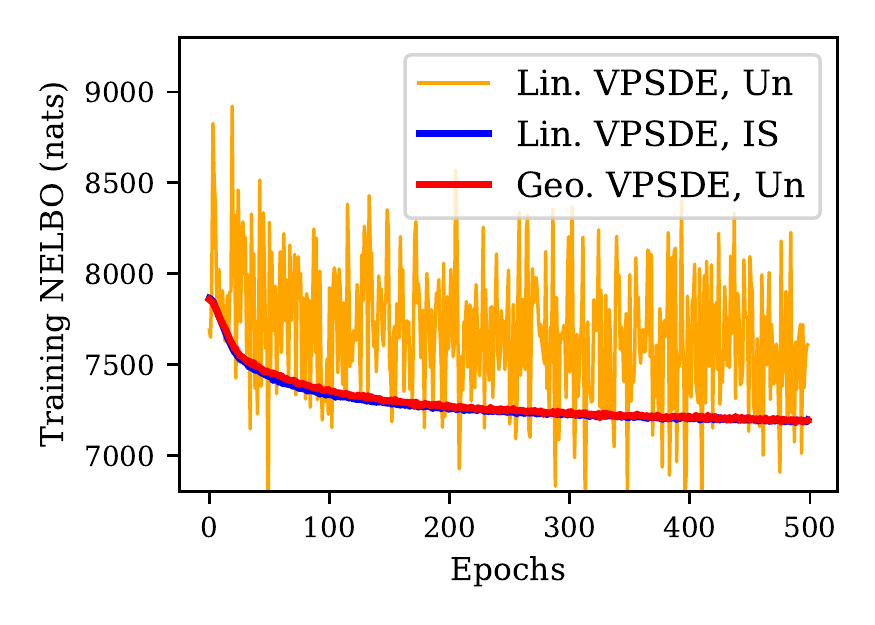}
  \end{center}
  \vspace{-0.3cm}
  \caption{\small Variance reduction}
  \label{fig:variance}
  \vspace{0.0cm}
  \begin{center}
    \includegraphics[scale=0.59, trim={10px, 10px, 10px, 10px}, clip=True]{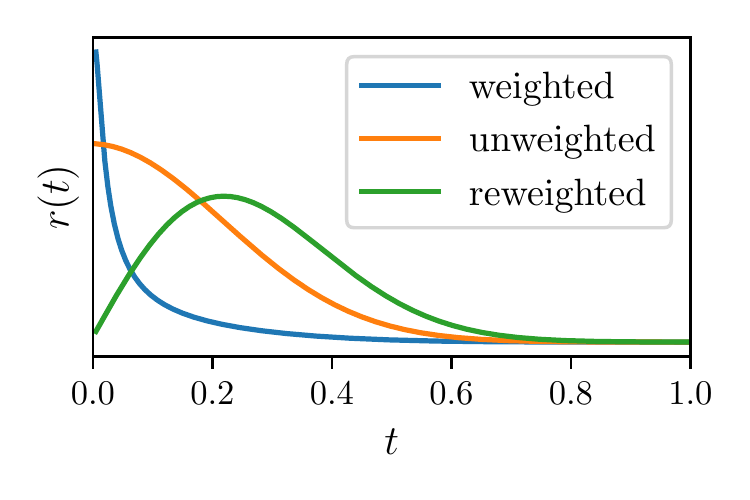}
  \end{center}
  \vspace{-0.3cm}
  \caption{\small IS distributions}
  \label{fig:is_dist}
  \vspace{-0.7cm}
\end{wrapfigure}

In Fig.~\ref{fig:variance}, we train a small LSGM on CIFAR-10 with $w_{\we}$ weighting using (i) the original VPSDE with uniform $t$ sampling, (ii) the same SDE but with our IS from $t$, and (iii) the proposed geometric VPSDE. Note how both (ii) and (iii) significantly reduce the variance and allow us to monitor the progress of the training objective. In this case, (i) has difficulty minimizing the objective due to the high variance. 
In App.~\ref{app:var}, we show how IS proposals can be formed for other SDEs, including the VESDE and Sub-VPSDE from \cite{song2021scoreSDE}.

\textbf{Variance reduction for unweighted and reweighted objectives:} When training with $w_{\un}$, analytically deriving IS proposal distributions for arbitrary $\beta(t)$ is challenging. For linear VPSDEs, we provide a derivation in App.~\ref{app:var} to obtain the optimal IS distribution. In contrast, defining IS proposal distributions is easier when training with $w_{\re}$. In App.~\ref{app:var}, we show that the optimal distribution is in the form $r(t) \propto {\D{\vt}}/{\D{t}}$  which is sampled by $t{=}\text{var}^{-1}((1 - \rho)\sigma^2_0 + \rho \sigma^2_1)$ with $\rho \sim \U[0, 1]$. In Fig.~\ref{fig:is_dist}, we visualize the IS distributions for the three weighting mechanisms for the linear VPSDE with the original $[\beta_0, \beta_1]$ parameters from~\cite{song2021scoreSDE}. $r(t)$ for the likelihood weighting is more tilted towards $t = 0$ due to the $1/\vt$ term in $w_{\we}$.

When using differently weighted objectives for training, we can either sample separate $t$ with different IS distributions for each objective, or use IS for the SGM objective (Eq.~\ref{eq:prior_obj}) and reweight the samples according to the likelihood objective for encoder training (Eq.~\ref{eq:vae_obj}). See App.~\ref{app:impl} for details.

\vspace{-0.3cm}
\section{Related Work}
\vspace{-0.21cm}
Our work builds on score-matching~\cite{hyvarinen2005scorematching,lyu2009scorematching,kingma2010scorematching,bengio2013denoising,geras2015scheduled,saremi2018deen,song2019slicedscorematching,li2019ebmsmhybrid,pang2020efficient},
specifically denoising score matching~\cite{vincent2011connection},
which makes our work related to recent generative models using denoising score matching- and denoising diffusion-based objectives~\cite{song2019scorematching,song2020improved,ho2020denoising,song2021scoreSDE,dhariwal2021diffusion}. Among those, \cite{ho2020denoising,dhariwal2021diffusion} use a discretized diffusion process with many noise scales, building on~\cite{sohldickstein2015thermodynamics}, while Song et al.~\cite{song2021scoreSDE} introduce the continuous time framework using SDEs. Experimentally, these works focus on image modeling and, contrary to us, work directly in pixel space.
Various works recently tried to address the slow sampling of 
these types of models and further improve output quality. \cite{jolicoeur-martineau2021DSM-ALS} add an adversarial objective,
\cite{song2021denoising} introduce non-Markovian diffusion processes that allow to trade off synthesis speed, quality, and sample diversity, \cite{gao2021recovery} learn a sequence of conditional energy-based models for denoising, \cite{luhman2021knowledge} distill the iterative sampling process into single shot synthesis, and \cite{sanroman2021noise} learn an adaptive noise schedule, which is adjusted during synthesis to accelerate sampling. Further, \cite{nichol2021improved} propose empirical variance reduction techniques for discretized diffusions and introduce a new, heuristically motivated, noise schedule. In contrast, our proposed noise schedule and our variance reduction techniques are analytically derived and directly tailored to our learning setting in the continuous time setup. 

Recently,~\cite{niu2020scorebasedgraphs} presented a method to generate graphs using score-based models, relaxing the entries of adjacency matrices to continuous values. 
LSGM would allow to model graph data more naturally using encoders and decoders tailored to graphs~\cite{simonovsky2018graphvae,jin2018jtvae,grover2019graphite,liao2019graphgeneration}.

Since our model can be considered a VAE~\cite{kingma2014vae,rezende2014stochastic} with score-based prior, it is related to approaches that improve VAE priors. For example, Normalizing flows and hierarchical distributions~\cite{rezendeICML15Normalizing,kingma2016improved,chen2016lossy,maaloe2019biva,vahdat2020nvae,child2021VDVAE}, as well as energy-based models~\cite{rolfe2016discrete,vahdat2018dvaes,Vahdat2018DVAE++,vahdat2019UndirectedPost,pang2020learning} have been proposed as VAE priors. Furthermore, classifiers~\cite{engel2018latent,bauer2019resampled,aneja2020ncpvae}, adversarial methods~\cite{makhzani2016adversarial}, and other techniques~\cite{takahashi2019variational,tomczak2018VampPrior} have been used to define prior distributions implicitly. In two-stage training, a separate generative model is trained in latent space as a new prior after training the VAE itself~\cite{dai2018diagnosing,oord2018neural,razavi2019generating,Ghosh2020From,esser2020vqgan,mittal2021symbolic}. Our work also bears a resemblance to recent methods on improving the sampling quality in generative adversarial networks using gradient flows in the latent space~\cite{ansari2021refining, che2020GANisEBM, Tanaka2019DOT, nie2021lace}, with the main difference that these prior works use a discriminator to update the latent variables, whereas we train an SGM.

\textbf{Concurrent works:} \cite{mittal2021symbolic} proposed to learn a denoising diffusion model in the latent space of a VAE for symbolic music generation.
This work does not introduce an end-to-end training framework of the combined VAE and denoising diffusion model and instead trains them in two separate stages. In contrast, concurrently with us \cite{wehenkel2021diffusion} proposed an end-to-end training approach, and \cite{sinha2021d2c} combines contrastive learning with diffusion models in the latent space of VAEs for controllable generation. However, \cite{mittal2021symbolic, wehenkel2021diffusion, sinha2021d2c} consider the discretized diffusion objective~\cite{ho2020denoising}, while we build on the continuous time framework. Also, these models are not equipped with the mixed score parameterization and variance reduction techniques, which we found crucial for the successful training of SGM priors. 

Additionally, \cite{kingma2021variational, song2021MLSGM, huang2021variational} concurrently with us proposed likelihood-based training of SGMs in data space\footnote{We build on the \href{https://arxiv.org/abs/2101.09258v1}{V1} version of \cite{song2021MLSGM}, which was substantially updated after the NeurIPS submission deadline.}. \cite{song2021MLSGM} developed a bound for the data likelihood in their Theorem 3 of their second version, using a denoising score matching objective, closely related to our cross entropy expression. However, our cross entropy expression is much simpler as we show how several terms can be marginalized out analytically for the diffusion SDEs employed by us (see our proof in App.~\ref{app:proof}). The same marginalization can be applied to Theorem 3 in \cite{song2021MLSGM} when the drift coefficient takes a special affine form (i.e., $\bff(\bz, t)=f(t) \bz$).
Moreover, \cite{huang2021variational} discusses the likelihood-based training of SGMs from a fundamental perspective and shows how several score matching objectives become a variational bound on the data likelihood. 
\cite{kingma2021variational} introduced a notion of signal-to-noise ratio (SNR) that results in a noise-invariant parameterization of time that depends only on the initial and final noise. Interestingly, our importance sampling distribution in Sec.~\ref{sec:variance} has a similar noise-invariant parameterization of time via $t = \text{var}^{-1}((\sigma_1^2)^\rho (\sigma_0^2)^{1 - \rho})$, which also depends only on the initial and final diffusion process variances. We additionally show that this time parameterization results in the optimal minimum-variance objective, if the distribution of latent variables follows a standard Normal distribution. Finally, \cite{kim2021score} proposed a modified time parameterization that allows modeling unbounded data scores. 
\vspace{-0.3cm}
\section{Experiments}
\vspace{-0.21cm}
Here, we examine the efficacy of LSGM in learning generative models for images.

\textbf{Implementation details:} We implement LSGM using the NVAE~\cite{vahdat2020nvae} architecture as VAE backbone and NCSN++~\cite{song2021scoreSDE} as SGM backbone. NVAE has a hierarchical latent structure. The diffusion process input
$\bz_0$ is constructed by concatenating the latent variables from all groups in the channel dimension. For NVAEs with multiple spatial resolutions in latent groups, we only feed the smallest resolution groups to the SGM prior and assume that the remaining groups have a standard Normal distribution. 

\textbf{Sampling:} To generate samples from LSGM at test time, we use a black-box ODE solver~\cite{chen2018neuralODE} to sample from the prior. Prior samples are then passed to the decoder to generate samples in data space. 

\textbf{Evaluation:} We measure NELBO, an upper bound on negative log-likelihood (NLL), using Eq.~\ref{eq:elbo_ce}. For estimating $\log p(\bz_0)$, we rely on the probability flow ODE~\cite{song2021scoreSDE}, which provides an unbiased but stochastic estimation of $\log p(\bz_0)$.
This stochasticity prevents us from performing an importance weighted estimation of NLL~\cite{burda2015importance} (see App.~\ref{app:bias} for details). For measuring sample quality, Fréchet inception distance (FID)~\cite{heusel2017gans} is evaluated with 50K samples. Implementation details in App.~\ref{app:impl}.

\begin{table*}
\small
\resizebox{0.45\linewidth}{!}{
\setlength{\tabcolsep}{2pt}
\begin{minipage}{0.48\textwidth}
\centering
\caption{\small Generative performance on CIFAR-10.} \label{table:main_cifar10}
    \begin{tabular}{llcc}
        \toprule
        & {\bf Method} & {\bf NLL$\downarrow$} & {\bf FID$\downarrow$} \\
        \midrule
        \multirow{3}{*}{\textbf{Ours}} & LSGM (FID) &  \hspace{-2.5mm}$\leq$3.43 & \bf 2.10  \\
        & LSGM (NLL) & \bf \hspace{-2mm}$\leq$2.87 & 6.89  \\
        & LSGM (balanced) & \hspace{-2mm}$\leq$2.95 & 2.17 \\
        & VAE Backbone  & 2.96 & 43.18 \\
        \midrule
        \multirow{6}{*}{\textbf{VAEs}} & VDVAE~\cite{child2021VDVAE} & \bf 2.87 & - \\
        & NVAE~\cite{vahdat2020nvae}   & 2.91 & 23.49 \\ 
        & VAEBM~\cite{xiao2021vaebm}  & - & 12.19 \\
        & NCP-VAE~\cite{aneja2020ncpvae} & - & 24.08 \\
        & BIVA~\cite{maaloe2019biva} & 3.08 & - \\
        & DC-VAE~\cite{parmar2020DC-VAE} & - & 17.90 \\
        \midrule
        \multirow{6}{*}{\textbf{Score}} & NCSN~\cite{song2019scorematching}   & - & 25.32\\
        & Rec. Likelihood~\cite{gao2021recovery} & 3.18 & 9.36 \\
        & DSM-ALS~\cite{jolicoeur-martineau2021DSM-ALS}  & 3.65 & - \\
        & DDPM~\cite{ho2020denoising} & 3.75 & 3.17 \\
        & Improved DDPM~\cite{nichol2021improved} & 2.94 & 11.47 \\
        & SDE (DDPM++)~\cite{song2021scoreSDE}  & 2.99 & 2.92 \\
        & SDE (NCSN++)~\cite{song2021scoreSDE}  & - & 2.20 \\
        \midrule
        \midrule
        \multirow{2}{*}{\textbf{Flows}} & VFlow~\cite{chen2020vflow} & 2.98 & - \\
        & ANF~\cite{huang2020anf} &  3.05 & - \\
        \midrule
        \multirow{5}{*}{\textbf{Aut.~Reg.}} & DistAug aug~\cite{jun2020distAug}  & 2.53 & 42.90 \\
        & Sp. Transformers~\cite{child2019sparse} & 2.80 & - \\
        &  $\delta$-VAE~\cite{razavi2019collapse} & 2.83 & - \\
        & PixelSNAIL~\cite{chen2018pixelsnail} & 2.85 & - \\
        & PixelCNN++~\cite{salimans2017pixelcnn++} & 2.92 & - \\
        \midrule 
        \multirow{2}{*}{\textbf{GANs}} & AutoGAN~\cite{Gong2019AutoGAN} & - & 12.42 \\
        & StyleGAN2-ADA~\cite{Karras2020ada}  & - & 2.92 \\
        \bottomrule
    \end{tabular}%
    \vspace{-0.2cm}
\end{minipage}
}
\hfill
\resizebox{0.45\linewidth}{!}{
\begin{minipage}{0.48\textwidth}
    \caption{\small Generative results on CelebA-HQ-256.}
    \centering
    \label{table:main_celeba}
    \begin{tabular}{llcc}
        \toprule
        & {\bf Method} & \bf NLL$\downarrow$ & {\bf FID$\downarrow$} \\
        \midrule
        \multirow{2}{*}{\textbf{Ours}}
        & LSGM  & \bf \hspace{-2mm}$\leq$0.70 & \bf 7.22 \\
        & VAE Backbone  & \bf 0.70 & 30.87 \\
        \midrule 
        \multirow{4}{*}{\textbf{VAEs}} & NVAE~\cite{vahdat2020nvae}   & \bf 0.70 & 29.76 \\ 
        & VAEBM~\cite{xiao2021vaebm}  & - & 20.38 \\
        & NCP-VAE~\cite{aneja2020ncpvae} & - & 24.79 \\
        & DC-VAE~\cite{parmar2020DC-VAE} & - & 15.80 \\
        \midrule 
        \multirow{1}{*}{\textbf{Score}}  &  SDE~\cite{song2021scoreSDE} & - & \bf 7.23 \\
        \midrule
        \midrule
        \multirow{1}{*}{\textbf{Flows}} & GLOW~\cite{kingma2018glow} & 1.03 & 68.93 \\
        \midrule 
        \multirow{1}{*}{\textbf{Aut.~Reg.}}  & SPN~\cite{menick2018spn} & 0.61 & - \\
        \midrule
        \multirow{3}{*}{\textbf{GANs}}  & Adv. LAE~\cite{pidhorskyi2020advLAE}   & - & 19.21 \\
        & VQ-GAN~\cite{esser2020vqgan}      & - & 10.70 \\
        & PGGAN~\cite{karras2018progressive}   & - & 8.03  \\
        \bottomrule
    \end{tabular}%
    
\bigskip
    \includegraphics[scale=0.5, trim={10px, 12px, 10px, 10px}, clip=True]{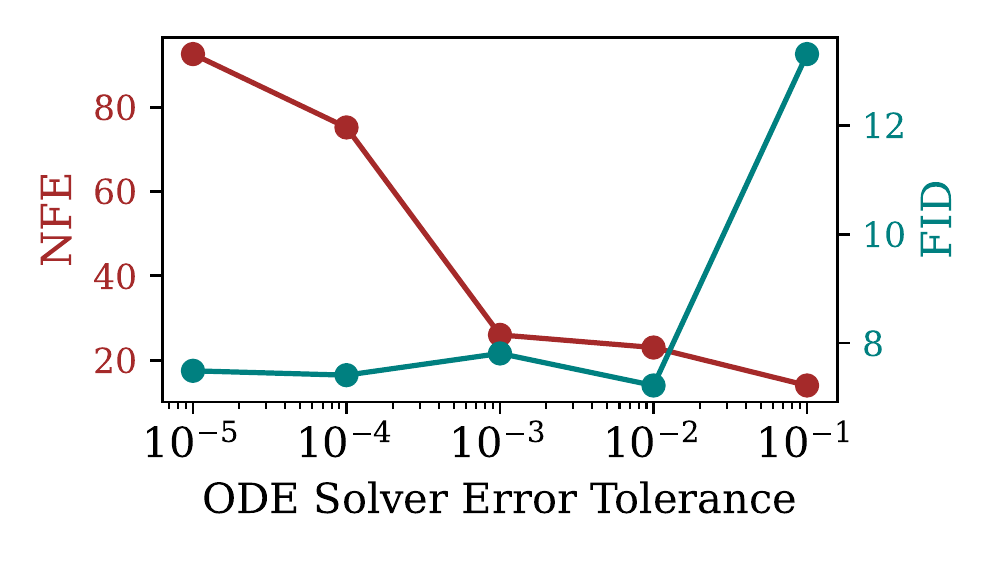}
  \captionof{figure}{\small FID and number of function evaluations (NFEs) for different ODE solver error tolerances on CelebA-HQ-256. LSGM takes 4.15 sec. for sampling while the original SGM~\cite{song2021scoreSDE} takes 45 min. with PC and 3.9 min. with ODE-based sampling.}
  \vspace{-0.3cm}
  \label{fig:run_time}
\end{minipage}
}
\end{table*}

\vspace{-0.2cm}
\subsection{Main Results} \label{sec:exp_main}
\textbf{Unconditional color image generation:}
Here, we present our main results for unconditional image generation on CIFAR-10~\cite{krizhevsky2009cifar} (Tab.~\ref{table:main_cifar10}) and CelebA-HQ-256 (5-bit quantized)~\cite{karras2018progressive} (Tab.~\ref{table:main_celeba}).
For CIFAR-10, we train 3 different models: \textit{LSGM (FID)} and \textit{LSGM (balanced)} both use the VPSDE with linear $\beta(t)$ and $w_{\un}$-weighting for the SGM prior in Eq.~\ref{eq:prior_obj}, while performing IS as derived in Sec.~\ref{sec:variance}.
They only differ in how the backbone VAE is trained.
\textit{LSGM (NLL)} is a model that is trained with our novel geometric VPSDE, using $w_{\we}$-weighting in the prior objective (further details in App.~\ref{app:impl}). 
When set up for high image quality, LSGM achieves a new state-of-the-art FID of 2.10. When tuned towards NLL, we achieve a NELBO of $2.87$, which is significantly better than previous score-based models. Only autoregressive models, which come with very slow synthesis, and VDVAE~\cite{child2021VDVAE} reach similar or higher likelihoods, but they usually have much poorer image quality.

For CelebA-HQ-256, we observe that when LSGM is trained with different SDE types and weighting mechanisms, it often obtains similar NELBO potentially due to applying the SGM prior only to small latent variable groups and using Normal priors at the larger groups. With $w_{\re}$-weighting and linear VPSDE, LSGM obtains the state-of-the-art FID score of 7.22 on a par with the original SGM~\cite{song2021scoreSDE}. 

For both datasets, we also report results for the VAE backbone used in our LSGM. Although this baseline achieves competitive NLL, its sample quality is behind our LSGM and the original SGM.

\begin{table*}
\small

\setlength{\tabcolsep}{4pt}
\resizebox{0.42\linewidth}{!}{
\begin{minipage}{0.47\textwidth}
    \caption{\small Dyn. binarized OMNIGLOT results.}
    \centering
    \label{table:omniglot}
        \begin{tabular}{llcc}
        \toprule
        & {\bf Method} & {\bf NELBO$\downarrow$} &  {\bf NLL$\downarrow$} \\
        \midrule
       \multirow{1}{*}{\textbf{Ours}} & LSGM & \bf 87.79 & \bf \hspace{-0.19cm}$\leq$87.79 \\
        \midrule
        \multirow{4}{*}{\textbf{VAEs}} 
        & NVAE~\cite{vahdat2020nvae}    & 93.92 & 90.75 \\
        & BIVA~\cite{maaloe2019biva}    & 93.54 & 91.34 \\
        & DVAE++~\cite{Vahdat2018DVAE++} &  -    & 92.38 \\
        & Ladder VAE~\cite{sonderby2016ladder} & - & 102.11 \\
        \midrule
        \midrule
        \multirow{3}{*}{\textbf{Aut.~Reg.}}
        & VLVAE~\cite{chen2016lossy} & - & 89.83 \\
        & VampPrior~\cite{tomczak2018VampPrior}  & -  & 89.76 \\
        & PixelVAE++~\cite{sadeghi2019pixelvae++} & -  & 88.29 \\ 
        \bottomrule
    \end{tabular}%
    \vspace{-0.4cm}
\end{minipage}
}
\hfill
\resizebox{0.42\linewidth}{!}{
\begin{minipage}{0.47\textwidth}
\centering
\caption{\small Dynamically binarized MNIST results.}
\label{table:mnist}
    \begin{tabular}{llcc}
        \toprule
        & {\bf Method} & {\bf NELBO$\downarrow$} &  {\bf NLL$\downarrow$} \\
        \midrule
       \multirow{1}{*}{\textbf{Ours}} & LSGM & \bf 78.47 & \hspace{-0.19cm}$\leq$78.47 \\
        \midrule
        \multirow{4}{*}{\textbf{VAEs}}
        & NVAE~\cite{vahdat2020nvae} &  79.56 & \bf 78.01 \\
        & BIVA~\cite{maaloe2019biva} & 80.06 & 78.41 \\
        & IAF-VAE~\cite{kingma2016improved} & 80.80  & 79.10 \\
        & DVAE++~\cite{Vahdat2018DVAE++}    & -   & 78.49 \\
        \midrule
        \midrule
        \multirow{3}{*}{\textbf{Aut.~Reg.}} & PixelVAE++~\cite{sadeghi2019pixelvae++} & -  & 78.00 \\
        & VampPrior~\cite{tomczak2018VampPrior} & -  & 78.45 \\
        & MAE~\cite{ma2019mae}                  & -  & 77.98 \\
        \bottomrule
    \end{tabular}%
    \vspace{-0.4cm}
\end{minipage}
}
\end{table*}

\textbf{Modeling binarized images:} 
Next, we examine LSGM on dynamically binarized MNIST~\cite{lecun1998mnist} and OMNIGLOT~\cite{burda2015importance}. We apply LSGM to binary images 
using a decoder with pixel-wise 
independent Bernoulli distributions.
For these datasets, we report both NELBO and NLL in nats in Tab.~\ref{table:omniglot} and Tab.~\ref{table:mnist}. On OMNIGLOT, LSGM achieves state-of-the-art likelihood of $\leq$87.79 nat, outperforming previous models including VAEs with autoregressive decoders, and even when comparing its NELBO against importance weighted estimation of NLL for other methods. On MNIST, LSGM outperforms previous VAEs in NELBO, reaching a NELBO 1.09 nat lower than the state-of-the-art NVAE.

\begin{figure}[!t]
     \centering
     \hspace{-0.2cm}
     \begin{subfigure}[b]{0.25\textwidth}
         \centering
         \includegraphics[scale=0.46, trim={0px, 64px, 64px, 0px}, clip=True]{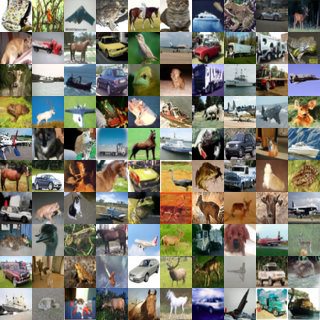}
         \caption{CIFAR-10}
     \end{subfigure}
     \hfill
     \begin{subfigure}[b]{0.45\textwidth}
         \centering
         \includegraphics[scale=0.23, trim={0px, 1px, 0px, 0px}, clip=True]{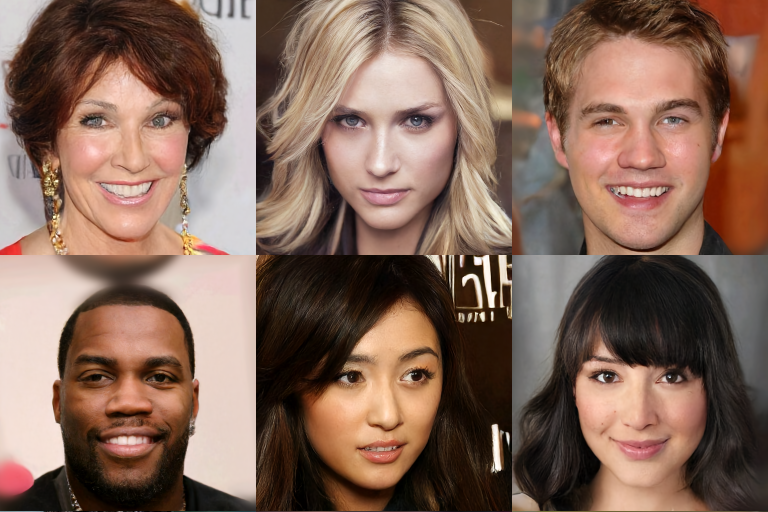}
         \caption{CelebA-HQ-256}
     \end{subfigure}
     \begin{minipage}[b]{0.25\textwidth}
     \begin{subfigure}[b]{\textwidth}
         \centering
         \includegraphics[scale=0.37, trim={0px, 0px, 0px, 0px}, clip=True]{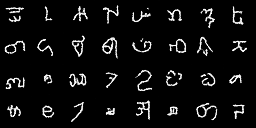}
         \caption{OMNIGLOT}
         \vspace{7px}
     \end{subfigure} \\ 
     \begin{subfigure}[b]{\textwidth}
         \centering
         \includegraphics[scale=0.37, trim={0px, 0px, 0px, 0px}, clip=True]{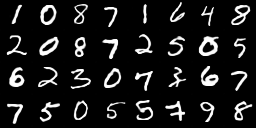}
         \caption{MNIST}
     \end{subfigure}
     \end{minipage}
    \caption{\small Generated samples for different datasets. For binary datasets, we visualize the decoder mean. 
    LSGM successfully generates sharp, high-quality, and diverse samples (additional samples in appendix).}
    \label{fig:qualitative}
    \vspace{-0.3cm}
\end{figure}

\textbf{Qualitative results}: 
We visualize qualitative results for all datasets in Fig.~\ref{fig:qualitative}. On the complex multimodal CIFAR-10 dataset, LSGM generates sharp and high-quality images. On CelebA-HQ-256, LSGM generates diverse samples from different ethnicity and age groups with varying head poses and facial expressions. On MNIST and OMNIGLOT, the generated characters are sharp and high-contrast.

\textbf{Sampling time}:
We compare LSGM against the original SGM~\cite{song2021scoreSDE} trained on the CelebA-HQ-256 dataset in terms of sampling time and number of function evaluations (NFEs) of the ODE solver.
Song et al.~\cite{song2021scoreSDE} propose two main sampling techniques including predictor-corrector (PC) and probability flow ODE. PC sampling involves 4000 NFEs and takes 44.6 min. on a Titan V for a batch of 16 images. It yields 7.23 FID score 
(see Tab.~\ref{table:main_celeba}). ODE-based sampling from SGM takes 3.91 min. with 335 NFEs, but it obtains a poor FID score of 128.13 with $10^{-5}$ as ODE solver error tolerance\footnote{We use the VESDE checkpoint at \url{https://github.com/yang-song/score_sde_pytorch}. Song et al.~\cite{song2021scoreSDE} report that ODE-based sampling yields worse FID scores for their models (see D.4 in \cite{song2021scoreSDE}). The problem is more severe for VESDEs. Unfortunately, at submission time only a VESDE model was released.}.

In a stark contrast, ODE-based sampling from our LSGM takes 0.07 min. with average of 23 NFEs, yielding 7.22 FID score. LSGM is 637$\times$ and 56$\times$ faster than original SGM's~\cite{song2021scoreSDE} PC and ODE sampling, respectively. 
In Fig.~\ref{fig:run_time}, we visualize FID scores and NFEs for different ODE solver error tolerances. Our LSGM achieves low FID scores for relatively large error tolerances.

We identify three main reasons for this significantly faster sampling from LSGM:
(i) The SGM prior in our LSGM models latent variables with 32$\times$32 spatial dim., whereas the original SGM~\cite{song2021scoreSDE} directly models 256$\times$256 images. 
The larger spatial dimensions require a deeper network to achieve a large receptive field. (ii) Inspecting the SGM prior in our model suggests that the score function is heavily dominated by the linear term at the end of training, as the mixing coefficients $\balpha$ are all $<0.02$. This makes our SGM prior smooth and numerically faster to solve. 
(iii) Since SGM is formed in the latent space in our model, 
errors from solving the ODE can be corrected to some degree using the VAE decoder, while in the original SGM~\cite{song2021scoreSDE} errors directly translate to artifacts in pixel space.

\vspace{-0.2cm}
\subsection{Ablation Studies} \label{sec:exp_ab}
\textbf{SDEs, objective weighting mechanisms and variance reduction.} In Tab.~\ref{tab:ablation_obj_weighting}, we analyze the different weighting mechanisms and variance reduction techniques and compare the geometric VPSDE with the regular VPSDE with linear $\beta(t)$~\cite{ho2020denoising,song2021scoreSDE}.
In the table, \textit{SGM-obj.-weighting} denotes the weighting mechanism used when training the SGM prior (via Eq.~\ref{eq:prior_obj}). 
$t$\textit{-sampling (SGM-obj.)} indicates the sampling approach for $t$, where $r_{\we}(t)$, $r_{\un}(t)$ and $r_{\re}(t)$ denote the IS distributions for the weighted (likelihood), the unweighted, and the reweighted objective, respectively.
For training the VAE encoder $q_\phi(\bz_0|\bx)$ (last term in Eq.~\ref{eq:vae_obj}), we either sample a separate batch $t$ with importance sampling following $r_{\we}(t)$ (only necessary when the SGM prior is not trained with $w_{\we}$ itself), or we \textit{reweight} the samples drawn for training the prior according to the likelihood objective (denoted by \textit{rew.}). \textit{n/a} indicates fields that do not apply: The geometric VPSDE has optimal variance for the weighted (likelihood) objective already with uniform sampling; there is no additional IS distribution. Also, we did not derive IS distributions for the geometric VPSDE for $w_{\un}$. \textit{NaN} indicates experiments that failed due to training instabilities.
Previous work~\cite{vahdat2020nvae, child2021VDVAE} have reported instability in training large VAEs. We find that our method inherits similar instabilities from VAEs; however, importance sampling often stabilizes training our LSGM. As expected, we obtain the best NELBOs (red) when training with the weighted, maximum likelihood objective ($w_{\we}$). Importantly, our new geometric VPSDE achieves the best NELBO. Furthermore, the best FIDs (blue) are obtained either by unweighted ($w_{\un}$) or reweighted ($w_{\re}$) SGM prior training, with only slightly worse NELBOs. These experiments were run on the CIFAR10 dataset, using a smaller model than for our main results above (details in App.~\ref{app:impl}).

\begin{table*}
\small
\setlength{\tabcolsep}{4pt}
\caption{\small Ablations on SDEs, objectives, weighting mechanisms, and variance reduction. Details in App.~\ref{app:impl}.}
\centering
\resizebox{0.9\linewidth}{!}{
    \begin{tabular}{cc|cc|cc|cc|cc|cc}
    \toprule
    \multicolumn{2}{c|}{\textbf{SGM-obj.-weighting}} & \multicolumn{2}{c|}{$w_{\we}$} & \multicolumn{4}{c|}{$w_{\un}$} & \multicolumn{4}{c}{$w_{\re}$} \\
    \midrule
    \multicolumn{2}{c|}{$t$\textbf{-sampling (SGM-obj.)}} & $\U[0, 1]$ &  $ r_{\we}(t)$ & \multicolumn{2}{c|}{$\U[0, 1]$} & \multicolumn{2}{c|}{$ r_{\un}(t)$} & \multicolumn{2}{c|}{$\U[0, 1]$} & \multicolumn{2}{c}{$ r_{\re}(t)$} \\
    \midrule
    \multicolumn{2}{c|}{$t$\textbf{-sampling (q-obj.)}} & \textit{rew.} &  \textit{rew.} & \textit{rew.} &  $ r_{\we}(t)$ & \textit{rew.} &  $ r_{\we}(t)$ & \textit{rew.} &  $ r_{\we}(t)$ & \textit{rew.} &  $ r_{\we}(t)$ \\
    \midrule\midrule
    \textbf{Geom.-} & \textbf{FID}$\downarrow$    & 10.18 & {\color{gray}\textit{n/a}} & \textit{NaN} & \textit{NaN} & {\color{gray}\textit{n/a}} & {\color{gray}\textit{n/a}} & 22.21 & \textit{NaN} & 7.29 & 7.18 \\
    \textbf{VPSDE} & \textbf{NELBO}$\downarrow$    & {\color{red}2.96} & {\color{gray}\textit{n/a}} & \textit{NaN} & \textit{NaN} & {\color{gray}\textit{n/a}} & {\color{gray}\textit{n/a}} & 3.04 & \textit{NaN} & 2.99 & 2.99 \\
    \midrule
    \multirow{2}{*}{\textbf{VPSDE}}
    & \textbf{FID}$\downarrow$    & 6.15 & 8.00 & \textit{NaN} & \textit{NaN} & {\color{blue}5.39} & {\color{blue}5.39} & \textit{NaN} & {\color{blue}4.99} & 15.12 & 6.19 \\
    & \textbf{NELBO}$\downarrow$    & {\color{red}2.97} & {\color{red}2.97} & \textit{NaN} & \textit{NaN} & 2.98 & 2.98 & \textit{NaN} & 2.99 & 3.03 & 2.99 \\
    \bottomrule
\end{tabular}%
}
\label{tab:ablation_obj_weighting}
\vspace{-0.4cm}
\end{table*}

\textbf{End-to-end training.} We proposed to train LSGM end-to-end, in contrast to~\cite{mittal2021symbolic}. Using a similar setup as above we compare end-to-end training of LSGM during the second stage with freezing the VAE encoder and decoder and only training the SGM prior in latent space during the second stage. When training the model end-to-end, we achieve an FID of $5.19$ and NELBO of $2.98$; when freezing the VAE networks during the second stage, we only get an FID of $9.00$ and NELBO of $3.03$. These results clearly motivate our end-to-end training strategy.

\textbf{Mixing Normal and neural score functions.} We generally found training LSGM without our proposed ``mixed score'' formulation (Sec.~\ref{sec:param})
to be unstable during end-to-end training, highlighting its importance.
To quantify the contribution of the mixed score parametrization for a stable model, we train a small LSGM with only one latent variable group.
In this case, without the mixed score, we reached an FID of $34.71$ and NELBO of $3.39$; with it, we got an FID of $7.60$ and NELBO of $3.29$. 
Without the inductive bias provided by the mixed score, learning that the marginal distribution is close to a Normal one for large $t$ purely from samples can be very hard in the high-dimensional latent space, where our diffusion is run. Furthermore, due to our importance sampling schemes, we tend to oversample small, rather than large $t$. However, synthesizing high-quality images requires an accurate score function estimate for all $t$. On the other hand, the log-likelihood of samples is highly sensitive to local image statistics and primarily determined at small $t$. It is plausible that we are still able to learn a reasonable estimate of the score function for these small $t$ even without the mixed score formulation. That may explain why log-likelihood suffers much less than sample quality, as estimated by FID, when we remove the mixed score parameterization.

Additional experiments and model samples are presented in App.~\ref{app:expr}.

\vspace{-0.2cm}
\section{Conclusions} \label{sec:conclusions}
\vspace{-0.2cm}
We proposed the \textit{Latent Score-based Generative Model}, a novel framework for end-to-end training of score-based generative models in the latent space of a variational autoencoder. Moving from data to latent space allows us to form more expressive generative models, model non-continuous data, and reduce sampling time using smoother SGMs. To enable training latent SGMs, we made three core contributions: (i) we derived a simple expression for the cross entropy term in the variational objective, (ii) we parameterized the SGM prior by mixing Normal and neural score functions, and (iii) we proposed several techniques for variance reduction in the estimation of the training objective. Experimental results show that latent SGMs outperform recent pixel-space SGMs in terms of both data likelihood and sample quality, and they can also be applied to binary datasets. In large image generation, LSGM generates data several orders of magnitude faster than recent SGMs. Nevertheless, LSGM's synthesis speed does not yet permit sampling at interactive rates, and our implementation of LSGM is currently limited to image generation. Therefore, future work includes further accelerating sampling, applying LSGMs to other data types, and designing efficient networks for LSGMs.
\vspace{-0.2cm}
\section{Broader Impact} \label{sec:impact}
\vspace{-0.2cm}
Generating high-quality samples while fully covering the data distribution has been a long-standing challenge in generative learning. A solution to this problem will likely help reduce biases in generative models and lead to improving overall representation of minorities in the data distribution. SGMs are perhaps one of the first deep models that excel at both sample quality and distribution coverage. However, the high computational cost of sampling limits their widespread use. Our proposed LSGM reduces the sampling complexity of SGMs by a large margin and improves their expressivity further. Thus, in the long term, it can enable the usage of SGMs in practical applications.  

Here, LSGM is examined on the image generation task which has potential benefits and risks discussed in \cite{Bailey2020thetools, vaccari2020deepfakes}. However, LSGM can be considered a generic framework that extends SGMs to non-continuous data types. In principle LSGM could be used to model, for example, language~\cite{bowman2016generating,li2020_Optimus}, music~\cite{dhariwal2020jukebox,mittal2021symbolic}, or molecules~\cite{sanchezlengeling2018inverse,alperstein2019smiles}. Furthermore, like other deep generative models, it can potentially be used also for non-generative tasks such as semi-supervised and representation learning~\cite{kingma2014semi,odena2016semisupervised,li2021semantic}. This makes the long-term social impacts of LSGM dependent on the downstream applications.

\vspace{-0.2cm}
\section*{Funding Statement}
\vspace{-0.2cm}
All authors were funded by NVIDIA through full-time employment. 

\bibliographystyle{unsrt}
{\small
\bibliography{bibliography}}

\clearpage

\tableofcontents
\clearpage

\appendix
\section{Proof for Theorem 1}\label{app:proof}
Without loss of generality, we state the theorem in general form without conditioning on $\bx$.

\setcounter{theorem}{0}
\begin{theorem} \label{th:simple_ce_app}
Given two distributions $q(\bz_0)$ and $p(\bz_0)$ defined in the continuous space $\R^D$, denote the marginal distributions of diffused samples under the SDE $\D{\bz} = f(t)\bz\D{t} + g(t)\D{\bw}$ at time $t\in[0,1]$ with $q(\bz_t)$ and $p(\bz_t)$.
Assuming that $\log q(\bz_t)$ and $\log p(\bz_t)$ are smooth with at most polynomial growth at $\bz_t\rightarrow\pm\infty$, and also assuming that $f(t)$ and $g(t)$ are chosen such that $q(\bz_1)=p(\bz_1)$ at $t=1$, the cross entropy is given by:
{\small
\begin{align*}
 \CE{q(\bz_0)}{p(\bz_0)} = \E_{t \sim \U[0, 1]}\left[\frac{g(t)^2}{2} \E_{q(\bz_t, \bz_0|\bx)} \left[||\nabla_{\bz_t} \log q(\bz_t|\bz_0)\!-\!\nabla_{\bz_t} \log p(\bz_t)||_2^2 \right]\right]\!+\!\frac{D}{2} \log\left(2\pi e\vo\right),
\end{align*}}%
with $q(\bz_t, \bz_0) = q(\bz_t|\bz_0)q(\bz_0)$ and a Normal transition kernel $q(\bz_t | \bz_0) = \N(\bz_t; \bmu_t(\bz_0), \vt \beye)$  where $\bmu_t$ and $\vt$ are obtained from $f(t)$ and $g(t)$ for a fixed initial variance $\vo$ at $t=0$.
\end{theorem}

Theorem~\ref{th:simple_ce_app} amounts to estimating the cross entropy between $q(\bz_0)$ and $p(\bz_0)$ with denoising score matching and can be understood intuitively in the context of LSGM: We are drawing samples from a potentially complex encoding distribution $q(\bz_0)$, add Gaussian noise with small initial variance $\vo$ to obtain a well-defined initial distribution, and then smoothly perturb the sampled encodings using a diffusion process, while learning a denoising model, the SGM prior. Note that from the perspective of the learnt SGM prior, which is defined by the separate reverse-time generative SDE with the learnt score function model  (see Sec.~\ref{sec:background}), the expression in our theorem becomes an upper bound (see discussion in Sec.~\ref{sec:ce}).

\begin{proof}
The first part of our proof follows a similar proof strategy as was used by Song et al.~\cite{song2021MLSGM}. We start the proof with a more generic diffusion process in the form:
\begin{equation*}
    d\bz = \bff(\bz, t) dt + g(t) d\bw 
\end{equation*}
The time-evolution of probability densities $q(\bz_t)$ and $p(\bz_t)$ under this SDE is described by the Fokker-Planck equation~\cite{saerkka2019sdebook} (note that we follow the same notation as in the main paper: We omit the $t$-subscript of the diffused distributions $q_t$, indicating the time dependence at the variable, i.e. $q(\bz_t)\equiv q_t(\bz_t)$):
\begin{equation}
\begin{split}
    \frac{\partial q(\bz_t)}{\partial t} & = \nabla_{\bz_t}\left(\frac{1}{2}g^2(t)q(\bz_t)\nabla_{\bz_t}\log\,q(\bz_t) - \bff(\bz, t)q(\bz_t) \right) \\
    & = \nabla_{\bz_t}\left( \bh_q(\bz_t, t) q(\bz_t)\right)
\end{split}
\end{equation}
with
\begin{equation}
\bh_q(\bz_t, t):= \frac{1}{2}g^2(t)\nabla_{\bz_t}\log\,q(\bz_t) - \bff(\bz, t)
\end{equation}
and analogously for $p(\bz_t)$.

The cross entropy can be written as
\begin{align*}
\CE{q(\bz_0)}{p(\bz_0)} &=\CE{q(\bz_1)}{p(\bz_1)} + \int_{1}^0\frac{\partial}{\partial t}\CE{q(\bz_t)}{p(\bz_t)}dt \\
&=\H{q(\bz_1)} - \int_{0}^1\frac{\partial}{\partial t}\CE{q(\bz_t)}{p(\bz_t)}dt
\end{align*}
since $q(\bz_1)=p(\bz_1)$, as assumed in the Theorem (in practice, the used SDEs are designed such that $q(\bz_1)=p(\bz_1)$).

Furthermore, we have
\begin{align*}
\frac{\partial}{\partial t}\CE{q(\bz_t)}{p(\bz_t)}
&=-\int\left[\frac{\partial q(\bz_t)}{\partial t}\log p(\bz_t) + \frac{q(\bz_t)}{p(\bz_t)}\frac{\partial p(\bz_t)}{\partial t}\right]d\bz \\
&\overset{\mathrm{(i)}}{=}-\int\left[\nabla_{\bz_t}(\bh_q(\bz_t,t)q(\bz_t))\log p(\bz_t) + \frac{q(\bz_t)}{p(\bz_t)} \nabla_{\bz_t}(\bh_p(\bz_t,t)p(\bz_t)) \right]d\bz \\
&\overset{\mathrm{(ii)}}{=} \int\left[\bh_q(\bz_t,t)^\top q(\bz_t)\nabla_{\bz_t}\log p(\bz_t) + \bh_p(\bz_t,t)^\top p(\bz_t)\nabla_{\bz_t} \frac{q(\bz_t)}{p(\bz_t)} \right]d\bz \\
&\overset{\mathrm{(iii)}}{=}\int q(\bz_t)\bigl[\bh_q(\bz_t,t)^\top\nabla_{\bz_t}\log p(\bz_t)  \\ &\qquad\qquad\quad + \bh_p(\bz_t,t)^\top\nabla_{\bz_t}\log q(\bz_t) \\ &\qquad\qquad\quad - \bh_p(\bz_t,t)^\top\nabla_{\bz_t}\log p(\bz_t) \bigr]d\bz  \\
&\overset{\mathrm{(iv)}}{=} \int q(\bz_t)\biggl[ -\frac{1}{2}g^2(t)||\nabla_{\bz_t}\log p(\bz_t)||^2 -\bff(\bz_t,t)^\top \nabla_{\bz_t}\log q(\bz_t) \\ & \qquad\quad\qquad + g^2(t)\nabla_{\bz_t} \log q(\bz_t)^\top\nabla_{\bz_t} \log p(\bz_t) \biggr]d\bz
\end{align*}
where $(i)$ inserts the Fokker Planck equations for $q(\bz_t)$ and $p(\bz_t)$, respectively. 
Furthermore, $(ii)$ is integration by parts assuming similar limiting behavior of $q(\bz_t)$ and $p(\bz_t)$ at $\bz_t\rightarrow\pm\infty$ as Song et al.~\cite{song2021MLSGM}. Specifically, we know that $q(\bz_t)$ and $p(\bz_t)$ must decay towards zero at $\bz_t\rightarrow\pm\infty$ to be normalized. Furthermore, we assumed $\log q(\bz_t)$ and $\log p(\bz_t)$ to have at most polynomial growth (or decay, when looking at it from the other direction) at $\bz_t\rightarrow\pm\infty$, which implies faster exponential growth/decay of $q(\bz_t)$ and $p(\bz_t)$. Also, $\nabla_{\bz_t}\log q(\bz_t)$ and $\nabla_{\bz_t}\log p(\bz_t)$ grow/decay at most polynomially, too, since the gradient of a polynomial is still a polynomial. Hence, one can work out that all terms to be evaluated at $\bz_t\rightarrow\pm\infty$ after integration by parts vanish.
Finally, $(iii)$ uses the log derivative trick and some rearrangements, and $(iv)$ is obtained by inserting $\bh_q$ and $\bh_p$.

Hence, we obtain
\begin{align*}
\CE{q(\bz_0)}{p(\bz_0)}=\H{q(\bz_1)} + \int_{0}^1\E_{q(\bz_t)}\biggl[ &\frac{1}{2}g^2(t)||\nabla_{\bz_t}\log p(\bz_t)||_2^2 +\bff(\bz_t,t) \nabla_{\bz_t}\log q(\bz_t) \\ & - g^2(t)\nabla_{\bz_t} \log q(\bz_t)^\top\nabla_{\bz_t} \log p(\bz_t) \biggr]dt,
\end{align*}
which we can interpret as a general score matching-based expression for calculating the cross entropy, analogous to the expressions for the Kullback-Leibler divergence and entropy derived by Song et al.~\cite{song2021MLSGM}.

However, as discussed in the main paper, dealing with the marginal score $\nabla_{\bz_t}\log q(\bz_t)$ is problematic for complex ``input'' distributions $q(\bz_0)$. Hence, 
we further transform the cross entropy expression into a denoising score matching-based expression:

{\small
\begin{align*}
\hspace{-0.5cm}
\CE{q(\bz_0)}{p(\bz_0)} &= \H{q(\bz_1)} + \int_{0}^1\E_{q(\bz_t)}\biggl[\frac{1}{2}g^2(t)||\nabla_{\bz_t}\log p(\bz_t)||_2^2 +\bff(\bz_t,t) \nabla_{\bz_t}\log q(\bz_t) \\ & \quad - g^2(t)\nabla_{\bz_t} \log q(\bz_t)^\top\nabla_{\bz_t} \log p(\bz_t) \biggr]dt \\
  &\overset{\mathrm{(i)}}{=} \frac{1}{2} \int_0^1 g(t)^2 \E_{q(\bz_0,\bz_t)} \left[ - 2\nabla_\bz \log q(\bz_t|\bz_0)^\top\nabla_{\bz_t} \log p(\bz_t) + ||\nabla_{\bz_t} \log p(\bz_t)||_2^2 \right] dt \\
   & \quad + \frac{1}{2} \int_0^1 \E_{q(\bz_0,\bz_t)} \left[2 \bff(\bz, t)^\top \nabla_{\bz_t} \log q(\bz_t|\bz_0) \right] dt + \H{q(\bz_1)} \\
 &\overset{\mathrm{(ii)}}{=} \frac{1}{2} \int_0^1 g(t)^2 \E_{q(\bz_0, \bz_t)} \left[||\nabla_{\bz_t} \log q(\bz_t|\bz_0)||_2^2 - 2\nabla_{\bz_t} \log q(\bz_t|\bz_0)^\top\nabla_{\bz_t} \log p(\bz_t) + ||\nabla_{\bz_t} \log p(\bz_t)||_2^2 \right] dt \\
 & \quad + \frac{1}{2} \int_0^1 \E_{q(\bz_t)} \left[2 \bff(\bz, t)^\top \nabla_{\bz_t} \log q(\bz_t|\bz_0) - g(t)^2 ||\nabla_{\bz_t} \log q(\bz_t|\bz_0)||_2^2 \right] dt + \H{q(\bz_1)} \\
 &\overset{\mathrm{(iii)}}{=} \frac{1}{2} \int_0^1 g(t)^2 \E_{q(\bz_0, \bz_t)} \left[||\nabla_{\bz_t} \log q(\bz_t|\bz_0) - \nabla_{\bz_t} \log p(\bz_t)||_2^2 \right] dt \\
 & \quad + \underbrace{\frac{1}{2} \int_0^1 \E_{q(\bz_0, \bz_t)} \left[ \left(2\bff(\bz, t) - g(t)^2 \nabla_{\bz_t} \log q(\bz_t|\bz_0) \right)^\top \nabla_{\bz_t} \log q(\bz_t|\bz_0) \right] dt}_{\text{(I): Model-independent term}} + \H{q(\bz_1)} \\
\end{align*}}
with $q(\bz_0,\bz_t)=q(\bz_t|\bz_0)q(\bz_0)$ and where
in $(i)$ we have used the following identity from Vincent~\cite{vincent2011connection}: 
\begin{align*}
    \E_{q(\bz_t)} \left[\nabla_{\bz_t} \log q(\bz_t) \right] &= 
    \E_{q(\bz_t)} \left[ \E_{q(\bz_0|\bz_t)} \left[\nabla_{\bz_t} \log q(\bz_t |\bz_0)  \right]  \right] = \E_{q(\bz_0)q(\bz_t|\bz_0)} \left[\nabla_{\bz_t} \log q(\bz_t |\bz_0) \right].
\end{align*}
 In $(ii)$, we have added and subtracted $g(t)^2||\nabla_{\bz_t} \log q(\bz_t|\bz_0)||_2^2$ and in $(iii)$ we rearrange the terms into denoising score matching. In the following, we show that the term marked by (I) depends only on the diffusion parameters and does not depend on $q(\bz_0)$ when $\bff(\bz, t)$ takes a special affine (linear) form $\bff(\bz, t) := f(t) \bz$, which is often used for training SGMs and which we assume in our Theorem.

Note that for linear $\bff(\bz, t) := f(t) \bz$, we can derive the mean and variance (there are no ``off-diagonal'' co-variance terms here, since all dimensions undergo diffusion independently) of the distribution $q(\bz_t | \bz_0)$ at any time $t$ in closed form, essentially solving the Fokker-Planck equation for this special case analytically. In that case, if the initial distribution at $t=0$ is Normal then the distribution stays Normal and the mean and variance completely describe the distribution, i.e. $q(\bz_t | \bz_0) = \N(\bz_t; \bmu_t(\bz_0), \vt \beye)$. The mean and variance are given by the differential equations and their solutions~\cite{saerkka2019sdebook}:
\begin{align}
    \frac{d \bmu}{dt} = f(t) \bmu \quad \quad \quad \quad &\rightarrow \bmu_t = \bz_0 e^{\int_0^t f(s) ds} \label{eq:gen_mean_diff_app} \\
    \frac{d \sigma^2}{dt} = 2f(t) \sigma^2 + g^2(t) &\rightarrow \vt = \frac{1}{\tilde{F}(t)}\left( \int_0^t \tilde{F}(s) g^2(s)ds + \sigma_0^2 \right),\;\; \tilde{F}(t):=e^{-2\int_0^t f(s)ds} \label{eq:gen_var_diff_app}
\end{align}
Here, $\bz_0$ denotes the mean of the distribution at $t=0$ and $\sigma_0^2$ the component-wise variance at $t=0$. After transforming into the denoising score matching expression above, what we are doing is essentially drawing samples $\bz_0$ from the potentially complex $q(\bz_0)$, then placing simple Normal distributions with variance $\sigma_0^2$ at those samples, and then letting those distributions evolve according to the SDE. $\sigma_0^2$ acts as a hyperparameter of the model.

In this case, i.e. when the distribution $q(\bz_t | \bz_0)$ is Normal at all $t$, we can represent samples $\bz_t$ from the intermediate distributions in reparameterized from $\bz_t = \bmu_t(\bz_0) + \sigma_t \beps$ where $\beps \sim \N(\beps; \bzero, \beye)$. We also know that $\nabla_\bz \log q(\bz_t|\bz_0)=-\frac{\beps}{\sigma_t}$ With this we can write down (i) as:
\begin{align}
(I) &= \frac{1}{2} \int_0^1 \E_{q(\bz_0),\beps} \left[ \left(2f(t)(\bmu_t(\bz_0) + \sigma_t \beps) + g(t)^2 \frac{\beps}{\sigma_t} \right)^T \left(-\frac{\beps}{\sigma_t}\right) \right] dt \\
&= \int_0^1 -\frac{f(t)}{\sigma_t} \underbrace{\E_{q(\bz_0),\beps} \left[\bmu_t(\bz_0)^T \beps \right]}_{=0} - \frac{2f(t)\vt + g(t)^2}{2\vt} \underbrace{\E_{\beps}[\beps^T \beps]}_{=D} dt \\
&= -\frac{D}{2} \int_0^1 \frac{2f(t)\vt + g(t)^2}{\vt} dt \\
&= -\frac{D}{2} \int_{\sigma_0^2}^{\sigma_1^2} \frac{1}{\vt} d \vt = \frac{D}{2} (\log \sigma_0^2 - \log \sigma_1^2),
\end{align}
where we have used Eq.~\ref{eq:gen_var_diff_app}. 

Furthermore, since $q(\bz_T) \rightarrow \N(\bz_T, \bzero, \sigma_1^2 \beye)$ at $t=1$, its entropy is $\H{q(\bz_T)} = \frac{D}{2} \log(2\pi e \sigma_1^2)$. With this, we get the following simple expression for the cross-entropy:
\begin{align*}
 \CE{q(\bz_0)}{p(\bz_0)} = \frac{1}{2} \int_0^1 g(t)^2 \E_{q(\bz_0, \bz_t)} \left[||\nabla_\bz \log q(\bz_t|\bz_0) - \nabla_\bz \log p(\bz_t)||_2^2 \right] dt + D \log(\sqrt{2\pi e\sigma_0^2})
\end{align*}
Expressing the integral as an expectation completes the proof:
\begin{align*}
 \CE{q(\bz_0)}{p(\bz_0)} = \E_{t \sim \U[0, 1]}\left[\frac{g(t)^2}{2} \E_{q(\bz_t, \bz_0)} \left[||\nabla_{\bz_t} \log q(\bz_t|\bz_0)\!-\!\nabla_{\bz_t} \log p(\bz_t)||_2^2 \right]\right]\!+\!\frac{D}{2} \log\left(2\pi e\vo\right)
\end{align*}
\end{proof}
The expression in Theorem~\ref{th:simple_ce_app} measures the cross entropy between $q$ and $p$ at $t = 0$. However, one should consider practical implications of the choice of initial variance $\vo$ when estimating the cross entropy between two distributions using our expression, as we discuss below.

Consider two arbitrary distributions $q'(\bz)$ and $p'(\bz)$. If the forward diffusion process has a non-zero initial variance (i.e., $\vo > 0$), the actual distributions $q$ and $p$ at $t = 0$ in the score matching expression are defined by $q(\bz_0) := \int  q'(\bz) \N(\bz_0, \bz, \vo \beye)d\bz $ and $p(\bz_0) := \int p'(\bz) \N(\bz_0, \bz, \vo \beye) d\bz$, which correspond to convolving $q'(\bz)$ and $p'(\bz)$ each with a Normal distribution with variance $\vo \beye$. In this case, $q'(\bz)$ and $p'(\bz)$ are not identical to $q(\bz_0)$ and $p(\bz_0)$, respectively, in general. However, we can approximate $q'(\bz)$ and $p'(\bz)$ using $p(\bz_0)$ and $q(\bz_0)$, respectively, when $\vo$ is small. That is why our expression in Theorem~\ref{th:simple_ce_app} that measures $\CE{q(\bz_0)}{p(\bz_0)}$, can be considered as an approximation of $\CE{q'(\bz)}{p'(\bz)}$ when $\vo$ takes a positive small value. Note that in practice, our $\vo$ is indeed generally very small (see Tab.~\ref{table:hyperparameters}).

On the other hand, when $\vo = 0$ (e.g., when using the VPSDE from Song et al.~\cite{song2021scoreSDE}), we know that $q'(\bz)$ and $p'(\bz)$ are identical to $q(\bz_0)$ and $p(\bz_0)$. However, in this case, the initial distribution at $t=0$ is essentially an infinitely sharp Normal and we cannot evaluate the integral over the full interval $t \in [0, 1]$. Hence, we limit its range to $t \in [\epsilon, 1]$, where $\epsilon$ is another hyperparameter. In this case, we can approximate the cross entropy $\CE{q'(\bz)}{p'(\bz)}$ using:
\begin{align*}
 \CE{q(\bz_0)}{p(\bz_0)} & \approx \frac{1}{2} \int_\epsilon^1 g(t)^2 \E_{q(\bz_0, \bz_t)} \left[||\nabla_\bz \log q(\bz_t|\bz_0) - \nabla_\bz \log p(\bz_t)||_2^2 \right] dt + D \log(\sqrt{2\pi e\sigma_\epsilon^2}) \\
 & = \E_{t \sim \U[\epsilon, 1]}\left[\frac{g(t)^2}{2} \E_{q(\bz_t, \bz_0)} \left[||\nabla_{\bz_t} \log q(\bz_t|\bz_0)\!-\!\nabla_{\bz_t} \log p(\bz_t)||_2^2 \right]\right]\!+\!\frac{D}{2} \log\left(2\pi e\sigma_\epsilon^2\right)
\end{align*}

\section{Variance Reduction}\label{app:var}
The variance of the cross entropy in a mini-batch update depends on the variance of $\CE{q(\bz_0)}{p(\bz_0)}$ where $q(\bz_0) := \E_{p_{\text{data}}(\bx)}[q(\bz_0|\bx)]$ is the aggregate posterior (i.e., the distribution of latent variables) and $p_{\text{data}}$ is the data distribution. This is because, for training, we use a mini-batch estimation of $\E_{p_{\text{data}}(\bx)}[\L(\bx, \bphi, \btheta, \bpsi)]$. For the cross entropy term in $\L(\bx, \bphi, \btheta, \bpsi)$, we have $\E_{p_{\text{data}}(\bx)}[\CE{q(\bz_0|\bx)}{p(\bz_0)}] = \CE{q(\bz_0)}{p(\bz_0)}$. 

In order to study the variance of the training objective, we derive $\CE{q(\bz_0)}{p(\bz_0)}$ analytically, assuming that both $q(\bz_0) = p(\bz_0)=\N(\bz_0; \bzero, \beye)$. This is a reasonable simplification for our analysis because pretraining our LSGM model with a $\N(\bz_0; \bzero, \beye)$ prior brings $q(\bz_0)$ close to $\N(\bz_0; \bzero, \beye)$ and our SGM prior is often dominated by the fixed Normal mixture component. Nevertheless, we empirically observe that the variance reduction techniques developed with this simplification still work well when $q(\bz_0)$ and $p(\bz_0)$ are not exactly $\N(\bz_0; \bzero, \beye)$.

In this section, we start with presenting the mixed score parameterization for generic SDEs in App.~\ref{app:mixed_score_general}. Then, we discuss variance reduction with importance sampling for these generic SDEs in App.~\ref{app:var_general}. Finally, in App.~\ref{app:variance_vpsde} and App.~\ref{app:variance_vesde}, we focus on variance reduction of the VPSDEs and VESDEs, respectively, and we briefly discuss the Sub-VPSDE~\cite{song2021scoreSDE} in App.~\ref{app:variance_subvpsde}.

\subsection{Generic Mixed Score Parameterization for Non-Variance Preserving SDEs} \label{app:mixed_score_general}
The mixed score parameterization uses the score that is obtained when dealing with Normal input data and just predicts an additional residual score. In the main text, we assume that the variance of the standard Normal data stays the same throughout the diffusion process, which is the case for VPSDEs. But the way Normal data diffuses depends generally on the underlying SDE and generic SDEs behave differently than the regular VPSDE in that regard. 

Consider the generic forward SDEs in the form:
\begin{equation} \label{eq:forward_ito_app}
    \D{\bz} = f(t)\bz\D{t} + g(t)\D{\bw}
\end{equation}

If our data distribution is standard Normal, i.e. $\bz_0 \sim \mathcal{N}(\bz_0; \bzero, \beye)$, using Eq.~\ref{eq:gen_var_diff_app}, we have
\begin{equation}
    \vn_t := \frac{1}{\tilde{F}(t)}\left( \int_0^t \tilde{F}(s) g^2(s)ds + 1 \right) = \frac{1}{\tilde{F}(t)}\left( \vtil_t + 1 \right)
\end{equation}
with the definition $\vtil_t:=\int_0^t \tilde{F}(s) g^2(s)ds$. Hence, the score function at time $t$ is $\nabla_{\bz_t}\log p(\bz_t)=-\frac{\bz_t}{\vn_t}$. Using the geometric mixture $p(\bz_t) \propto \N(\bz_t; 0, \vn_t)^{1 - \alpha} p'_\btheta(\bz_t)^\alpha$, we can generally define our mixed score parameterization as
\begin{equation}
    \beps_\theta(\bz_t, t) := \frac{\sigma_t}{\vn_t} (1 - \balpha) \odot \bz_t + \balpha \odot \beps'_\theta(\bz_t, t).
\end{equation}
In the case of VPSDEs, we have $\vn_t = 1$ which corresponds to the mixed score introduced in the main text.

\textbf{Remark:} It is worth noting that both $\vn_t$ and $\vt$ are solutions to the same differential equation in Eq.~\ref{eq:gen_var_diff_app} with different initial conditions. It is easy to see that $\vn_t - \vt = (1 - \vo) \tilde{F}(t)^{-1}$.

\subsection{Variance Reduction of Cross Entropy with Importance Sampling for Generic SDEs} \label{app:var_general}
Let's consider the cross entropy expression for $p(\bz_0) = \N(\bz_0, \bzero, \beye)$ and $q(\bz_0) = \N(\bz_0, \bzero, (1 - \vo)\beye)$ where we have scaled down the variance of $q(\bz_0)$ to $(1 - \vo)$ to accommodate the fact that the diffusion process with initial variance $\vo$ applies a perturbation with variance $\vo$ in its initial step (hence, the marginal distribution at $t=0$ is $\N(\bz_0, \bzero, \beye)$ and we know that the optimal score is $\beps_\theta(\bz_t, t)=\frac{\sigma_t}{\vn_t} \bz_t$, i.e., the Normal component).

The cross entropy $\CE{q(\bz_0)}{p(\bz_0)}$ with the optimal score $\beps_\theta(\bz_t, t)=\frac{\sigma_t}{\vn_t} \bz_t$ 
is:
\begin{align}
\text{CE} - \text{const.} & =
\frac{1}{2}\int_\epsilon^1 \frac{g^2(t)}{\vt} \E_{\bz_0, \beps} \left[||\beps - \beps_\theta(\bz_t, t) ||_2^2 \right] dt \\
& = \frac{1}{2}\int_\epsilon^1 \frac{g^2(t)}{\vt} \E_{\bz_0, \beps} \left[||\beps - \frac{\sigma_t}{\vn_t} \bz_t ||_2^2 \right] dt \\
& = \frac{1}{2}\int_\epsilon^1 \frac{g^2(t)}{\vt} \E_{\bz_0, \beps} \left[||\beps - \frac{\sigma_t}{\vn_t} (\tilde{F}(t)^{-\frac{1}{2}} \bz_0  + \beps  \sigma_t) ||_2^2 \right] dt \label{eq:general_zt} \\
& = \frac{1}{2}\int_\epsilon^1 \frac{g^2(t)}{\vt} \E_{\bz_0, \beps} \left[||\frac{\vn_t - \vt}{\vn_t} \beps - \frac{\sigma_t}{\vn_t} \tilde{F}(t)^{-\frac{1}{2}} \bz_0  ||_2^2 \right] dt \\
& = \frac{1}{2}\int_\epsilon^1 \frac{g^2(t)}{\vt} \left( \frac{(\vn_t - \vt)^2}{(\vn_t)^2}   \E_{\beps} \left[||\beps||_2^2 \right] + \frac{\vt}{(\vn_t)^2} \tilde{F}(t)^{-1} \E_{\bz_0} \left[ ||\bz_0 ||_2^2 \right] \right) dt \label{eq:general_indep} \\
& = \frac{D}{2}\int_\epsilon^1 \frac{g^2(t)}{\vt} \left( \frac{(\vn_t - \vt)^2}{(\vn_t)^2}  + \frac{\vt}{(\vn_t)^2} \tilde{F}(t)^{-1} (1 - \vo) \right) dt \\
& = \frac{D}{2}\int_\epsilon^1 \frac{g^2(t)}{\vt} \left( \frac{(\vn_t - \vt)^2}{(\vn_t)^2}  + \frac{\vt (\vn_t - \vt) }{(\vn_t)^2} \right) dt \label{eq:general_Ft} \\
& = \frac{D}{2}\int_\epsilon^1 \frac{g^2(t)}{\vt}  dt - \frac{D}{2}\int_\epsilon^1 \frac{g^2(t)}{\vn_t}  dt  \\
& = \frac{D}{2}\int_\epsilon^1 \frac{\frac{d}{dt}\vt + 2f(t)\vt}{\vt}  dt - \frac{D}{2}\int_\epsilon^1 \frac{\frac{d}{dt}\vn_t + 2f(t) \vn_t}{\vn_t}  dt \label{eq:general_gt} \\
& = \frac{D}{2}\int_\epsilon^1 \frac{\frac{d}{dt}\vt}{\vt}dt - \frac{D}{2}\int_\epsilon^1 \frac{\frac{d}{dt}\vn_t}{\vn_t}dt \label{eq:general_weighted} \\
& = D\frac{1-\epsilon}{2} \E_{t \sim \U[\epsilon, 1]}\left[\frac{d}{dt}\log\left(\frac{\vt}{\vn_t}\right)\right] \\
& = D\frac{1-\epsilon}{2} \E_{t \sim \U[\epsilon, 1]}\left[\frac{d}{dt}\log\left(\frac{\vtil_t + \vo}{\vtil_t + 1}\right)\right],
\end{align}
where in Eq.~\ref{eq:general_zt}, we have used $\bz_t = \tilde{F}(t)^{-\frac{1}{2}} \bz_0  + \beps  \sigma_t$. In Eq.~\ref{eq:general_indep}, we have used the fact that $\bz_0$ and $\beps$ are independent. In Eq.~\ref{eq:general_Ft}, we have used the identity $\vn_t - \vt = (1 - \vo) \tilde{F}(t)^{-1}$. In Eq.~\ref{eq:general_gt}, we have used $g^2(t) = \frac{d}{dt}\vt + 2f(t)\vt$ from Eq.~\ref{eq:gen_var_diff_app}.

Therefore, the IW distribution with minimum variance for $\CE{q(\bz_0)}{p(\bz_0)}$ is
\begin{equation}
    r(t) \propto \frac{d}{dt}\log\left(\frac{\vtil_t + \vo}{\vtil_t + 1}\right)
\end{equation}
with normalization constant
\begin{equation}
    \tilde{R} = \log\left(\left(\frac{\vtil_1 + \vo}{\vtil_1 + 1}\right) \left(\frac{\vtil_\epsilon + 1}{\vtil_\epsilon + \vo}\right)\right)
\end{equation}
and CDF
\begin{equation}
    R(t) = \frac{1}{\tilde{R}} \log\left(\left(\frac{\vtil_t + \vo}{\vtil_t + 1}\right) \left(\frac{\vtil_\epsilon + 1}{\vtil_\epsilon + \vo}\right)\right)
\end{equation}
Hence, the inverse CDF is
\begin{equation}
    t = \left(\vtil_t\right)^{inv}\left(\frac{\vo - \left(\frac{\vtil_\epsilon + \vo}{\vtil_\epsilon + 1} \right)^{1-\rho}\left(\frac{\vtil_1 + \vo}{\vtil_1 + 1} \right)^\rho}{\left(\frac{\vtil_\epsilon + \vo}{\vtil_\epsilon + 1} \right)^{1-\rho}\left(\frac{\vtil_1 + \vo}{\vtil_1 + 1} \right)^\rho - 1}\right)
\end{equation}
Finally, the cross entropy objective with importance weighting becomes
\begin{align}
    \frac{1}{2}\int_\epsilon^1  &\frac{g^2(t)}{\vt} \E_{\bz_0, \beps} \left[||\beps - \beps_\theta(\bz_t, t) ||_2^2 \right] dt
    = \frac{\tilde{R}}{2} \E_{t \sim r(t)}\left[\frac{1 + \vtil_t}{1-\vo} \E_{\bz_0, \beps}||\beps - \beps_\theta(\bz_t, t) ||_2^2\right] \\
    & = \frac{1}{2} \log\left(\left(\frac{\vtil_1 + \vo}{\vtil_1 + 1}\right) \left(\frac{\vtil_\epsilon + 1}{\vtil_\epsilon + \vo}\right)\right) \E_{t \sim r(t)}\left[ \frac{1 + \vtil_t}{1-\vo} \E_{\bz_0, \beps}||\beps - \beps_\theta(\bz_t, t) ||_2^2\right]
\end{align}
The idea here is to write everything as a function of $\vtil_t=\int_0^t \tilde{F}(s) g^2(s)ds$. We see that $\vtil_t$ is monotonically increasing for any $g(t)$ and $f(t)$; hence, it always has an inverse and inverse transform sampling is, in principle, always possible. However, we should pick $g(t)$ and $f(t)$ such that $\vtil_t$ and its inverse are also analytically tractable to avoid dealing with numerical methods.

\subsection{VPSDE} \label{app:variance_vpsde}
Consider the simple forward diffision process in the form:
\begin{equation}
    d\bz = -\frac{1}{2}\beta(t) \bz dt + \sqrt{\beta(t)} d\bw 
\end{equation}
which corresponds to the VPSDE from Song et al.~\cite{song2021scoreSDE}. The appealing characteristic of this diffusion model is that if $\bz_0 \sim \N(\bz_0; \bzero, \beye)$, intermediate $\bz(t)$ will also have a standard Normal distribution and its variance is constant (i.e., $\frac{d}{dt}\vn_t= 0$). In the original VPSDE, $\beta(t)$ is defined by a linear function $\beta(t)=\beta_0+(\beta_1-\beta_0)t$ that interpolates between $[\beta_0, \beta_1]$.

\subsubsection{Variance Reduction for Likelihood Weighting (Geometric VPSDE)}
Our analysis in App.~\ref{app:var_general}, Eq.~\ref{eq:general_weighted} shows that the cross entropy can be expressed as:
\begin{align}
    \CE{q(\bz_0)}{p(\bz_0)} - \text{const} & = \frac{D}{2}\int_\epsilon^1 \frac{\frac{d}{dt}\vt}{\vt}dt - \frac{D}{2}\int_\epsilon^1 \frac{\frac{d}{dt}\vn_t}{\vn_t}dt \\
    & = \frac{D}{2}\int_\epsilon^1 \frac{\frac{d}{dt}\vt}{\vt}dt \\
    & = D\frac{1-\epsilon}{2} \E_{t \sim \U[\epsilon, 1]}\left[ \frac{\frac{d}{dt}\vt}{\vt} \right] \label{eq:vpsde_exp}
\end{align}
where for the VPSDE we have used $\frac{d}{dt}\vn_t= 0$. 

A sample-based estimation of this expectation has a low variance if $\frac{1}{\vt}\frac{d \vt}{dt}$ is constant for all $t\in[0,1]$. By solving the ODE $\frac{1}{\vt}\frac{d \vt}{dt} = const.$, we can see that a log-linear noise schedule of the form $\vt = \sigma^2_\text{min}(\frac{\sigma^2_\text{max}}{\sigma^2_\text{min}})^t$ satisfies this condition, with $t\in[0,1]$, $0\!<\!\vmin\!<\!\vmax\!<\!1$, and $\sigma^2_\text{min}=\vo$.

Using Eq.~\ref{eq:gen_var_diff_app}, we can find an expression for $\beta(t)$ that generates such noise schedule:
\begin{equation}
    \beta(t) = \frac{1}{1 - \vt} \frac{d \vt}{dt} = \frac{\vt}{1 - \vt} \log(\frac{\sigma^2_\text{max}}{\sigma^2_\text{min}})=\frac{\sigma^2_\text{min}(\frac{\sigma^2_\text{max}}{\sigma^2_\text{min}})^t}{1-\sigma^2_\text{min}(\frac{\sigma^2_\text{max}}{\sigma^2_\text{min}})^t}\log(\frac{\sigma^2_\text{max}}{\sigma^2_\text{min}})
\end{equation}
We call a VPSDE with $\beta(t)$ defined as above a \textit{geometric VPSDE}. For small $\sigma^2_\text{min}$ and $\sigma^2_\text{max}$ close to $1$, all inputs diffuse closely towards the standard Normal prior at $t=1$. In that regard, notice that our geometric VPSDE is well-behaved with positive $\beta(t)$ only within the relevant interval $t\in[0,1]$ and for $0\!<\!\vmin\!<\!\vmax\!<\!1$. These conditions also imply $\sigma^2_t<1$ for all $t\in[0,1]$. This is expected for any VPSDE. We can approach unit variance arbitrarily closely but not reach it exactly.

Importantly, our geometric VPSDE is different from the ``variance-exploding'' SDE (VESDE), proposed by Song et al.~\cite{song2021denoising} (also see App.~\ref{app:transitionkernels}). The VESDE leverages a SDE in which the variance grows in an almost unbounded way, while the mean of the input distribution stays constant. Because of this, the hyperparameters of the VESDE must be chosen carefully in a data-dependent manner~\cite{song2020improved}, which can be problematic in our case (see discussion in App.~\ref{app:variance_vesde}). Furthermore, Song et al. also found that the VESDE does not perform well when used with probability flow-based sampling~\cite{song2021scoreSDE}. In contrast, our geometric VPSDE combines the variance preserving behavior (i.e. standard Normal input data remains standard Normal throughout the diffusion process; all individual inputs diffuse towards standard Normal prior) of the VPSDE with the geometric growth of the variance in the diffusion process, which was first used in the VESDE.

Finally, for the geometric VPSDE we also have that $\frac{\partial}{\partial t}\CE{q(\bz_t)}{p(\bz_t)}=const.$ for Normal input data. Hence, data is encoded ``as continuously as possible'' throughout the diffusion process. This is in line with the arguments made by Song et al. in~\cite{song2020improved}. We hypothesize that this is particularly beneficial towards learning models with strong likelihood or NELBO performance. Indeed, in our experiments we observe the geometric VPSDE to perform best on this metric.

\subsubsection{Variance Reduction for Likelihood Weighting (Importance Sampling)}

Above, we have assumed that we sample from a uniform distribution for $t$ and we have defined $\beta(t)$ and $\vt$ such that the variance of a Monte-Carlo estimation of the expectation is minimum. Another approach for improving the sample-based estimate of the expectation is to keep $\beta(t)$ and $\vt$ unchanged and to use importance sampling such that the variance of the estimate is minimum. 

Using importance sampling, we can rewrite the expectation in Eq.~\ref{eq:vpsde_exp} as:
\begin{equation}
 \E_{t \sim \U[\epsilon, 1]} \left[ \frac{1}{\vt}\frac{d \vt}{dt} \right] = \E_{t \sim r(t)} \left[\frac{1}{r(t)} \frac{1}{\vt}\frac{d \vt}{dt} \right]
\end{equation}
where $r(t)$ is a proposal distribution. The theory of importance sampling~\cite{owenMCbook} shows that $r(t) \propto \frac{1}{\vt}\frac{d \vt}{dt} = \frac{d \log \vt}{dt}$ will have the smallest variance. In order to use this proposal distribution, we require (i) sampling from $r(t)$ and (ii) evaluating the objective using this importance sampling technique.

\textbf{Sampling from $r(t)$ by inverse transform sampling:}
It's easy to see that the normalization constant for $r(t)$ is $\int_\epsilon^1 \frac{d \log \vt}{dt} dt = \log \sigma_1^2 - \log \sigma_\epsilon^2$. Thus, the PDF $r(t)$ is:
\begin{equation}
r(t) = \frac{1}{\log \sigma_1^2 - \log \sigma_\epsilon^2} \frac{1}{\vt}\frac{d \vt}{dt} = \frac{\beta(t)(1-\vt)}{(\log \sigma_1^2 - \log \sigma_\epsilon^2) \vt}
\end{equation}

We can derive inverse transform sampling by deriving the inverse CDF:
\begin{equation}
    R(t) = \frac{\log \frac{\vt}{\sigma_\epsilon^2}}{\log \frac{\sigma_1^2}{\sigma_\epsilon^2}} = \rho \Rightarrow \frac{\vt}{\sigma_\epsilon^2} = \left(\frac{\sigma_1^2}{\sigma_\epsilon^2}\right)^\rho \Rightarrow t = \text{var}^{-1}\left(\left(\sigma_1^2\right)^\rho \left(\sigma_\epsilon^2\right)^{1 - \rho}\right)
\end{equation}
where $\text{var}^{-1}$ is the inverse of $\vt$. 

\textbf{Importance Weighted Objective:} The cross entropy is then written as (ignoring the constants here):
\begin{align}
\frac{1}{2}\int_\epsilon^1 & \frac{\beta(t)}{\vt} \E_{\bz_0, \beps} \left[||\beps - \beps_\theta(\bz_t, t) ||_2^2 \right] dt = \frac{1}{2} \E_{t \sim r(t)} \left[ \frac{(\log \sigma_1^2 - \log \sigma_\epsilon^2)}{(1-\vt)} \E_{\bz_0, \beps} ||\beps - \beps_\theta(\bz_t, t) ||_2^2 \right]
\end{align}

\subsubsection{Variance Reduction for Unweighted Objective}
Using a similar derivation as in App.~\ref{app:var_general}, we can show that for the unweighted objective for $p(\bz_0) = \N(\bz_0, \bzero, \beye)$ and $q(\bz_0) = \N(\bz_0, \bzero, (1 - \vo)\beye)$, we have
\begin{align}
 \int_\epsilon^1 \E_{\bz_0, \beps} \left[||\beps - \beps_\theta(\bz_t, t) ||_2^2 \right] dt & = \frac{D}{2}\int_\epsilon^1  \left( \frac{(\vn_t - \vt)^2}{(\vn_t)^2}  + \frac{\vt (\vn_t - \vt) }{(\vn_t)^2} \right) dt \\
 & = D\frac{1-\epsilon}{2} \E_{t \sim \U[\epsilon, 1]} \left[ 1-\vt \right] \\
 & = D\frac{1-\epsilon}{2} \E_{t \sim r(t)} \left[\frac{ 1-\vt}{r(t)} \right]
\end{align}
with proposal distribution $r(t)\propto 1-\vt$. Recall that in the VPSDE with linear $\beta(t)=\beta_0+(\beta_1-\beta_0)t$, we have
\begin{equation}
    1-\vt=(1-\sigma^2_0)e^{-\int_0^t\beta(s)ds}=(1-\sigma^2_0)e^{-\beta_0 t-(\beta_1-\beta_0)\frac{t^2}{2}}
\end{equation}
Hence, the normalization constant of $r(t)$ is
\begin{align}
    \tilde{R} & = \int_\epsilon^1 (1-\sigma^2_0)e^{-\beta_0 t-(\beta_1-\beta_0)\frac{t^2}{2}} dt \\
    & = \underbrace{(1-\sigma^2_0)e^{\frac{1}{2}\frac{\beta_0}{\beta_1-\beta_0}}\sqrt{\frac{\pi}{2(\beta_1-\beta_0)}}}_{:=A_{\tilde{R}}} \left[\textrm{erf}\left(\sqrt{\frac{\beta_1-\beta_0}{2}}\left[1+\frac{\beta_0}{\beta_1-\beta_0}\right]\right) - \textrm{erf}\left(\sqrt{\frac{\beta_1-\beta_0}{2}}\left[\epsilon + \frac{\beta_0}{\beta_1-\beta_0}\right]\right) \right]
\end{align}
Similarly, we can write the CDF of $r(t)$ as
\begin{align}
    R(t) & =\frac{A_{\tilde{R}}}{\tilde{R}}\left[\textrm{erf}\left(\sqrt{\frac{\beta_1-\beta_0}{2}}\left[t+\frac{\beta_0}{\beta_1-\beta_0}\right]\right) - \textrm{erf}\left(\sqrt{\frac{\beta_1-\beta_0}{2}}\left[\epsilon + \frac{\beta_0}{\beta_1-\beta_0}\right]\right) \right]
\end{align}
solving $\rho=R(t)$ for $t$ then results in
\begin{align}
    t = \sqrt{\frac{2}{\beta_1-\beta_0}}\textrm{erfinv}\left(\frac{\rho \tilde{R}}{A_{\tilde{R}}}+\textrm{erf}\left(\sqrt{\frac{\beta_1-\beta_0}{2}}\left[\epsilon + \frac{\beta_0}{\beta_1-\beta_0}\right]\right) \right)-\frac{\beta_0}{\beta_1-\beta_0}
\end{align}

\textbf{Importance Weighted Objective:}
\begin{align}
\int_\epsilon^1 & \E_{\bz_0, \beps} \left[||\beps - \beps_\theta(\bz_t, t) ||_2^2 \right] dt = \E_{t \sim r(t)} \left[ \frac{\tilde{R}}{(1-\vt)} \E_{\bz_0, \beps} ||\beps - \beps_\theta(\bz_t, t) ||_2^2 \right]
\end{align}

\subsubsection{Variance Reduction for Reweighted Objective}

For the reweighted mechanism, we drop only $\vt$ from the cross entropy objective but we keep $g^2(t) = \beta(t)$. Using a similar derivation in App.~\ref{app:var_general}, we can show that unweighted objective for $p(\bz_0) = \N(\bz_0, \bzero, \beye)$ and $q(\bz_0) = \N(\bz_0, \bzero, (1 - \vo)\beye)$, we have
\begin{equation}
\int_\epsilon^1 \beta(t) \E_{\bz_0, \beps} \left[||\beps - \beps_\theta(\bz_t, t) ||_2^2 \right] dt = D\frac{1-\epsilon}{2} \E_{t \sim \U[\epsilon, 1]} \left[ \frac{d \vt}{dt} \right] = D\frac{1-\epsilon}{2} \E_{t \sim r(t)} \left[\frac{ \frac{d \vt}{dt}}{r(t)} \right]
\end{equation}
with proposal distribution $r(t)\propto \frac{d \vt}{dt} = \beta(t)(1-\vt)$.

In this case, we have the following proposal $r(t)$, its CDF $R(t)$ and inverse CDF $R^{-1}(\rho)$:
\begin{equation}
r(t) = \frac{\beta(t)(1-\vt)}{\sigma^2_1 - \sigma^2_\epsilon}, \quad R(t)=\frac{\vt - \sigma^2_\epsilon}{\sigma^2_1 - \sigma^2_\epsilon}, \quad t = R^{-1}(\rho) = \text{var}^{-1}((1 - \rho)\sigma^2_\epsilon + \rho \sigma^2_1)
\end{equation}
Note that usually $\sigma^2_\epsilon \gtrapprox 0$ and $\sigma^2_1 \lessapprox 1$. In that case, the inverse CDF can be thought of as $R^{-1}(\rho) \approx \text{var}^{-1}(\rho)$.

\textbf{Importance Weighted Objective:}
\begin{align}
\frac{1}{2}\int_\epsilon^1 & \beta(t) \E_{\bz_0, \beps} \left[||\beps - \beps_\theta(\bz_t, t) ||_2^2 \right] dt = \frac{1}{2} \E_{t \sim r(t)} \left[ \frac{(\sigma_1^2 - \sigma_\epsilon^2)}{(1-\vt)} \E_{\bz_0, \beps} ||\beps - \beps_\theta(\bz_t, t) ||_2^2 \right]
\end{align}

\textbf{Remark:} It is worth noting that the derivation of the importance sampling distribution for the reweighted objective does not make any assumption on the form of $\beta(t)$. Thus, the IS distribution can be formed for any VPSDE when training with the reweighted objective, including the original VPSDE with linear $\beta(t)$ and also our new geometric VPSDE.

\subsection{VESDE} \label{app:variance_vesde}

The VESDE~\cite{song2021scoreSDE} is defined by:
\begin{align}
    d\bz & = \sqrt{\frac{d}{dt}\sigma(t)^2} d\bw \\
    & = \sqrt{\vmin\log\left(\frac{\vmax}{\vmin}\right) \left(\frac{\vmax}{\vmin}\right)^t } d\bw
\end{align}
with $\sigma(t)^2=\vmin\left(\frac{\vmax}{\vmin}\right)^t$.

Solving the Fokker-Planck equation for input distribution $\mathcal{N}(\mu_0,\vo)$ results in
\begin{equation} \label{eq:vesde_normal_diff}
    \mu_t = \mu_0; \qquad \vt = \vo - \vmin + \vmin\left(\frac{\vmax}{\vmin}\right)^t
\end{equation}
Typical values for $\vmin$ and $\vmax$ are $\vmin=0.01^2$ and $\vmax=50^2$ (CIFAR10). Usually, we use $\vmin=\vo$.

Note that when the input data is distributed as $\bz_0 \sim \N(\bz_0; \bzero, \beye)$, the variance 
at time $t$ in VESDE is given by:
\begin{equation}
    \vn_t = 1 - \vmin + \vmin\left(\frac{\vmax}{\vmin}\right)^t
\end{equation}

Note that $\vmax$ is typically very large and chosen empirically based on the scale of the data~\cite{song2020improved}. However, this is tricky in our case, as the role of the data is played by the latent space encodings, which themselves are changing during training. We did briefly experiment with the VESDE and calculated $\vmax$ as suggested in~\cite{song2020improved} using the encodings after the VAE pre-training stage. However, these experiments were not successful and we suffered from significant training instabilities, even with variance reduction techniques. Therefore, we did not further explore this direction.

Nevertheless, our proposed variance reduction techniques via importance sampling can be derived also for the VESDE. Hence, for completeness, they are shown below.

\subsubsection{Variance Reduction for Likelihood Weighting}
Let's have a closer look at the likelihood objective when using the VESDE for modeling the standard Normal data. Following similar arguments as in previous sections, we have $\bz_0 \sim \mathcal{N}(\bz_0; \bzero, (1-\vmin) \beye)$. With the optimal score $\beps_\theta(\bz_t, t)=\frac{\sigma_t}{\vn_t} \bz_t$ (i.e., the Normal component), we have the following expression for $\CE{q(\bz_0)}{p(\bz_0)}$ from Eq.~\ref{eq:general_weighted}:
\begin{align}
\frac{1}{2}\int_\epsilon^1 & \frac{g^2(t)}{\vt} \E_{\boldsymbol{\mu}_0, \beps} \left[||\beps - \beps_\theta(\bz_t, t) ||_2^2 \right] dt = \frac{D}{2}\int_\epsilon^1 \frac{\frac{d}{dt}\vt}{\vt}dt - \frac{D}{2}\int_\epsilon^1 \frac{\frac{d}{dt}\vn_t}{\vn_t}dt =\\
\frac{D}{2}\int_\epsilon^1 & \left[ \frac{\frac{d}{dt}\vt}{\vt} -  \frac{\frac{d}{dt}\vn_t}{\vn_t}\right]dt = D\frac{1-\epsilon}{2} \E_{t \sim \mathcal{U}[\epsilon,1]}\left[ \frac{\frac{d}{dt}\vt}{\vt} -  \frac{\frac{d}{dt}\vn_t}{\vn_t}\right]
\end{align}
\textit{Since the term inside the expectation is not constant in $t$, the VESDE does not result in an objective with naturally minimal variance, opposed to our proposed geometric VPSDE.}

We derive an importance sampling scheme with a proposal distribution
\begin{align}
    r(t)\propto \frac{1}{\vt}\frac{d\vt}{dt} - \frac{1}{\vn_t}\frac{d\vn_t}{dt} = \log\left(\frac{\vmax}{\vmin}\right) \left(1 - \frac{\vmin\left(\frac{\vmax}{\vmin}\right)^t}{1 - \vmin + \vmin\left(\frac{\vmax}{\vmin}\right)^t} \right)
\end{align}
Note that the quantity above is always positive as $ \frac{\vmin\left(\frac{\vmax}{\vmin}\right)^t}{1 - \vmin + \vmin\left(\frac{\vmax}{\vmin}\right)^t} \leq 1$ with $\vmin < 1$. In this case the normalization constant of $r(t)$ is $\tilde{R} = \log\left(\frac{\vn_\epsilon}{\sigma_\epsilon^2}\frac{\vmax}{\vn_1} \right)$ and the CDF is:
\begin{equation}
    R(t) = \frac{1}{\tilde{R}} \left[\log \vt - \log \sigma_\epsilon^2 + \log \vn_\epsilon - \log \vn_t \right] = \frac{1}{\tilde{R}} \log \left(\frac{\vn_\epsilon \vt}{\vn_t \sigma_\epsilon^2} \right)
\end{equation}
And the inverse CDF is:
\begin{align}
    t= \mathring{\text{var}}^{-1}\left(\frac{1 - \vmin}{1 - \left(\frac{\sigma_\epsilon^2}{\vn_\epsilon}\right)^{1-\rho} \left(\frac{\vmax}{\vn_1} \right)^\rho} \right)
\end{align}
where $\mathring{\text{var}}^{-1}$ is the inverse of $\vn_t$.

So, the objective with importance sampling is then:
\begin{align*}
    \frac{1}{2}\int_\epsilon^1 \frac{g^2(t)}{\vt} \E_{\bz_0, \beps} \left[||\beps - \beps_\theta(\bz_t, t) ||_2^2 \right] dt &= \frac{1}{2}\E_{t \sim r(t)} \left[ \log\left(\frac{\vn_\epsilon}{\sigma_\epsilon^2}\frac{\vmax}{\vn_1} \right) \frac{\vn_t}{1-\vmin} \E_{\bz_0, \beps}||\beps - \beps_\theta(\bz_t, t) ||_2^2 \right]
\end{align*}

In contrast to the VESDE, the geometric VPSDE combines the geometric progression in diffusion variance directly with minimal variance in the objective by design. Furthermore, it is simpler to set up, because we can always choose $\vmax\sim1$ for the geometric VPSDE and do not have to use a data-specific $\vmax$ as proposed by \cite{song2020improved}.

\subsubsection{Variance Reduction for Unweighted Objective}
When we drop all ``prefactors'' in the objective, the importance sampling distribution stays the same as above, since $\frac{g^2(t)}{\vt}$ is constant in $t$. The objective becomes:
\begin{equation}
\int_\epsilon^1 \E_{\bz_0, \beps} \left[||\beps - \beps_\theta(\bz_t, t) ||_2^2 \right] dt = \E_{t \sim r(t)} \left[ \frac{\log\left(\frac{\vn_\epsilon}{\sigma_\epsilon^2}\frac{\vmax}{\vn_1} \right)}{\log\left(\frac{\vmax}{\vmin}\right)} \frac{\vn_t}{1-\vmin} \E_{\bz_0, \beps}||\beps - \beps_\theta(\bz_t, t) ||_2^2 \right]
\end{equation}

\subsubsection{Variance Reduction for Reweighted Objective}
To define the importance sampling for the reweighted objective by $\vt$, we use the fact that $\frac{d\vt}{dt} = \frac{d\vn_t}{dt}$ in VESDEs. Using a similar derivation as in App.~\ref{app:var_general}, we show:
\begin{align}
    \frac{1}{2}\int_\epsilon^1 g^2(t) \E_{\bz_0, \beps} \left[||\beps - \beps_\theta(\bz_t, t) ||_2^2 \right] dt &= \frac{D}{2}\int_\epsilon^1 \frac{d \vt}{dt} dt - \frac{D}{2}\int_\epsilon^1 \frac{d \vn_t}{dt} \frac{\vt}{\vn_t}dt \\
    & = \frac{D}{2}\int_\epsilon^1 \frac{d \vn_t}{dt} \left(\frac{\vn_t - \vt}{\vn_t} \right) dt \\
    & = \frac{D (1 - \vo)}{2}\int_\epsilon^1 \frac{1}{\vn_t} \frac{d \vn_t}{dt} dt
\end{align}

Thus, the optimal proposal for reweighted objective and the inverse CDF are:
\begin{equation}
    r(t) \sim \frac{1}{\vn_t} \frac{d\vn_t}{dt} \Rightarrow r(t) = \frac{1}{\log(\frac{\vn_1}{\vn_\epsilon})}\frac{1}{\vn_t}\frac{d\vn_t}{dt} \Rightarrow R(t) =\frac{\log(\frac{\vn_t}{\vn_\epsilon})}{\log(\frac{\vn_1}{\vn_\epsilon})} \Rightarrow t= \mathring{\text{var}}^{-1}\left((\vn_\epsilon)^{1-\rho}(\vn_1)^\rho\right) 
\end{equation}

So, the reweighted objective with importance sampling is:
\begin{equation}
\frac{1}{2}\int_\epsilon^1 g^2(t) \E_{\boldsymbol{\mu}_0, \beps} \left[||\beps - \beps_\theta(\bz_t, t) ||_2^2 \right] dt = \frac{1}{2}\E_{t \sim r(t)} \left[ \log \left(\frac{\vn_1}{\vn_\epsilon}\right) \vn_t \E_{\boldsymbol{\mu}_0, \beps}||\beps - \beps_\theta(\bz_t, t) ||_2^2 \right]
\end{equation}

Note that in practice, we can safely set $\epsilon = 0$ as initial $\vo$ is non-zero in the VESDE.

\subsection{Sub-VPSDE} \label{app:variance_subvpsde}
Song et al. also proposed the Sub-VPSDE~\cite{song2021scoreSDE}. It is defined as:
\begin{align}
    d\bz = -\frac{1}{2}\beta(t) \bz dt + \sqrt{\beta(t)\left(1 - e^{- 2\int_0^t \beta(s) ds}\right)} d\bw
\end{align}
with the same linear $\beta(t)$ as for the regular VPSDE.

Solving the Fokker-Planck equation for input distribution $\mathcal{N}(\mu_0,\vo)$ at $t=0$ results in
\begin{equation} \label{eq:sub_vpsde_normal_diff}
    \mu_t = e^{-\frac{1}{2} \int_0^t \beta(s) ds} \mu_0; \qquad \vt = \left(1.0 - e^{- \int_0^t \beta(s) ds}\right)^2 + \vo \, e^{- \int_0^t \beta(s) ds}
\end{equation}

\begin{wrapfigure}{r}{0.5\textwidth}
  \begin{center}
    \includegraphics[width=0.5\textwidth]{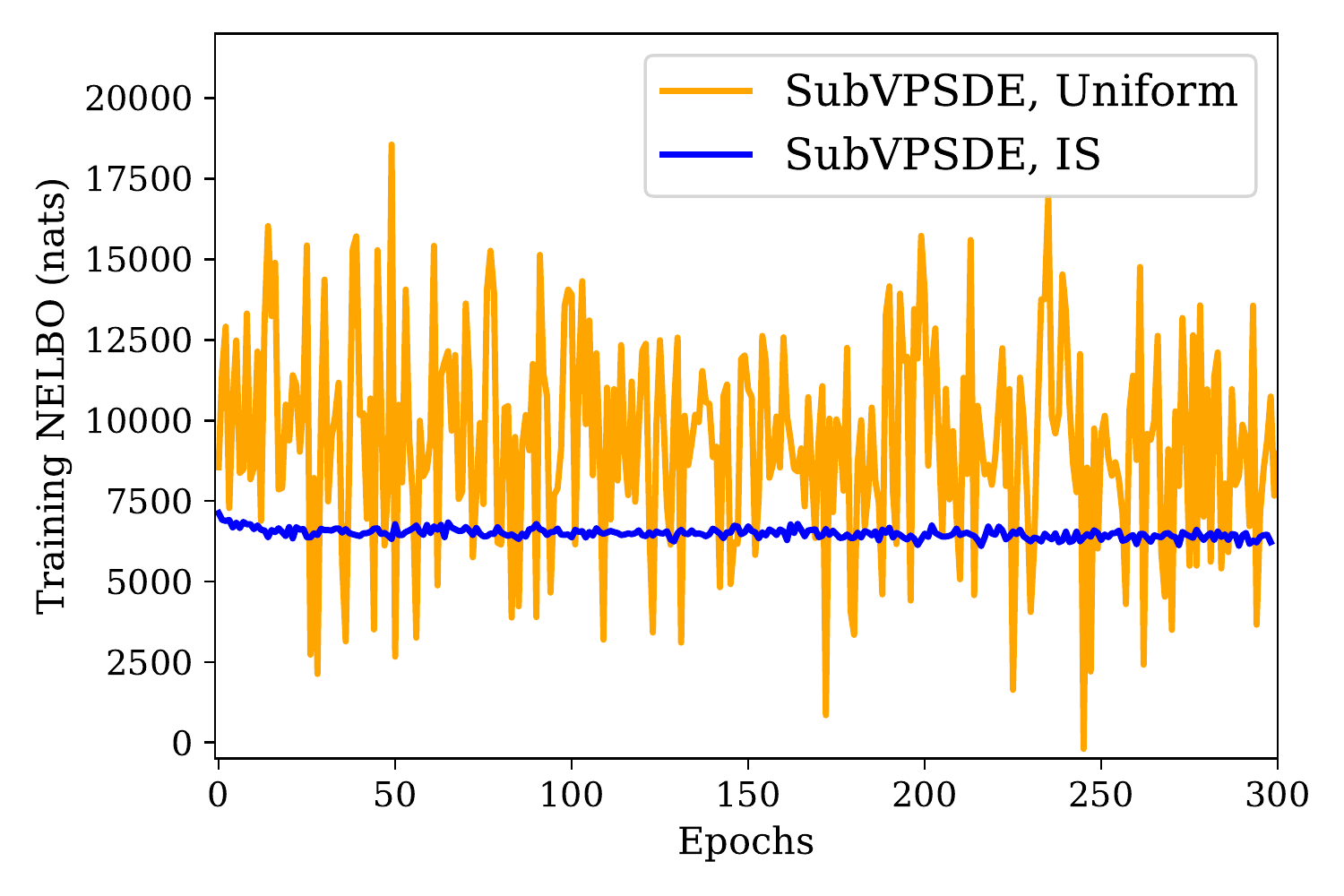}
  \end{center}
  \caption{\small Variance reduction of the sample-based estimate of the training objective for the Sub-VPSDE, using an IS distribution derived from the regular VPSDE.}
  \label{fig:variance_subvpsde}
\end{wrapfigure}
Deriving importance sampling distributions for variance reduction for the Sub-VPSDE can be more complicated than for the VPSDE, Geometric VPSDE, and VESDE and we did not investigate this in detail. However, for the same linear $\beta(t)$ the Sub-VPSDE is close to the VPSDE, only slightly reducing the the variance $\sigma^2_t$ of the diffusion process distribution for small $t$. This suggests that the IS distribution derived using the regular VPSDE will likely also significantly reduce the variance of the objective due to $t$-sampling of the Sub-VPSDE, just not as optimally as theoretically possible. 
In Fig.~\ref{fig:variance_subvpsde}, we show the training NELBO of an LSGM trained on CIFAR-10 with $w_\we$-weighting using the Sub-VPSDE. We show the NELBO both for uniform $t$ sampling as well as for $t$ sampling from the IS distribution that was originally derived for the regular VPSDE with the same $\beta(t)$ (the experiment and model setup is otherwise the same as the one for the ablation study on SDEs, weighting mechanisms and variance reduction).
We indeed observe a significantly reduced training objective variance. We were consequently able to train large LSGM models in a stable manner using the Sub-VPSDE with VPSDE-based IS. However, the strongest generative performance in either NLL or FID was not achieved using the Sub-VPSDE, but with the Geometric VPSDE or regular VPSDE. For that reason, we did not focus on the Sub-VPSDE in our main experiments. However, a generative modeling performance comparison of the VPSDE vs. Sub-VPSDE in a smaller LSGM model is presented in App.~\ref{app:subvpvsvp}.

\section{Expressions for the Normal Transition Kernel} \label{app:transitionkernels}

In our derivations of the Normal transition kernel $q(\bz_t|\bz_0)$, we only considered the general case in Eq.~\ref{eq:gen_mean_diff_app} and Eq.~\ref{eq:gen_var_diff_app}. However, the expression for $q(\bz_t|\bz_0)$ can be further simplified for different SDEs that are considered in this paper. For completeness, we provide the expressions for $q(\bz_t|\bz_0)$ below:
\begin{equation}
    q(\bz_t|\bz_0) = \begin{cases}
        \N\Big(\bz_t; e^{-\frac{1}{2} \int_0^t \beta(s) ds} \bz_0, \left[ 1 - (1 - \vo)e^{- \int_0^t \beta(s) ds} \right] \beye \Big) &\text{VPSDE (linear $\beta(t)$)}  \\
        \N\Big(\bz_t; \sqrt{\frac{1 - \sigma^2_\text{min}(\frac{\sigma^2_\text{max}}{\sigma^2_\text{min}})^t}{1 - \sigma^2_\text{min}}} \bz_0, \sigma^2_\text{min}(\frac{\sigma^2_\text{max}}{\sigma^2_\text{min}})^t \beye\Big) &\text{Geometric VPSDE} \\
        \N\Big(\bz_t; \bz_0, \sigma^2_\text{min}(\frac{\sigma^2_\text{max}}{\sigma^2_\text{min}})^t \beye\Big) &\text{VESDE}
        \end{cases}
\end{equation}
In both VESDE and Geometric VPSDE, the initial variance $\vo$ is denoted by $\sigma^2_\text{min} > 0$. These diffusion processes start from a slightly perturbed version of the data at $t\!=\!0$. In VESDE, $\sigma^2_\text{max}$ by definition is large (as the name variance exploding SDE suggests) and it is set based on the scale of the data~\cite{song2020improved}. In contrast, $\sigma^2_\text{max}$ in the Geometric VPSDE does not depend on the scale of the data and it is set to $\sigma^2_\text{max} \approx 1$. In the VPSDE, the initial variance is denoted by the hyperparameter $\vo$. In contrast to VESDE and Geometric VPSDE, we often set the initial variance to zero in VPSDE, meaning that the diffusion process models the data distribution exactly at $t\!=\!0$. However, using the VPSDE with $\vo=0$ comes at the cost of not being able to sample $t$ in the full interval $[0,1]$ during training and also prevents us from solving the probability flow ODE all the way to zero during sampling~\cite{song2021scoreSDE}.

\section{Probability Flow ODE} \label{app:prob_flow_ode}
In LSGM, to sample from our SGM prior in latent space and to estimate NELBOs, we follow Song et al.~\cite{song2021scoreSDE} and build on the connection between SDEs and ODEs. We use black-box ODE solvers to solve the probability flow ODE. Here, we briefly recap this approach.

All SDEs used in this paper can be written in the general form
\begin{equation*}
    d\bz = \bff(\bz, t) dt + g(t) d\bw 
\end{equation*}

The reverse of this diffusion process is also a diffusion process running backwards in time~\cite{anderson1982reverse,song2021scoreSDE}, defined by
\begin{equation*}
    d\bz = \left[\bff(\bz, t) - g^2(t) \nabla_{\bz_t} \log q(\bz_t) \right]dt + g(t) d\bar{\bw},
\end{equation*}
where $d\bar{\bw}$ denotes a standard Wiener process going backwards in time, $dt$ now represents a negative infinitesimal time increment, and $\nabla_{\bz_t} \log q(\bz_t)$ is the score function of the diffusion process distribution at time $t$. Interestingly, Song et al. have shown that there is a corresponding ODE that generates the same marginal probability distributions $q(\bz_t)$ when acting upon the same prior distribution $q(\bz_1)$. It is given by
\begin{equation*}
    d\bz = \left[\bff(\bz, t) - \frac{g^2(t)}{2} \nabla_{\bz_t} \log q(\bz_t) \right]dt
\end{equation*}
and usually called the \textit{probability flow ODE}. This connects score-based generative models using diffusion processes to continuous Normalizing flows, which are based on ODEs~\cite{chen2018neuralODE,grathwohl2019ffjord}. Note that in practice $\nabla_{\bz_t} \log q(\bz_t)$ is approximated by a learnt model. Therefore, the generative distributions defined by the ODE and SDE above are formally not exactly equivalent when inserting this learnt model for the score function expression. Nevertheless, they often achieve quite similar performance in practice~\cite{song2021scoreSDE}. This aspect is discussed in detail in concurrent work by Song et al.~\cite{song2021MLSGM}.

We can use the above ODE for efficient sampling of the model via black-box ODE solvers. Specifically, we can draw samples from the standard Normal prior distribution at $t=1$ and then solve this ODE towards $t=0$. In fact, this is how we perform sampling from the latent SGM prior in our paper. Similarly, we can also use this ODE to calculate the probability of samples under this generative process using the instantaneous change of variables formula (see \cite{chen2018neuralODE,grathwohl2019ffjord} for details). We rely on this for calculating the probability of latent space samples under the score-based prior in LSGM. Note that this involves calculating the trace of the Jacobian of the ODE function. This is usually approximated via Hutchinson's trace estimator, which is unbiased but has a certain variance (also see discussion in Sec.~\ref{app:bias}).

This approach is applicable similarly for all diffusion processes and SDEs considered in this paper.
\setcounter{section}{4}
\section{Converting VAE with Hierarchical Normal Prior to Standard Normal Prior} \label{app:hvae}
Converting a VAE with hierarchical prior to a standard Normal prior can be done using a simple change of variables. Consider a VAE with hierarchical encoder $q(\bz|\bx) = \prod_l q(\bz_l | \bz_{<l}, \bx)$ and hierarchical prior $p(\bz) = \prod_l p(\bz_l |\bz_{<l})$ where $\bz=\{\bz_l\}_{l=1}^{L}$ represent all latent variables and:
\begin{align}
 p(\bz_l |\bz_{<l}) &= \N(\bz_l; \bmu_l(\bz_{<l}),  \sigma^2_l(\bz_{<l}) \beye)  \label{eq:hprior} \\
q(\bz_l | \bz_{<l}, \bx) &= \N(\bz_l; \bmu'_l(\bz_{<l}, \bx), \sigma'^2_l(\bz_{<l}, \bx) \beye) 
\end{align}
where for simplicity we have assumed that the variance is shared for all the components. We can reparameterize the latent variables by introducing $\beps_l = \frac{\bz_l - \bmu_l(\bz_{<l})}{\sigma_l(\bz_{<l})}$. With this reparameterization, the equivalent VAE is:
\begin{align}
 p(\beps_l) &= \N(\beps_l; \bzero,  \beye)   \\
 q(\beps_l|\beps_{<l}, \bx) &= \N(\beps_l; \frac{\bmu'_l(\bz_{<l}, \bx)- \bmu_l(\bz_1)}{\sigma_l(\bz_{<l})}, \frac{\sigma'^2_l(\bz_{<l}, \bx)}{\sigma^2_l(\bz_{<l})} \beye), \\
\end{align}
where $\bz_l = \bmu_l(\bz_{<l}) + \sigma_l(\bz_{<l}) \beps_l$. In this equivalent parameterization, we can consider $\beps_l$ as latent variables with a standard Normal prior.

\subsection{Converting NVAE Prior to Standard Normal Prior}
In NVAE~\cite{vahdat2020nvae}, the prior has the same hierarchical form as in Eq.~\ref{eq:hprior}. However, the authors observe that the \textit{residual parameterization} of the encoder often improves the generative performance. In this parameterization, with a small modification, the encoder is defined by:
\begin{align}
 q(\bz_l|\bz_{<l}, \bx) &= \N(\bz_l; \bmu_l(\bz_{<l}) + \sigma_l(\bz_{<l}) \Delta \bmu'_l(\bz_{<l}, \bx), \sigma^2_l(\bz_{<l}) \Delta \sigma'^2_l(\bz_{<l}, \bx) \beye),
\end{align}
where the encoder is tasked to predict the residual parameters $\Delta \bmu'_l(\bz_{<l}, \bx)$ and $\Delta \sigma'^2_l(\bz_{<l}, \bx)$. Using the same reparameterization as above ($\beps_l = \frac{\bz_l - \bmu_l(\bz_{<l})}{\sigma_l(\bz_{<l})}$), we have the equivalent VAE in the form:
\begin{align}
 p(\beps_l) &= \N(\beps_l; \bzero,  \beye)  \label{eq:nvae_prior} \\
 q(\beps_l|\beps_{<l}, \bx) &= \N(\beps_l; \Delta \bmu'_l(\bz_{<l}, \bx), \Delta \sigma'^2_l(\bz_{<l}, \bx) \beye),  \label{eq:nvae_encoder} 
\end{align}
where $\bz_l = \bmu_l(\bz_{<l}) + \sigma_l(\bz_{<l}) \beps_l$. In other words, the residual parameterization of encoder, introduced in NVAE, predicts the mean and variance for the $\beps_l$ distributions directly.

\section{Bias in Importance Weighted Estimation of Log-Likelihood} \label{app:bias}
A common approach for estimating test log-likelihood in VAEs is to use the importance weighted bound on log-likelihood~\cite{burda2015importance}. In LSGM, we have access to an unbiased but stochastic estimation of the prior likelihood $\log p(\bz_0)$ which we obtain using the probability flow ODE~\cite{song2021scoreSDE}. The stochasticity in the estimation comes from Hutchinson's trick~\cite{grathwohl2019ffjord}. In VAEs, the test log-likelihood is estimated using importance weighted (IW) estimation~\cite{burda2015importance}:
{\small 
\begin{equation} \hspace{-0.4cm} \label{eq:iw}
    \E_{\bz^{(1)}, \dots, \bz^{(K)} \sim q(\bz|\bx)}[\log (\frac{1}{K} \sum_{k=1}^{K} \exp( w^{(k)}))] \quad \text{where} \quad w^{(k)} = \log p(\bz^{(k)}) + \log p(\bx|\bz^{(k)}) - \log q(\bz^{(k)}|\bx)
\end{equation}}
which is a statistical lower bound on $\log p(\bx)$.

In this section, we provide an informal analysis that shows that IW estimation with $K>1$ can overestimate the log-likelihood when $\log p(z)$ is measured with an unbiased estimator with variance $\sigma^2$. In our analysis we assume that $\sigma^2$ is small and we use Taylor expansion to study how the IW bound varies. 
Under our analysis, we observe that the bias has $O(\sigma^2)$ and it can be minimized by ensuring that $\sigma^2$ is sufficiently small.

Consider the Taylor expansion around $\bw$ up to second order of the function $\log\sum\exp(\bw) = \log \sum_k e^{w_i}$ where $\bw =\{w^{(k)}\}_{k=1}^K$ ($\log\sum\exp: \R^K \rightarrow \R$). 
With $\beps \sim \N(\beps, \bzero, \beye)$ and assuming that $\sigma^2$ is sufficiently small so that all terms beyond second order contribute negligibly, we have:
\begin{align}
    \E_{\beps} [\log\sum\exp(\bw + \sigma \beps)] \approx \log\sum\exp(\bw) + \sigma \cancelto{\bzero}{\E_{\beps} [\beps^T]} \nabla_{\bw} \log\sum\exp(\bw) + \sigma^2 \cancelto{\text{trace}(\bH)}{\E_{\beps} [\beps^T \bH \beps]}
\end{align}
where $\bH$ is the Hessian matrix for the $\log\sum\exp$ function at $\bw$. Note that the gradient $\nabla_{\bw} \log\sum\exp(\bw) = \frac{e^{w_i}}{\sum_j e^{w_j}}$ is the softmax function and $\text{trace}(\bH)\!=\!\sum_i\!\frac{e^{w_i}}{\sum_j e^{w_j}}\!\left(1 - \frac{e^{w_i}}{\sum_j e^{w_j}}\right) \leq 1$. Thus, we have:
\begin{align}
    \E_{\beps} [\log\sum\exp(\bw + \sigma \beps)] \lessapprox \log\sum\exp(\bw) + \sigma^2
\end{align}
So, when the importance weights $\bw =\{w^{(k)}\}_{k=1}^K$ are estimated with sufficiently small variance $\sigma^2$, the bias is proportional to the variance of this estimate.

In our experiments, we observe that the variance of the $\log p(\bz_0)$ estimate is not small enough to obtain a reliable estimate of test likelihood using the importance weighted bound. One way to reduce the variance is to use many randomly sampled noise vectors in Hutchinson's trick. However, this makes NLL estimation computationally too expensive. Fortunately, when evaluating NELBO (which corresponds to $K=1$ here), the NELBO estimate is unbiased and its variance is small because of averaging across big test datasets (with often 10k samples). For example, on MNIST the standard deviation of our $\log p(\bz_0)$ estimate is 0.36 nat, while the standard deviation of NELBO is 0.07 nat.
\section{Additional Implementation Details}\label{app:impl}
All hyperparameters for our main models are provided in Tab.~\ref{table:hyperparameters}.

\subsection{VAE Backbone}
The VAE backbone for all LSGM models is NVAE~\cite{vahdat2020nvae}\footnote{\url{https://github.com/NVlabs/NVAE} (NVIDIA Source Code License)}, one of the best-performing VAEs in the literature. It has a hierarchical latent space with group-wise autoregressive latent variable dependencies and it leverages residual neural networks (for architecture details see~\cite{vahdat2020nvae}). It uses depth-wise separable convolutions in the decoder. Although both the approximate posterior and the prior are hierarchical in its original version, we can reparametrize the prior and write it as a product of independent Normal distributions (see Sec.~\ref{app:hvae}).

The VAE's most important hyperparameters include the number of latent variable groups and their spatial resolution, the channel depth of the latent variables, the number of residual cells per group, and the number of channels in the convolutions in the residual cells. Furthermore, when training the VAE during the first stage we are using KL annealing and KL balancing, as described in~\cite{vahdat2020nvae}. For some models, we complete KL annealing during the pre-training stage, while for other models we found it beneficial to anneal only up to a KL-weight $\beta_\textrm{KL}<1.0$ in the ELBO during the first stage and complete KL annealing during the main end-to-end LSGM training stage. This provides additional flexibility in learning an expressive distribution in latent space during the second training stage, as it prevents more latent variables from becoming inactive while the prior is being trained gradually. However, when using a very large backbone VAE together with an SGM objective that does not correspond to maximum likelihood training, i.e. $w_\un$- or $w_\re$-weighting, we empirically observe that this approach can also hurt NLL, while slightly improving FID (see \textit{CIFAR10 (best FID)} model).

Note that the VAE Backbone performance for CIFAR10 reported in Tab. 2 in the main paper corresponds to the 20-group backbone VAE (trained to full KL-weight $\beta_\textrm{KL}=1.0$) from the \textit{CIFAR10 (balanced)} LSGM model (see hyperparameter Tab.~\ref{table:hyperparameters}).

\textbf{Image Decoders:} Since SGMs~\cite{song2021scoreSDE} assume that the data is continuous, they rely on uniform dequantization when measuring data likelihood. However, in LSGM, we rely on decoders designed specifically for images with discrete intensity values. On color images, we use mixtures of discretized logistics~\cite{salimans2017pixelcnn++}, and on binary images, we use Bernoulli distributions. These decoder distributions are both available from the NVAE implementation.

\subsection{Latent SGM Prior}
Our denoising networks for the latent SGM prior are based on the NCSN++ architecture from Song et al.~\cite{song2021scoreSDE}, adapted such that the model ingests and predicts tensors according to the VAE's latent variable dimensions. We vary hyperparameters such as the number of residual cells per spatial resolution level and the number of channels in convolutions. Note that all our models use $0.2$ dropout in the SGM prior. Some of our models use upsampling and downsampling operations with anti-aliasing based on Finite Impulse Response (FIR)~\cite{zhang2019shiftinvar}, following Song et al.~\cite{song2021scoreSDE}.

NVAE has a hierarchical latent structure. For small image datasets including CIFAR-10, MNIST and OMNIGLOT all the latent variables have the same spatial dimensions. Thus, the diffusion process input
$\bz_0$ is constructed by concatenating the latent variables from all groups in the channel dimension. Our NVAE backbone on the CelebA-HQ-256 dataset comes with multiple spatial resolutions in latent groups. In this case, we only feed the smallest resolution groups to the SGM prior and assume that the remaining groups have a standard Normal distribution. 

\subsection{Training Details}
To optimize our models, we are mostly following the previous literature. The VAE's encoder and decoder networks are trained using an Adamax optimizer~\cite{kingma2014adam}, following NVAE~\cite{vahdat2020nvae}. In the second stage, the whole model is trained with an Adam optimizer~\cite{kingma2014adam} and we perform learning rate annealing for the VAE network optimization, while we keep the learning rate constant when optimizing the SGM prior parameters. At test time, we use an exponential moving average (EMA) of the parameters of the SGM prior with $0.9999$ EMA decay rate, following~\cite{ho2020denoising,song2021scoreSDE}. Note that, when using the VPSDE with linear $\beta(t)$, we are also generally following~\cite{ho2020denoising,song2021scoreSDE} and use $\beta_0=0.1$ and $\beta_1=20.0$. We did not observe any benefits in using the EMA parameters for the VAE networks.

\subsection{Evaluation Details}
For evaluation, we are drawing samples and calculating log-likelihoods using the probability flow ODE, leveraging black-box ODE solvers, following~\cite{chen2018neuralODE,grathwohl2019ffjord,song2021scoreSDE}. Similar to~\cite{song2021scoreSDE}, we are using an RK45 ODE solver~\cite{dormand1980odes}, based on \texttt{scipy}, using the \texttt{torchdiffeq} interface~\footnote{\url{https://github.com/rtqichen/torchdiffeq} (MIT License)}. Integration cutoffs close to zero and ODE solver error tolerances used for evaluation are indicated in Tab.~\ref{table:hyperparameters} (for example, for the VPSDE with linear $\beta(t)$ we usually use $\vo=0$ and therefore have that $\sigma^2_t$ goes to $0$ at $t=0$, hence preventing us from integrating the probability flow ODE all the way to exactly $0$. This was handled similarly by Song et al.~\cite{song2021scoreSDE}).

Following the conventions established by previous work~\cite{karras2018progressive,song2019scorematching,ho2020denoising,song2021denoising}, when evaluating our main models we compute FID at frequent intervals during training and report FID and NLL at the minimum observed FID.

Vahdat and Kautz in NVAE~\cite{vahdat2020nvae} observe that setting the batch normalization (BN) layers to train mode during sampling (i.e., using batch statistics for normalization instead of moving average statistics) improves sample quality. We similarly observe that setting BN layers to train mode improves sample quality by about 1 FID score on the CelebA-HQ-256 dataset, but it does not affect performance on the CIFAR-10 dataset. In contrast to NVAE, we do not change the temperature of the prior during sampling, as we observe that it hurts generation quality.

\begin{table*}
\setlength{\tabcolsep}{4pt}
\caption{\small Hyperparameters for our main models. We use the same notations and abbreviations as in Tab. 6 in main paper.}
\centering
\resizebox{0.99\linewidth}{!}{
    \begin{tabular}{l|ccc|cc|c|c}
    \toprule
    \textbf{Hyperparameter} & \textbf{CIFAR10} & \textbf{CIFAR10} & \textbf{CIFAR10} & \textbf{CelebA-HQ-256} & \textbf{CelebA-HQ-256} &  \textbf{OMNIGLOT} & \textbf{MNIST} \\
      & (best FID) & (balanced) & (best NLL) & (best quantitative)  & (best qualitative)  &  &  \\
    \midrule\midrule
    \textbf{VAE Backbone}  &  &  &  &  &  &  & \\
    \# normalizing flows  & 0 & 0 & 2 & 2 & 2 & 0 & 0 \\
    \# latent variable scales  & 1 & 1 & 1 & 3 & 2 & 1 & 1 \\
    \# groups in each scale  & 20 & 20 & 4 & 8 & 10 & 3 & 2 \\
    spatial dims. of $\bz$ in each scale  & $16^2$ & $16^2$ & $16^2$ & $128^2$, $64^2$, $32^2$ & $128^2$, $64^2$ &  $16^2$ & $8^2$ \\
    \# channel in $\bz$  & 9 & 9 & 45 & 20 & 20 & 20 & 20 \\
    \# initial channels in enc. & 128 & 128 & 256 & 64 & 64 & 64 & 64 \\
    \# residual cells per group  & 2 & 2 & 3 & 2 & 2 & 3 & 1 \\
    NVAE's spectral reg. $\lambda$   & $10^{-2}$ & $10^{-2}$ & $10^{-2}$ & $3\times10^{-2}$ & $3\times10^{-2}$ & $10^{-2}$ &  $10^{-2}$ \\
    \midrule
    \textbf{Training}  &  &  &  &  &  &  & \\
    \textbf{(VAE pre-training)}  &  &  &  &  &  &  & \\
    \# epochs  & 400 & 600 & 400 & 200 & 200 & 200 & 200 \\
    learning rate VAE & $10^{-2}$ & $10^{-2}$ & $10^{-2}$ & $10^{-2}$ & $10^{-2}$ & $10^{-2}$ & $10^{-2}$  \\
    batch size per GPU  & 32 & 32 & 64 & 4 & 4 & 64 & 100 \\
    \# GPUs  & 8 & 8 & 4 & 16 & 16 & 2 & 2 \\
    KL annealing to  & $\beta_\textrm{KL}{=}0.7$ & $\beta_\textrm{KL}{=}1.0$ & $\beta_\textrm{KL}{=}0.7$ & $\beta_\textrm{KL}{=}1.0$ & $\beta_\textrm{KL}{=}1.0$ & $\beta_\textrm{KL}{=}1.0$ & $\beta_\textrm{KL}{=}0.7$ \\
    \midrule
    \textbf{Latent SGM Prior}  &  &  &  &  &  &  & \\
    \# number of scales  & 3 & 3 & 3 & 4 & 5 & 3 & 2 \\
    \# residual cells per scale  & 8 & 8 & 8 & 8 & 8 & 8 & 8 \\
    \# conv. channels at each scale  & [512]$\times$3 & [512]$\times$3 & [512]$\times$3 & 256, [512]$\times$3 & [320]$\times$2, [640]$\times$3 & [256]$\times$3 & [256]$\times$2 \\
    use FIR~\cite{zhang2019shiftinvar}  & yes & yes & yes & yes & yes & no & no \\
    \midrule
    \textbf{Training}  &  &  &  &  &  &  & \\
    \textbf{(Main LSGM training)}  &  &  &  &  &  &  & \\
    \# epochs  & 1875 & 1875 & 1875 & 1000 & 2000 & 1500 & 800 \\
    learning rate VAE & $10^{-4}$ & $10^{-4}$ & $10^{-4}$ & $10^{-4}$ & - & $10^{-4}$ & $10^{-4}$ \\
    learning rate SGM prior & $10^{-4}$ & $10^{-4}$ & $10^{-4}$ & $10^{-4}$ & $10^{-4}$ & $3\times10^{-4}$ & $3\times10^{-4}$ \\
    batch size per GPU  & 16 & 16 & 16 & 4 & 8 & 32 & 32 \\
    \# GPUs  & 16 & 16 & 16 & 16 & 16 & 4 & 4 \\
    KL annealing  & continued & no & continued & no & no & continued & continued \\
    SDE  & VPSDE & VPSDE & Geo. VPSDE & VPSDE & VPSDE & VPSDE & VPSDE \\
    $\vo$ ($=\vmin$ for Geo. VPSDE)  & 0.0 & 0.0 & $3\times10^{-5}$ & 0.0 & 0.0 & 0.0 & 0.0 \\
    $\vmax$ (only for Geo. VPSDE)  & - & - & 0.999 & - & - & - & - \\
    $t$-sampling cutoff during training  & 0.01 & 0.01 & 0.0 & 0.01 & 0.01 & 0.01 & 0.01 \\
    SGM prior weighting mechanism    & $w_{\un}$ & $w_{\un}$ & $w_{\we}$ & $w_{\re}$ & $w_{\re}$ & $w_{\we}$  & $w_{\we}$ \\
    t-sampling approach (SGM-obj.)    & $r_{\un}(t)$ & $r_{\un}(t)$ & $\U[0, 1]$ & $r_{\re}(t)$ &  $r_{\re}(t)$ & $r_{\we}(t)$ &  $r_{\we}(t)$ \\
    t-sampling approach (q-obj.)  & \textit{rew.} & \textit{rew.} & \textit{rew.} & $r_{\we}(t)$ & - & \textit{rew.} & \textit{rew.} \\
    \midrule
    \textbf{Evaluation}  &  &  &  &  &  &  & \\
    ODE solver integration cutoff    & $10^{-6}$ & $10^{-6}$ & $10^{-6}$ & $10^{-5}$ & $10^{-5}$ & $10^{-5}$ & $10^{-5}$ \\
    ODE solver error tolerance   & $10^{-5}$ & $10^{-5}$ & $10^{-5}$ & $10^{-5}$ &  $10^{-5}$ & $10^{-5}$ & $10^{-5}$ \\
    \bottomrule
\end{tabular}%
}
\label{table:hyperparameters}
\vspace{-0.4cm}
\end{table*}

\subsection{Ablation Experiments}
Here we provide additional details and discussions about the ablation experiments performed in the paper.

\subsubsection{Ablation: SDEs, Objective Weighting Mechanisms and Variance Reduction} \label{app:experiments_sdeetc}
The models that were used for the ablation experiment on SDEs, objective weighting mechanisms and variance reduction and produced the results in Tab. 6 in the main paper use an overall similar setup as the \textit{CIFAR10 (best NLL)} one, with a few exceptions: They are trained only for 1000 epochs and evaluation always happens using the checkpoint at the end of training. Furthermore, the total batchsize over all GPUs is reduced from 256 to 128. Additionally, only 2 instead of 8 cells per residual are used in the latent SGM prior networks. Finally, the VAE's KL term is annealed all the way to $\beta_\textrm{KL}=1.0$ during the first training stage for these experiments. All other hyperparameters correspond to the \textit{CIFAR10 (best NLL)} setup, except those that are explicitly varied as part of the ablation study and mentioned in Tab. 6 in the paper.

As discussed in the main paper, the results of this ablation study overall validate that importance sampling is important to stabilize training, that the $w_\we$-weighting mechanism as well as our novel geometric VPSDE are well suited for training towards strong likelihood, and that the $w_\un$- and $w_\re$-weighting mechanisms tend to produce better FIDs. Although these trends generally hold, it is noteworthy that not all results translate perfectly to our large models that we used to produce our main results. For instance, the setting with $w_\re$-weighting and no importance sampling for the SGM objective, which produced the best FID in Tab. 6 (main paper), is generally unstable for our bigger models, in line with our observation that IS is usually necessary to stabilize training. The stable training run for this setting in Tab. 6 can be considered an outlier.

Furthermore, for CIFAR10 we obtained our very best FID results using the VPSDE, $w_\un$-weighting, IS, and sample reweighting for the $q$-objective, while for the slightly smaller models used for the results in Tab. 6, there is no difference between using sample reweighting and drawing a separate batch $t$ with $r_\we(t)$ for training $q$ for this case (see Tab. 6 main paper, VPSDE, $w_\un$, $r_\un(t)$ fields). Also, CelebA-HQ-256 behaves slightly different for the large models in that the VPSDE with $w_\re$-weighting and sampling a separate batch $t$ with $r_\we(t)$ for $q$-training performed best by a small margin (see hyperparameter Tab.~\ref{table:hyperparameters}).

\subsubsection{Ablation: End-to-End Training}
The model used for the results on the ablation study regarding end-to-end training vs. fully separate VAE and SGM prior training is the same one as used for the ablation study on SDEs, objective weighting mechanisms and variance reduction above, evaluated in a similar way. For this experiment, we used the VPSDE, $w_\un$-objective weighting, IS for $t$ with $r_\un(t)$ when training the SGM prior, and we did draw a second batch $t$ with $r_\we(t)$ for training $q$ (only relevant for the end-to-end training setup). 

\subsubsection{Ablation: Mixing Normal and Neural Score Functions}
The model used for the ablation study on mixing Normal and neural score functions is again similar to the one used for the other ablations with the exception that the underlying VAE has only a single latent variable group, which makes it much smaller and removes all hierarchical dependencies between latent variables. We tried training multiple models with larger backbone VAEs, but they were generally unstable when trained without our mixed score parametrization, which only hightlights its importance. As for the previous ablation, for this experiment we used the VPSDE, $w_\un$-objective weighting, IS for $t$ with $r_\un(t)$ when training the SGM prior, and we did draw a second batch $t$ with $r_\we(t)$ for training $q$.

\subsection{Training Algorithms}
To unambiguously clarify how we train our LSGMs, we summarized the training procedures in three different algorithms for different situations:
\begin{enumerate}
    \item \textit{Likelihood training with IS}. In this case, the SGM prior and the encoder share the same weighted likelihood objective and do not need to be updated separately.
    \item \textit{Un/Reweighted training with separate IS of $t$ for SGM-objective and $q$-objective}. Here, the SGM prior and the encoder need to be updated with different weightings, because the encoder always needs to be trained using the weighted (maximum likelihood) objective. We draw separate batches $t$ using separate IS distribution for the two differently weighted objectives (i.e. last term in Eq. 8 from main paper vs. Eq. 9).
    \item \textit{Un/Reweighted training with IS of $t$ for the SGM-objective and reweighting for the $q$-objective}. What this means is that when training the encoder with the score-based cross entropy term (last term in Eq. 8 from main paper), we are using an importance sampling distribution that was actually tailored to un- or reweighted training for the SGM objective (Eq. 9 from main paper) and therefore isn't optimal for the weighted (maximum likelihood) objective necessary for encoder training. However, if we nevertheless use the same importance sampling distribution, we do not need to draw a second batch of $t$ for encoder training. In practice, this boils down to different (re-)weighting factors in the cross entropy term (see Algorithm 3).
\end{enumerate}

For efficiency comparison between approaches (2) and (3), we observe that (3) consumes more memory than (2) in general but it can be faster due to the shared computation for the denoising step. Due to the memory limitations, we use (2) on large image datasets. Note that the choice between (2) and (3) may affect generative performance as we empirically observed in our experiments.

\begin{algorithm}[H]
\caption{Likelihood training with IS}
    \label{alg:likelihood_uniform}
    \begin{algorithmic}
    \State \textbf{Input:} data $\bx$, parameters $\{ \btheta, \bphi, \bpsi\}$
    \State Draw $\bz_0 \sim q_\bphi(\bz_0|\bx)$ using encoder.
    \State Draw $t \sim r_{\we}(t)$ with IS distribution of likelihood weighting (Sec.~\ref{app:var}).
    \State Calculate $\bmu_t(\bz_0)$ and $\vt$ according to SDE.
    \State Draw $\bz_t \sim q(\bz_t | \bz_0)$ using $\bz_t = \bmu_t(\bz_0) + \vt \beps$ where $\beps \sim \N(\beps, \bzero, \beye)$.
    \State Calculate score $\beps_\btheta(\bz_t, t) = \sigma_t (1 - \balpha) \odot \bz_t + \balpha \odot \beps'_\theta(\bz_t, t)$.
    \State Calculate cross entropy $\CE{q_\bphi(\bz_0|\bx)}{p_\btheta(\bz_0)} \approx \frac{1}{r_{\we}(t)}\frac{w_{\we}(t)}{2} ||\beps\!-\!\beps_\btheta(\bz_t, t)||_2^2$.
    \State Calculate objective $\L(\bx, \btheta, \bphi, \bpsi) = -\log p_\bpsi(\bx|\bz_0)+\log q_\bphi(\bz_0 | \bx) + \CE{q_\bphi(\bz_0|\bx)}{p_\btheta(\bz_0)} $.
    \State Update all parameters $\{ \btheta, \bphi, \bpsi\}$ by minimizing $\L(\bx, \btheta, \bphi, \bpsi)$.
    \end{algorithmic}
\end{algorithm}

\begin{algorithm}[H]
\caption{Un/Reweighted training with separate IS of $t$}
    \label{alg:dual_objective}
    \begin{algorithmic}
    \State \textbf{Input:} data $\bx$, parameters $\{ \btheta, \bphi, \bpsi\}$
    \State Draw $\bz_0 \sim q_\bphi(\bz_0|\bx)$ using encoder.
    \State
    \State \texttt{$\rhd$ Update SGM prior}
    \State Draw $t \sim r_{\un/\re}(t)$ with IS distribution for un/reweighted objective (Sec.~\ref{app:var}).
    \State Calculate $\bmu_t(\bz_0)$ and $\vt$ according to SDE.
    \State Draw $\bz_t \sim q(\bz_t | \bz_0)$ using $\bz_t = \bmu_t(\bz_0) + \vt \beps$ where $\beps \sim \N(\beps, \bzero, \beye)$.
    \State Calculate score $\beps_\btheta(\bz_t, t) = \sigma_t (1 - \balpha) \odot \bz_t + \balpha \odot \beps'_\theta(\bz_t, t)$.
    \State Calculate objective $\L(\btheta) \approx \frac{1}{r_{\un/\re}(t)}\frac{w_{\un/\re}(t)}{2} ||\beps\!-\!\beps_\btheta(\bz_t, t)||_2^2$.
    \State Update SGM prior parameters $\btheta$ by minimizing $\L(\btheta)$.
    \State
    \State \texttt{$\rhd$ Update VAE Encoder and Decoder with new $t$ sample}
    \State Draw $t \sim r_{\we}(t)$ with IS distribution for likelihood weighting (Sec.~\ref{app:var}).
    \State Calculate $\bmu_t(\bz_0)$ and $\vt$ according to SDE.
    \State Draw $\bz_t \sim q(\bz_t | \bz_0)$ using $\bz_t = \bmu_t(\bz_0) + \vt \beps$ where $\beps \sim \N(\beps, \bzero, \beye)$.
    \State Calculate score $\beps_\btheta(\bz_t, t) = \sigma_t (1 - \balpha) \odot \bz_t + \balpha \odot \beps'_\theta(\bz_t, t)$.
    \State Calculate cross entropy $\CE{q_\bphi(\bz_0|\bx)}{p_\btheta(\bz_0)} \approx \frac{1}{r_{\we}(t)}\frac{w_{\we}(t)}{2} ||\beps\!-\!\beps_\btheta(\bz_t, t)||_2^2$.
    \State Calculate objective $\L(\bx, \bphi, \bpsi) = -\log p_\bpsi(\bx|\bz_0)+\log q_\bphi(\bz_0 | \bx) + \CE{q_\bphi(\bz_0|\bx)}{p_\btheta(\bz_0)} $.
    \State Update VAE parameters $\{\bphi, \bpsi\}$ by minimizing $\L(\bx, \bphi, \bpsi)$.
    \end{algorithmic}
\end{algorithm}

\begin{algorithm}[H]
\caption{Un/Reweighted training with IS of $t$ for the SGM objective}
    \label{alg:reweighted}
    \begin{algorithmic}
    \State \textbf{Input:} data $\bx$, parameters $\{ \btheta, \bphi, \bpsi\}$ 
    \State Draw $\bz_0 \sim q_\bphi(\bz_0|\bx)$ using encoder.
    \State Draw $t \sim r_{\un/\re}(t)$ with IS distribution for un/reweighted objective (Sec.~\ref{app:var}).
    \State Calculate $\bmu_t(\bz_0)$ and $\vt$ according to SDE.
    \State Draw $\bz_t \sim q(\bz_t | \bz_0)$ using $\bz_t = \bmu_t(\bz_0) + \vt \beps$ where $\beps \sim \N(\beps, \bzero, \beye)$.
    \State Calculate score $\beps_\btheta(\bz_t, t) = \sigma_t (1 - \balpha) \odot \bz_t + \balpha \odot \beps'_\theta(\bz_t, t)$.
    \State Compute $\L_{DSM} := ||\beps\!-\!\beps_\btheta(\bz_t, t)||_2^2$
    \State
    \State \texttt{$\rhd$ SGM prior loss}
    \State Calculate objective $\L(\btheta) \approx \frac{1}{r_{\un/\re}(t)}\frac{w_{\un/\re}(t)}{2} \L_{DSM}$.
    \State
    \State \texttt{$\rhd$ VAE Encoder and Decoder loss computed with the same $t$ sample}
    \State Calculate cross entropy $\CE{q_\bphi(\bz_0|\bx)}{p_\btheta(\bz_0)} \approx \frac{1}{r_{\un/\re}(t)}\frac{w_{\we}(t)}{2} \L_{DSM}$.
    \State Calculate objective $\L(\bx, \bphi, \bpsi) = -\log p_\bpsi(\bx|\bz_0)+\log q_\bphi(\bz_0 | \bx) + \CE{q_\bphi(\bz_0|\bx)}{p_\btheta(\bz_0)}$.
    \State
    \State \texttt{$\rhd$ Update all parameters}
    \State Update SGM prior parameters $\btheta$ by minimizing $\L(\btheta)$.
    \State Update VAE parameters $\{\bphi, \bpsi\}$ by minimizing $\L(\bx, \bphi, \bpsi)$.
    \State

    \end{algorithmic}
\end{algorithm}

\subsection{Computational Resources}
In total, the research project consumed $\approx 350,000$ GPU hours, which translates to an electricity consumption of about $\approx50$ MWh. We used an in-house GPU cluster of V100 NVIDIA GPUs.

\section{Additional Experiments}\label{app:expr}
\subsection{Additional Samples}
\begin{figure}
    \centering
    \includegraphics[scale=0.6]{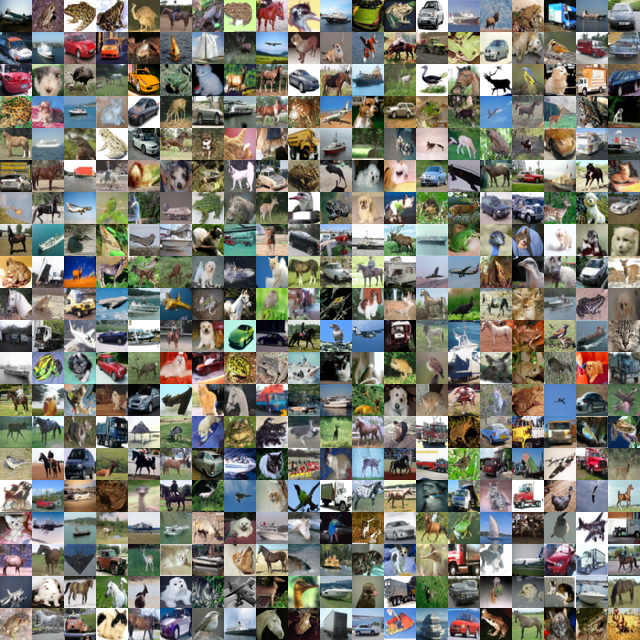}
    \caption{Additional uncurated samples generated by LSGM on the CIFAR-10 dataset (best FID model). Sampling in the latent space is done using the probability flow ODE.}
    \label{fig:cifar10-additional}
\end{figure}

\begin{figure}
    \centering
    \includegraphics[scale=0.31]{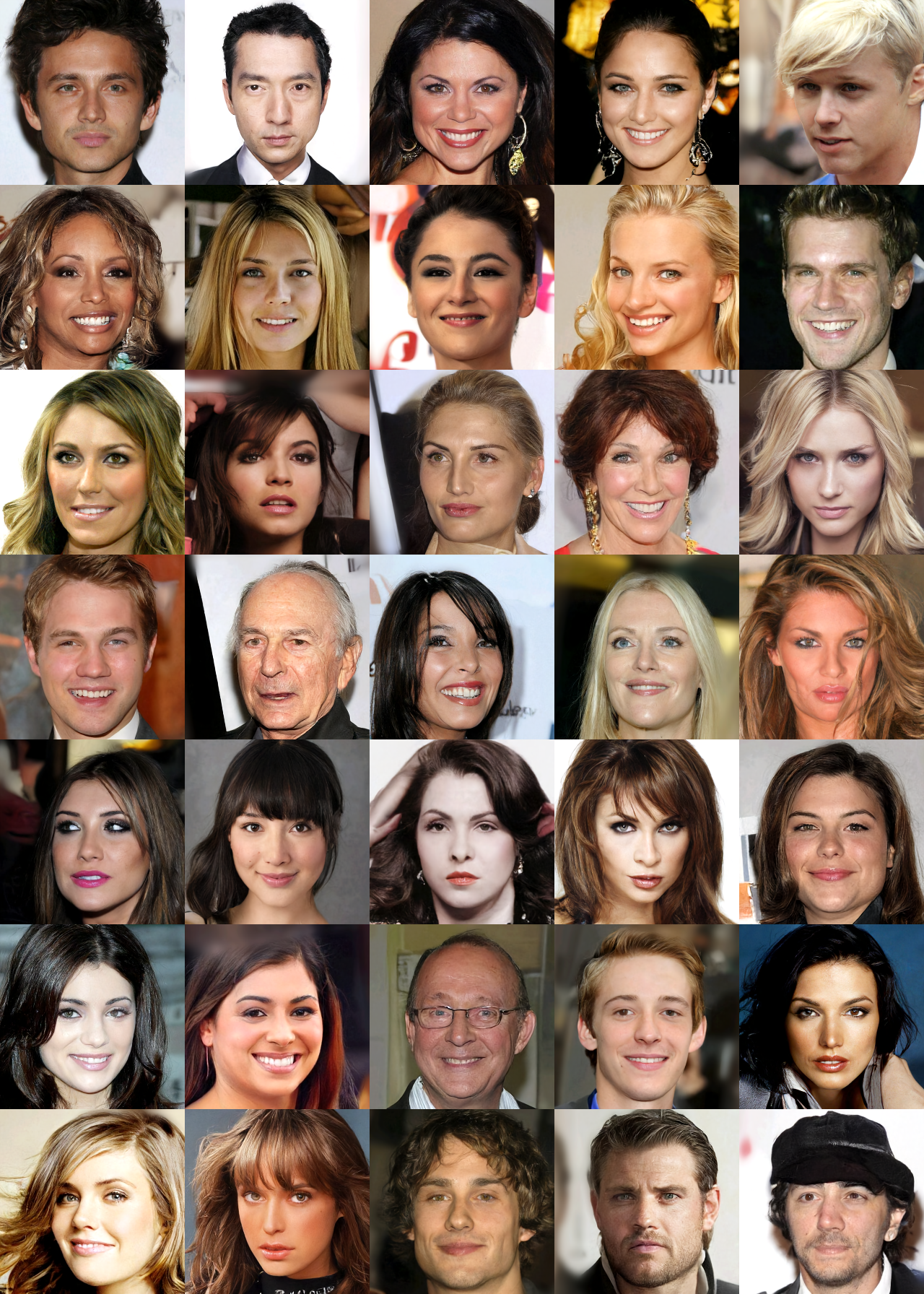}
    \caption{Additional uncurated samples generated by LSGM on the CelebA-HQ-256 dataset. Sampling in the latent space is done using the probability flow ODE.}
    \label{fig:celebA-additional}
\end{figure}

In this section, we provide additional samples generated by our models for CIFAR-10 in Fig.~\ref{fig:cifar10-additional}, and CelebA-256-HQ in Fig.~\ref{fig:celebA-additional}.

\subsection{MNIST: Small VAE Experiment}
Here, we examine our LSGM on a small VAE architecture. We specifically follow \cite{kingma2016improved} and build a small VAE in the NVAE codebase. In particular, the model does not have hierarchical latent variables, but only a single latent variable group with a total of 64 latent variables. Encoder and decoder consist of small ResNets with 6 residual cells in total (every two cells there is a down- or up-sampling operation, so we have 3 blocks with 2 residual cells per block). The experiments are done on dynamically binarized MNIST. As we can see in Table~\ref{table:smallvae}, our implementation of the VAE obtains a similar test NELBO as \cite{kingma2016improved}. However, our LSGM improves the NELBO by almost 4.6 nats. This simple experiment shows that we can even obtain good generative performance with our LSGM using small VAE architectures. 

\begin{table*}
\setlength{\tabcolsep}{16pt}
\centering
\caption{\small Experiment with a small VAE architecture on dynamically binarized MNIST.}
    \begin{tabular}{lcc}
        \toprule
        {\bf Method} & {\bf NELBO $\downarrow$} (nats) \\
        \midrule
        Small VAE~\cite{kingma2016improved} & 84.08$\pm$0.10 \\
        Small VAE + inverse autoregressive flow~\cite{kingma2016improved} & 80.80$\pm$0.07 \\
        \midrule
        Our small VAE & 83.85 \\
        Our LSGM w/ small VAE & \bf 79.23 \\
        \bottomrule
    \end{tabular}%
\label{table:smallvae}
\end{table*}

\subsection{CIFAR-10: Neural Network Evaluations during Sampling}
In Tab.~\ref{table:cifarnfe}, we report the number of neural network evaluations performed by the ODE solver during sampling from our CIFAR-10 models. ODE solver error tolerance is $10^{-5}$ and time integration cutoff is $10^{-6}$. CIFAR-10 is a highly diverse and more multimodal dataset, compared to CelebA-HQ-256. Because of that, the latent SGM prior that is learnt is more complex, requiring more function evaluations.

\begin{table*}
\setlength{\tabcolsep}{16pt}
\centering
\begin{minipage}{0.7\textwidth}
\centering
\caption{\small Number of function evaluations (NFE) of ODE solver during probability flow-based latent SGM prior sampling and corresponding sampling time for our main CIFAR-10 models. Sampling was done in batches of size 16 using a single Titan V GPU. Results are averaged over 20 sampling runs. See Tab.~\ref{table:main_cifar10} in main text for generative performance metrics.}
    \begin{tabular}{lcc}
        \toprule
        {\bf Method} & {\bf NFE $\downarrow$} & {\bf Sampling Time $\downarrow$} \\
        \midrule
        LSGM (FID) &  \hspace{-2.5mm} 138 & 11.07 sec.  \\
        LSGM (NLL) &  \hspace{-2mm} 120 & 9.58 sec.  \\
        LSGM (balanced) & \hspace{-2mm} 128 & 10.26 sec. \\
        \bottomrule
    \end{tabular}%
\label{table:cifarnfe}
\end{minipage}
\end{table*}

\subsection{CIFAR-10: Sub-VPSDE vs. VPSDE} \label{app:subvpvsvp}
In App.~\ref{app:variance_subvpsde} we discussed how variance reduction techniques derived based on the VPSDE can also help reducing the variance of the sample-based estimate of the training objective when using the Sub-VPSDE in the latent space SGM. Here, we perform a quantitative comparison between the VPSDE and the Sub-VPSDE, following the same experimental setup and using the same models as for the ablation study on SDEs, objective weighting mechanisms, and variance reduction (experiment details in App.~\ref{app:experiments_sdeetc}). The results are reported in Tab.~\ref{tab:subvp_vs_vp}. We find that the VPSDE generally performs slightly better in FID, while we observed little difference in NELBO in these experiments. Importantly, the Sub-VPSDE also did not outperform our novel geometric VPSDE in NELBO. We also see that the combination of Sub-VPSDE with $w_\re$-weighting performs poorly. Consequently, we did not explore the Sub-VPSDE further in our main experiments.

\begin{table*}
\small
\setlength{\tabcolsep}{16pt}
\centering
\begin{minipage}{0.8\textwidth}
\centering
\caption{\small Comparing the VPSDE and Sub-VPSDE in LSGM. For detailed explanations of abbreviations in the table, see Tab.~\ref{tab:ablation_obj_weighting} in main paper. Note that importance sampling distributions are generally based on derivations with the VPSDE, even when using the Sub-VPSDE, as discussed in App.~\ref{app:variance_subvpsde}.}
\centering
\resizebox{1.0\linewidth}{!}{
    \begin{tabular}{cc|c|c|c}
    \toprule
    \multicolumn{2}{c|}{\textbf{SGM-obj.-weighting}} & $w_{\we}$ & $w_{\un}$ & $w_{\re}$ \\
    \midrule
    \multicolumn{2}{c|}{$t$\textbf{-sampling (SGM-obj.)}} & $ r_{\we}(t)$ & $r_{\un}(t)$ & $ r_{\re}(t)$ \\
    \midrule
    \multicolumn{2}{c|}{$t$\textbf{-sampling (q-obj.)}} & \textit{rew.} & $r_{\we}(t)$ & $r_{\we}(t)$ \\
    \midrule\midrule
    \multirow{2}{*}{\textbf{VPSDE}} 
     & \textbf{FID}$\downarrow$    & 8.00 & 5.39 & 6.19 \\
     & \textbf{NELBO}$\downarrow$   & 2.97 & 2.98 & 2.99 \\
    \midrule
    \multirow{2}{*}{\textbf{Sub-VPSDE}}
    & \textbf{FID}$\downarrow$    & 8.46 & 5.73 & 19.10 \\
    & \textbf{NELBO}$\downarrow$    & 2.97 & 2.97 & 3.04 \\
    \bottomrule
\end{tabular}%
}
\label{tab:subvp_vs_vp}
\end{minipage}
\vspace{-0.4cm}
\end{table*}

\subsection{CelebA-HQ-256: Different ODE Solver Error Tolerances}
\begin{figure}[!t]
     \centering
     \hspace{-0.2cm}
     \begin{subfigure}[b]{\textwidth}
         \centering
         \includegraphics[scale=0.5]{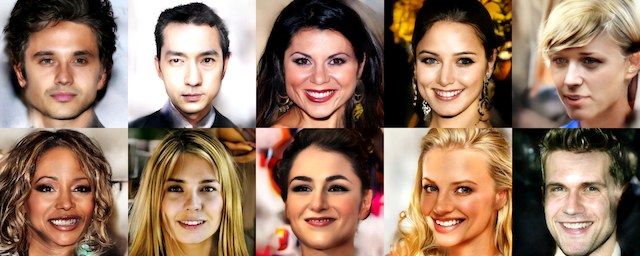}
         \caption{ODE solver error tolerance $10^{-2}$}
     \end{subfigure} \bigskip \\
     \begin{subfigure}[b]{\textwidth}
         \centering
         \includegraphics[scale=0.5]{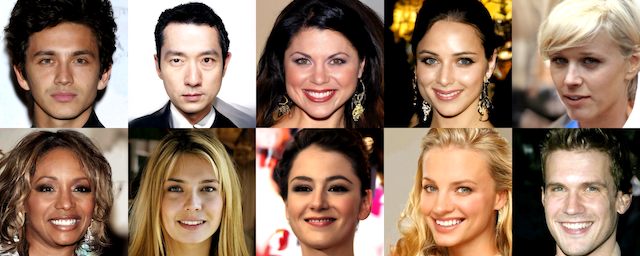}
         \caption{ODE solver error tolerance $10^{-3}$}
     \end{subfigure} \smallskip \\
     \begin{subfigure}[b]{\textwidth}
         \centering
         \includegraphics[scale=0.5]{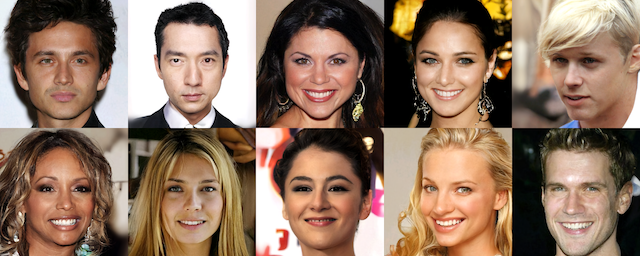}
         \caption{ODE solver error tolerance $10^{-4}$}
     \end{subfigure} \smallskip \\
     \begin{subfigure}[b]{\textwidth}
         \centering
         \includegraphics[scale=0.5]{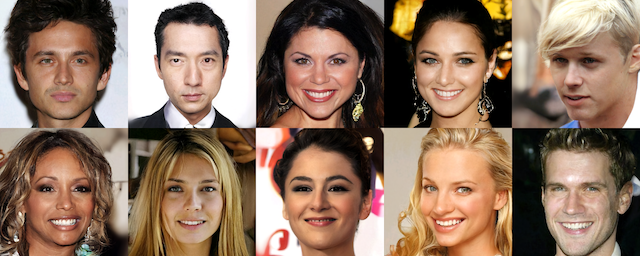}
         \caption{ODE solver error tolerance $10^{-5}$}
     \end{subfigure}
    \caption{\small The effect of ODE solver error tolerance on the quality of samples. In contrast to the original SGM~\cite{song2021scoreSDE} where high error tolerance results in pixelated images (see Fig.~3 in~\cite{song2021scoreSDE}), in our case high error tolerances create low-frequency artifacts. Reducing the error tolerance improves subtle details slightly.}
    \label{fig:error_tolerance}
    \vspace{-0.3cm}
\end{figure}

In Fig.~\ref{fig:error_tolerance}, we visualize CelebA-HQ-256 samples from our LSGM model for varying ODE solver error tolerances. 

\subsection{CelebA-HQ-256: Ancestral Sampling}
\begin{figure}
    \centering
    \includegraphics[scale=0.31]{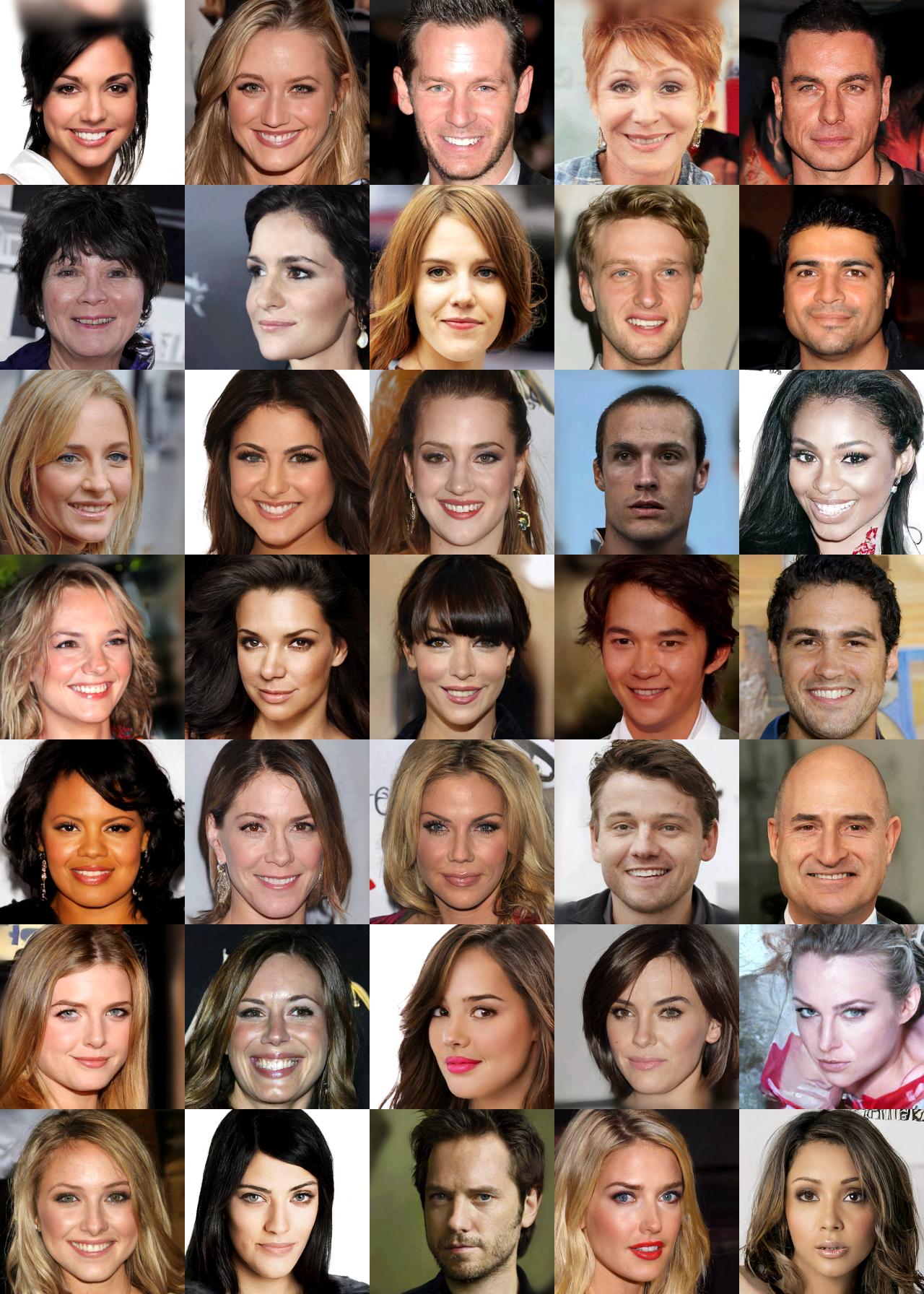}
    \caption{Uncurated samples generated by LSGM on the CelebA-HQ-256 dataset using 200-step ancestral sampling for the prior.}
    \label{fig:celebA-ancestral-200}
\end{figure}

\begin{figure}
    \centering
    \includegraphics[scale=0.31]{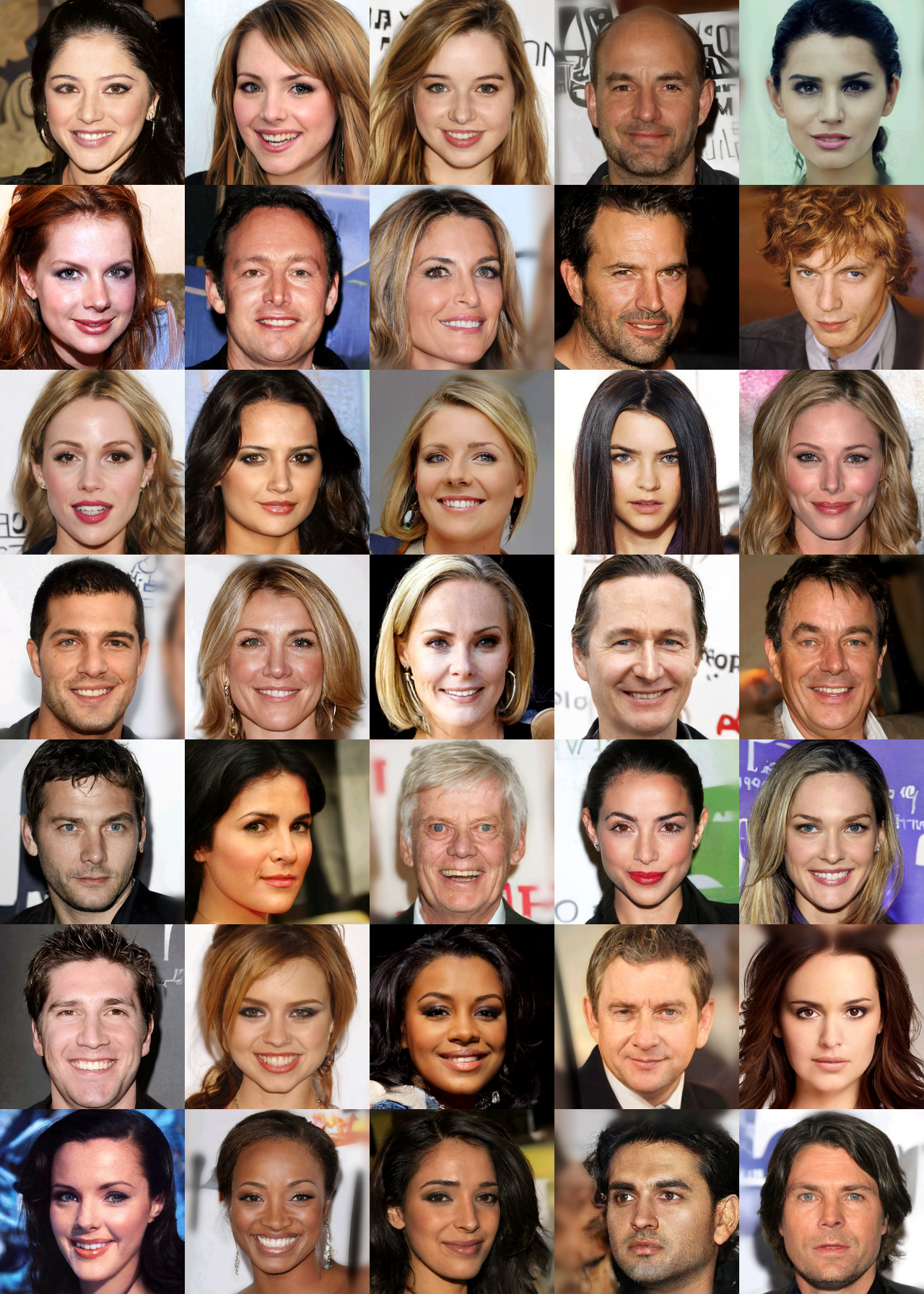}
    \caption{Uncurated samples generated by LSGM on the CelebA-HQ-256 dataset using 1000-step ancestral sampling for the prior. }
    \label{fig:celebA-ancestral-1000_1}
\end{figure}

\begin{figure}
    \centering
    \includegraphics[scale=0.31]{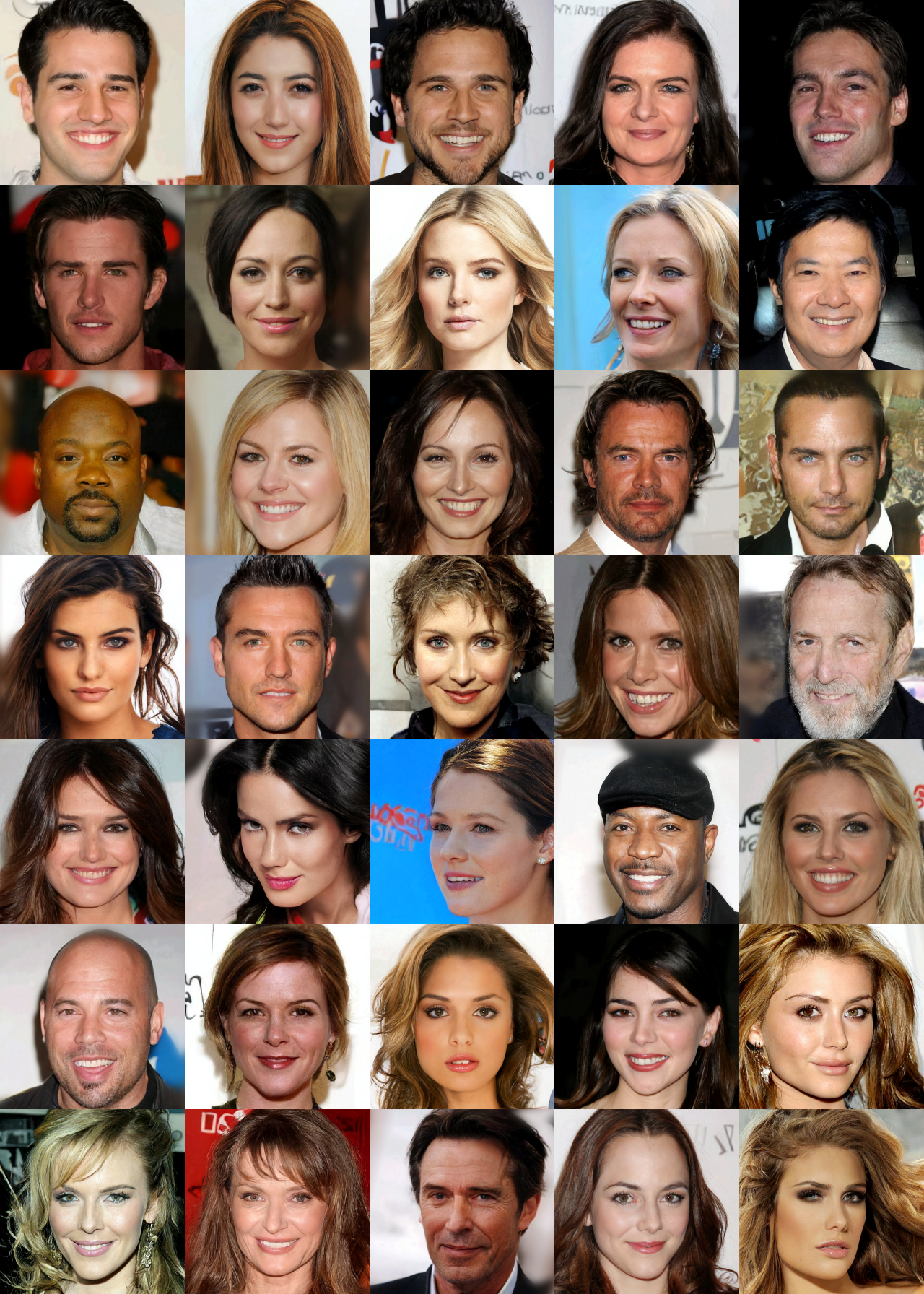}
    \caption{Additional uncurated samples generated by LSGM on the CelebA-HQ-256 dataset using 1000-step ancestral sampling. }
    \label{fig:celebA-ancestral-1000_2}
\end{figure}

For our experiments in this paper, we use the probability flow ODE to sample from the model. However, on CelebA-HQ-256, we observe that ancestral sampling~\cite{song2021scoreSDE, ho2020denoising, sohldickstein2015thermodynamics} from the prior instead of solving the probability flow ODE often generates much higher quality samples. However, the FID score is slightly worse for this approach. In Fig.~\ref{fig:celebA-ancestral-200}, Fig.~\ref{fig:celebA-ancestral-1000_1}, and Fig.~\ref{fig:celebA-ancestral-1000_2}, we visualize samples generated with different numbers of steps in ancestral sampling.

\subsection{CelebA-HQ-256: Sampling from VAE Backbone vs. LSGM}
\begin{figure}
    \centering
    \includegraphics[scale=0.31]{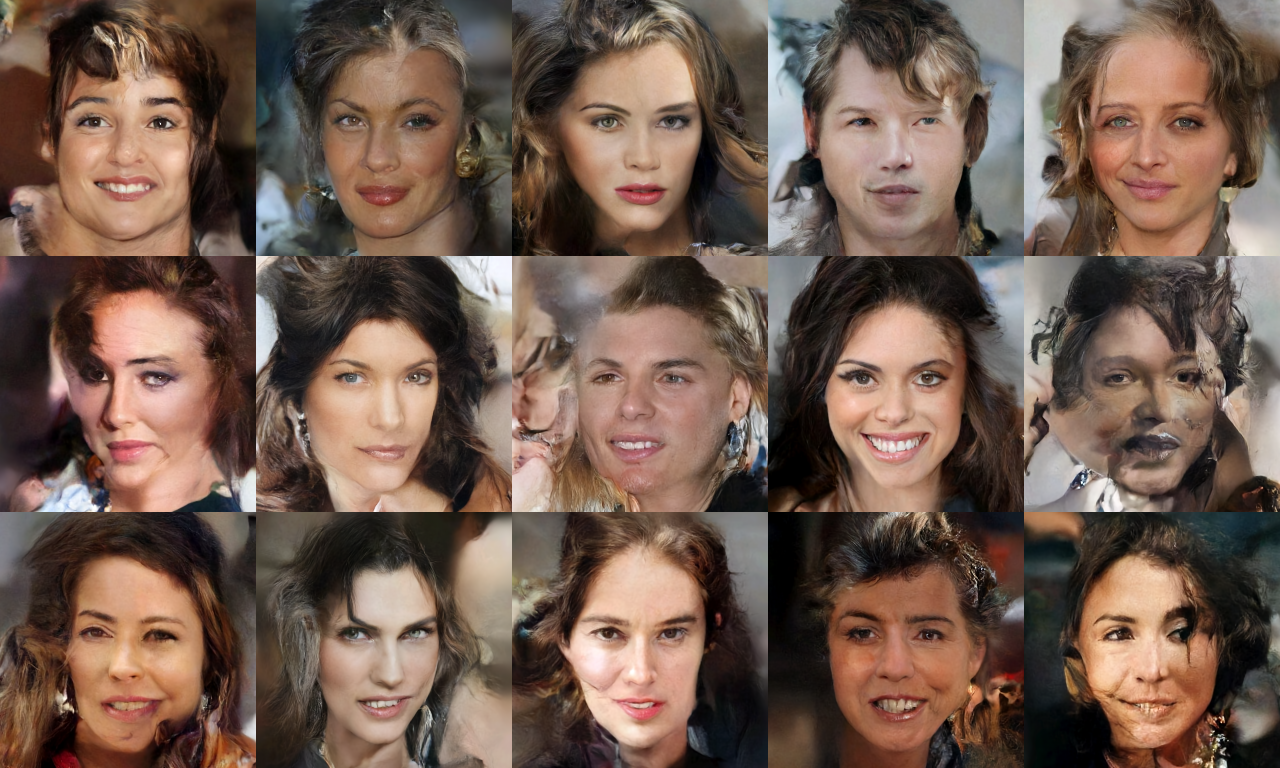}
    \caption{Uncurated samples generated by our VAE backbone without changing the temperature of the prior. The poor quality of the samples from the VAE backbone is partially due to the large spatial dimensions of the latent space in which long-range correlations are not encoded well. }
    \label{fig:VAE_backbone}
\end{figure}

\begin{figure}
    \centering
    \includegraphics[scale=0.31]{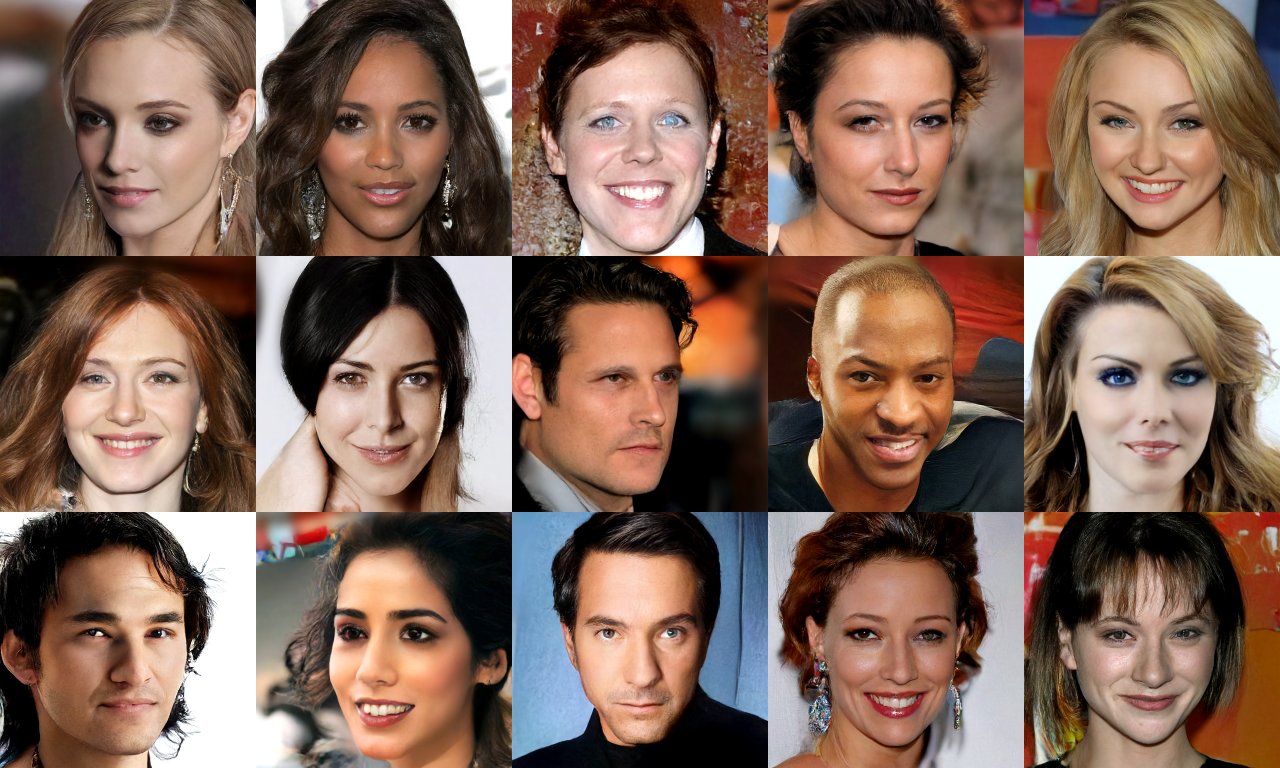}
    \caption{Uncurated samples generated by LSGM with the SGM prior applied to the latent variables of 32$\times$32 spatial dimensions, on the CelebA-HQ-256 dataset. Sampling in the latent space is done using the probability flow ODE.}
    \label{fig:celebA_small}
\end{figure}

For the quantitative results on the CelebA-HQ-256 dataset in the main text, we use an LSGM with spatial dimension of 32$\times$32 for the latent variables in the SGM prior. However, for the qualitative results we used an LSGM with the prior spatial dimension of 64$\times$64. The 32$\times$32 dimensional model achieves a better FID score compared to the 64$\times$64 dimensional model (FID 7.22 vs. 8.53) and sampling from it is much faster (2.7 sec. vs. 39.9 sec.). However, the visual quality of the samples is slightly worse. In this section, we visualize samples generated by the 32$\times$32 dimensional model as well as the VAE backbone for this model. In this experiment, the VAE backbone is fully trained. Samples from our VAE backbone are visualized in Fig.~\ref{fig:VAE_backbone} and for our 32$\times$32 dimensional LSGM in Fig.~\ref{fig:celebA_small}.

\subsection{Evolution Samples on the ODE and SDE Reverse Generative Process}
\begin{figure}
     \centering
     \hspace{-0.2cm}
     \begin{subfigure}[b]{\textwidth}
         \centering
         \includegraphics[scale=0.22]{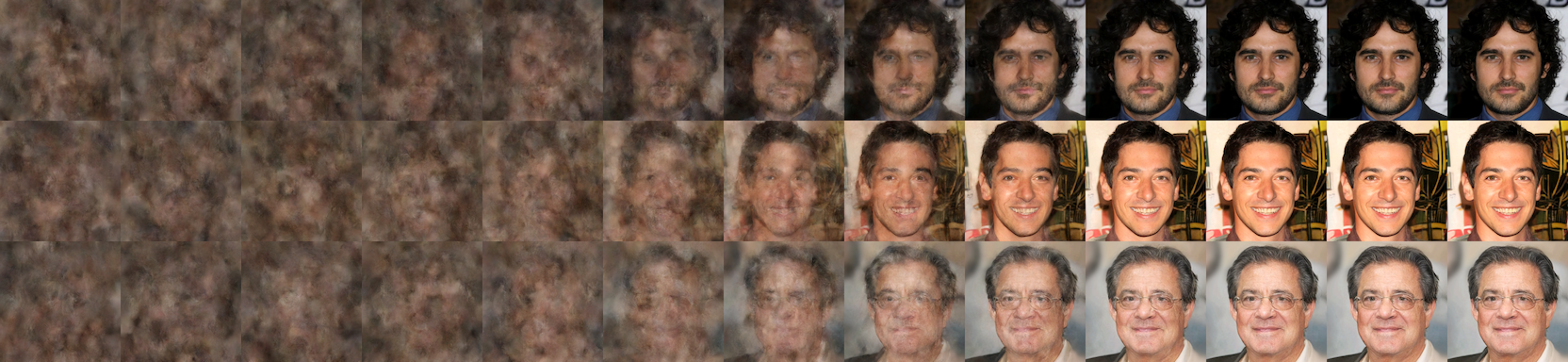}
         \caption{Evolution of latent variables under the SDE}
     \end{subfigure} \bigskip \\
     \begin{subfigure}[b]{\textwidth}
         \centering
         \includegraphics[scale=0.22]{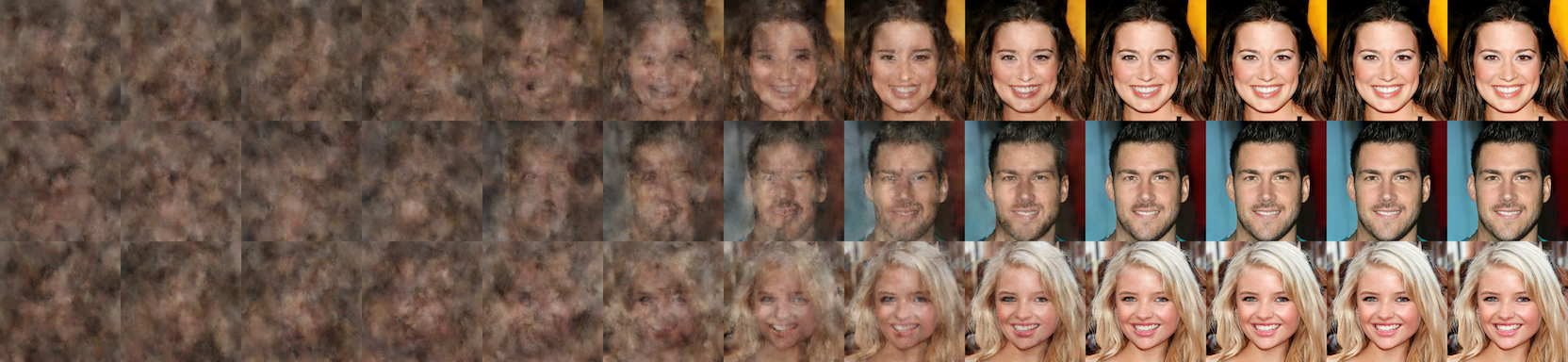}
         \caption{Evolution of latent variables under the SDE}
     \end{subfigure} \smallskip \\
     \begin{subfigure}[b]{\textwidth}
         \centering
         \includegraphics[scale=0.22]{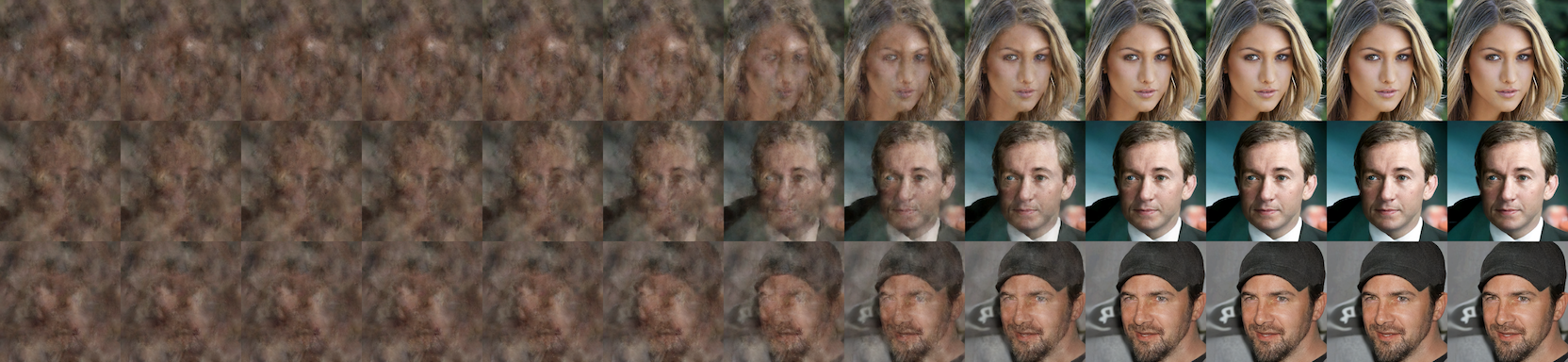}
         \caption{Evolution of latent variables under the ODE}
     \end{subfigure} \bigskip \\
     \begin{subfigure}[b]{\textwidth}
         \centering
         \includegraphics[scale=0.22]{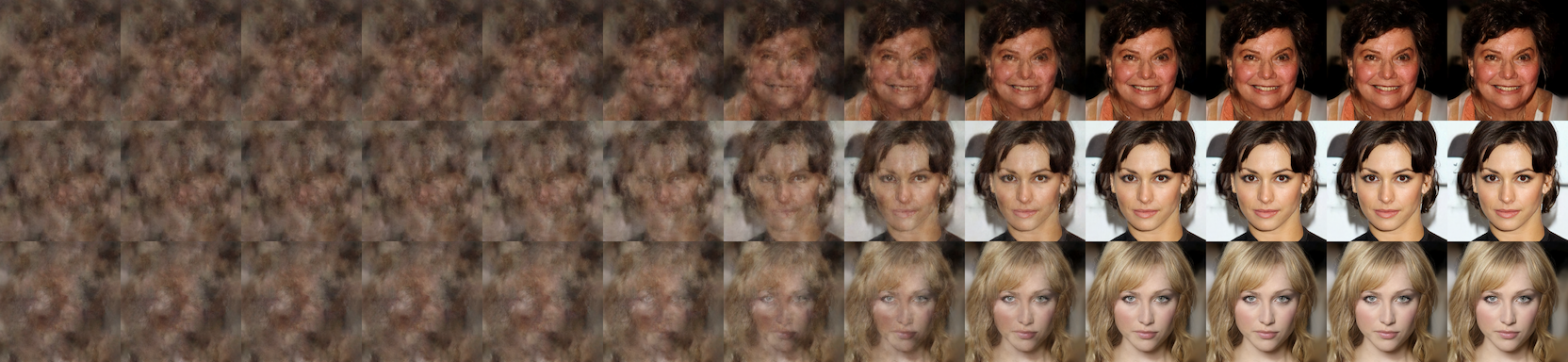}
         \caption{Evolution of latent variables under the ODE}
     \end{subfigure} \smallskip 
    \caption{We visualize the evolution of the latent variables under both the reverse generative SDE (a-b) and also the probability flow ODE (c-d). Specifically, we feed latent variables from different stages along the generative denoising diffusion process to the decoder to map them back to image space. The 13 different images in each row correspond to the times $t=[1.0, 0.9, 0.8, 0.7, 0.6, 0.5, 0.4, 0.3, 0.2, 0.1, 0.05, 0.01, 10^{-5}]$ along the reverse denoising diffusion process. The evolution of the images is noticeably different from diffusion models that are run directly in pixel space (see, for example, Fig. 1 in \cite{song2021scoreSDE}).}
    \label{fig:latent_trajectory}
\end{figure}

In Fig.~\ref{fig:latent_trajectory}, we visualize the evolution of the latent variables under both the reverse generative SDE and also the probability flow ODE. We are decoding the intermediate latent samples along the reverse-time generative process via the decoder to pixel space.

\end{document}